\documentclass[11pt,oneside,final]{ntu_qe}

\newcommand{\numcol}{1}
\usepackage[svgnames, username, dvipsnames]{xcolor}
\usepackage{caption}
\usepackage{subcaption}
\usepackage{NTUdis}
\usepackage{setspace}
\usepackage{graphicx} 
\graphicspath{{Figures/}}
\usepackage{float}
\usepackage{algorithm}
\usepackage{amsfonts}
\usepackage{ifthen}
\usepackage{booktabs}
\usepackage{amssymb}
\usepackage{supertabular}
\usepackage{array}
\usepackage{multirow}
\usepackage{fancyhdr}
\usepackage{fancyheadings}
\usepackage{psfrag}
\usepackage{amsmath}
\usepackage{hhline}
\usepackage{type1cm}
\usepackage{lettrine}
\usepackage{cite}
\usepackage{fancybox}
\usepackage{comment}
\usepackage{hyperref}
\usepackage[titletoc]{appendix}
\usepackage{pdflscape}
\usepackage{tabularx}
\usepackage{url}
\usepackage{ulem}
\usepackage{tabularx}
\usepackage{array}
\usepackage{colortbl}
\usepackage{adjustbox}
\usepackage{mathtools}

\usepackage{tikz}
\usepackage{enumitem}
\usetikzlibrary{positioning}
\usepackage{ctable}
\newif\ifblackandwhite

\usepackage[hmargin=2cm,vmargin=2.5cm]{geometry}
\usepackage{etoolbox}
\usepackage{longtable}%

\usepackage{colortbl}%

\usepackage{booktabs}

\ifblackandwhite
  
\else

\fi

\makeatletter
\let\@currsize\normalsize
\makeatother

\newcommand{\checkOneCol}{\ifthenelse{\numcol = 1}}

\newcommand{\wabstract} {\begin{abstract}}
\newcommand{\wendabstract} {\end{abstract}}

\newcommand{\weqnarray}{\begin{eqnarray}}
\newcommand{\wendeqnarray} {\end{eqnarray}}
\newcommand{\waligneqnarray}{\begin{align}}
\newcommand{\walignendeqnarray} {\end{align}}

\newcommand{\wtable}{\begin{table}}
\newcommand{\wendtable}{\end{table}}

\newcommand{\wfigure}{\begin{figure}}
\newcommand{\wendfigure}{\end{figure}}

\newcommand{\wlist}{\begin{itemize}}
\newcommand{\wendlist}{\end{itemize}}
\newcommand{\wdesc}{\begin{description}}
\newcommand{\wenddesc}{\end{description}}

\newcommand{\wnumlist}{\begin{enumerate}}
\newcommand{\wendnumlist}{\end{enumerate}}

\newtheorem{theorem}{Theorem}
\newtheorem{lemma}{Lemma}
\newtheorem{corollary}{Corollary}
\newtheorem{definition}{Definition}
\newtheorem{example} {Example}
\newtheorem{case} {Case}
\newcommand{\wtheorem}{\begin{theorem}}
\newcommand{\wendtheorem}{\end{theorem}}
\newcommand{\wlemma}{\begin{lemma}}
\newcommand{\wendlemma}{\end{lemma}}
\newcommand{\wcorollary}{\begin{corollary}}
\newcommand{\wendcorollary}{\end{corollary}}
\newcommand{\wdefinition}{\begin{definition}}
\newcommand{\wenddefinition}{\end{definition}}
\newcommand{\wexample}{\begin{example}}
\newcommand{\wendexample}{\end{example}}
\newcommand{\wproof}{\begin{proof}}
\newcommand{\wendproof}{\end{proof}}
\newcommand{\wcase}{\begin{case}}
\newcommand{\wendcase}{\end{case}}
\newcommand{\eg}{e.g., }
\newcounter{ifDRR}
\newcommand{\spellDRR}{Deficit Round Robin}
\newcommand{\abbrevDRR}{DRR}
\newcommand{\DRR}{%
\ifthenelse{\value{ifDRR}= 0}%
{\spellDRR\  (\abbrevDRR) \cite{scheduler:drr}\setcounter{ifDRR}{1}}%
{\abbrevDRR}%
}
\newcounter{ifGOODPUT}
\newcommand{\spellGOODPUT}{goodput}
\newcommand{\goodput}{%
\ifthenelse{\value{ifGOODPUT}=0}%
{\spellGOODPUT\  
\footnote{Goodput is the effective throughput, 
as determined by the data successfully
received and decoded at the receiver. }
\setcounter{ifGOODPUT}{1}}%
{\spellGOODPUT}%
}
\newcommand{\setonehalfspace}{\hsp}
\newcommand{\setdoublespace}{\dsp}

\begin{document}
\setdoublespace

\newcommand{\HRule}[1]{\rule{\linewidth}{#1}} 	

\thispagestyle{empty}

\makeatletter							
\def\printtitle{%
    {\centering \@title\par}}
\makeatother									

\makeatletter							
\def\printauthor{%
    {\centering \large \@author}}				
\makeatother							

\title{	\normalsize \textsc{} 	
		 	\\[2.0cm]								
			\HRule{0.5pt} \\						
			\LARGE \textbf{{Understanding Video Transformers for Segmentation: A Survey of Application and Interpretability}}	
			\HRule{2pt} \\ [0.5cm]		
			\normalsize 
		}

\author{
	Rezaul Karim and Richard P. Wildes\\	
        Lassonde School of Engineering\\
        Department of Electrical Engineering and Computer Science\\
        York University\\
}

\printtitle					
  	\vfill
\printauthor				
\hsp

\begin{nabstract}
Video segmentation encompasses a wide range of categories of problem formulation, e.g., object, scene, actor-action and multimodal video segmentation, for delineating task-specific scene components with pixel-level masks. Recently, approaches in this research area shifted from concentrating on ConvNet-based to transformer-based models. In addition, various interpretability approaches have appeared for transformer models and video temporal dynamics, motivated by the growing interest in basic scientific understanding, model diagnostics and societal implications of real-world deployment. Previous surveys mainly focused on ConvNet models on a subset of video segmentation tasks or transformers for classification tasks. Moreover, component-wise discussion of transformer-based video segmentation models has not yet received due focus. In addition, previous reviews of interpretability methods focused on transformers for classification, while analysis of video temporal dynamics modelling capabilities of video models received less attention. In this survey, we address the above with a thorough discussion of various categories of video segmentation, a component-wise discussion of the state-of-the-art transformer-based models, and a review of related interpretability methods. We first present an introduction to the different video segmentation task categories, their objectives, specific challenges and benchmark datasets. Next, we provide a component-wise review of recent transformer-based models and document the state of the art on different video segmentation tasks. Subsequently, we discuss post-hoc and ante-hoc interpretability methods for transformer models and interpretability methods for understanding the role of the temporal dimension in video models. Finally, we conclude our discussion with future research directions.
\newpage
\end{nabstract}

\setonehalfspace
\addcontentsline{toc}{section}{Table of Contents}
\tableofcontents
\listoffigures
\listoftables

\newpage
\startarabicpagination

\hsp

\chapter{Introduction}\label{CH1}

\section{Motivation}
Over the past several years, computer vision research has shifted from concentrating on ConvNet-based to transformer-based models. This transition is evident from the proposal of numerous transformer models for video classification (\eg~\cite{arnab2021vivit,liu2021swin,fan2021multiscale,liu2022video,li2022mvitv2}) and segmentation (\eg~\cite{mei2021transvos,duke2021sstvos,cheng2021mask2formervid}). In parallel, video segmentation continues to be a subject of great interest due to its theoretical as well as practical importance. From a theoretical perspective, video segmentation addresses the fundamental problem of assigning category membership to video on a pixel-wise basis. From a practical standpoint, applications to downstream video understanding tasks, e.g., autonomous driving and video editing, also contribute to continuing growth in interest from the research community.

Corresponding to the recent emphasis on transformers and video segmentation, a number of related literature reviews have appeared, e.g., \cite{wang2021survey, vandenhende2021multi, selva2023video}. Still, there remains a gap, as surveys have yet to appear that provide a detailed study of transformers for video segmentation. More specifically, the following gaps need to be filled. First, recent surveys of approaches to video segmentation mainly cover ConvNet-based methods to the problem~\cite{yao2020video,wang2021survey}. Alternatively, recent surveys of video transformers focus on video classification~\cite{selva2023video}. To our knowledge, surveys of transformer-based video segmentation models have yet to be available. Second, recent surveys of video segmentation mainly focused on two categories of video segmentation: object-based and semantic-based segmentation~\cite{wang2021survey}. However, due to the wide variety of video segmentation tasks and the recent formulation of several new tasks (e.g., depth-aware video panoptic segmentation~\cite{qiao2021vip} and audio-guided video object segmentation~\cite{pan2022wnet}), there is a need for a survey with a taxonomy of task categories. Third, previous surveys do not include component-wise discussion of the segmentation approaches~\cite{yao2020video,wang2021survey}. However, recent models for different types of video segmentation exhibit several high-level components with a specialization in the design of the components. Several unified models also have been proposed to do multiple segmentation tasks using a single model. Component-wise discussion of video segmentation models can play a vital role in understanding the generalization of multiple segmentation tasks in a single model. Fourth, although there is some recent research on the interpretability of transformers in vision, there is a gap in the availability of a survey covering transformer interpretability.

To address the above concerns, we present a survey on transformers for video segmentation. In this survey, we compile a taxonomy of video segmentation tasks, a component-wise analysis of recent transformer models for video segmentation and a discussion of transformer interpretability.

\section{Overview of video segmentation}
Video segmentation is a fundamental computer vision task that divides each frame into regions based on specific qualitative objectives, \eg delineation of objects, motion, actions and semantics. Segmentation algorithms assign each pixel a label from a finite set of classes to delineate the pixel's membership. This label set depends on the qualitative criteria that define an output space based on the application of interest. Based on the qualitative criteria, there are various categories of video segmentation tasks, e.g., object segmentation, actor-action segmentation and semantic segmentation~\cite{wang2021survey}. This inference from colour space to label space supports analysis and application for downstream objectives of an artificial intelligence processor or a human observer. Applications of interest include video surveillance~\cite{tian2005robust}, human-object interaction~\cite{zhou2020cascaded} and autonomous driving~\cite{chen2015deepdriving}.

Video segmentation shares many challenges with single image segmentation (e.g., photometric and geometric variation within a segment); however, additional challenges arise in the setting of temporal image streams, as segments need to be delineated and tracked or otherwise associated across time. Attendant challenges include occlusion, motion blur, camera motion and object deformation. Moreover, operation with multiple video frames incurs additional computational expense compared to single frames.

Video segmentation is important for reasons both theoretical and practical. From a theoretical perspective, association of pixels according to common attributes (e.g., colour, texture, dynamics) to form meaningful segments (\eg objects, motion, actions) is a fundamental problem in visual information processing. Video segmentation also is useful for various applications including video compression~\cite{hadizadeh2013saliency}, object tracking~\cite{kim2002fast}, activity recognition~\cite{holloway2014activity}, video surveillance~\cite{tian2005robust}, robotics~\cite{xu2016deep} and autonomous driving~\cite{chen2015deepdriving}. More generally, video segmentation allows for manipulation of each segment independently, tracking objects of interest, recognizing human activities, analyzing scene content, compressing video data and retrieving specific segments or objects from a video database based on their semantic content or visual appearance. Despite being very challenging, the theoretical interest and application areas have driven a continuing increase in research efforts on video segmentation.

\section{Early approaches to video segmentation}
Given its theoretical and practical motivation, video segmentation has attracted much research effort for decades \cite{wang2021survey,thounaojam2014survey,zhang2006overview}. Although numerous categories of video segmentation have proliferated lately \cite{wang2021survey}, early approaches primarily focused on the most basic video segmentation task of foreground-background separation. Early proposals for video segmentation are mainly analytic approaches that apply geometric constraints along with heuristics to design a solution. In addition to their historical importance, some analytic methods or components from those approaches are still relevant in the literature. The dominant analytical techniques of the early attempts fall into four categories: background subtraction, trajectory-based, object proposal-based and saliency-based. The remainder of this subsection overviews each of these approaches.


\paragraph{Background subtraction.}
Background subtraction approaches typically rely on heuristic assumptions about the background in the scene, e.g., pixels along frame boundaries are likely to be background. These approaches then attempt to detect moving objects by comparing pixel attributes to those of the hypothesized background. Connected component algorithms are used to estimate the connected region corresponding to an object or other distinctive region. Therefore, the above process is called background subtraction. Overall, video object segmentation is achieved by constructing a representation of the scene called the background model and then finding deviations from the model for each input frame.

There are various background subtraction methods that have been proposed in the literature. The variety of the proposed approaches primarily differ in terms of the heuristics and types of background model used \eg stationary background \cite{wren1997pfinder, stauffer2000learning,elgammal2002background,han2011density}, background undergoing 2D parametric motion \cite{ren2003statistical,barnich2010vibe} and background undergoing 3D motion \cite{irani1998unified,torr1998concerning,brutzer2011evaluation}. They commonly use multi-class statistical models of colour pixels with a single Gaussian or a Gaussian Mixture Model (GMM). All of these methods rely on the restrictive assumption that the camera is stable and slowly moving; so, they are sensitive to model selection (2D or 3D) and cannot handle many challenging cases such as non-rigid objects and dynamic backgrounds. 
A detailed overview and comparison of background subtraction based video object segmentation is available elsewhere~\cite{brutzer2011evaluation}.

\paragraph{Point trajectories.}
Other early analytic approaches for video segmentation initially cluster point trajectories and use them as prior information to obtain the foreground object. The common motivation of these approaches is considering motion as a perceptual cue for segmenting a video into separate objects and point trajectories have potential to support analysis of motion information over long periods~\cite{shi1998motion}. Interestingly, the notion of ``common fate'' for grouping points and objects can be traced to Gestalt psychology studies of perceptual organization \cite{kohler1970gestalt}. Point trajectory-based methods generally use optical flow~\cite{brox2010object,ochs2013segmentation,chen2015video} or feature tracking ~\cite{shi2000normalized,ochs2012higher,fragkiadaki2012video} as a way of obtaining motion information. Optical flow allows the analysis of motion information between two consecutive frames using dense motion fields. Feature tracking allows object motion analysis over a longer range by tracking a sparse set of key feature points over a larger temporal extent of a video sequence. Methods with point trajectories benefit from explicitly exploiting motion information compared to background subtraction methods. Still, they face the challenge of grouping local motion estimates into coherent segments.


\paragraph{Object proposals.}
Object proposal based approaches generate likely regions (proposals) where objects appear based on colour, texture and motion. The proposals are then ranked based on confidence scores for subsequent aggregation across those that are highly ranked~\cite{lee2011key,ma2012maximum,zhang2013video,fu2014object,koh2017primary,koh2018sequential}. These methods vary in the information used to generate object proposals, the object proposal generation methods and the method of ranking and aggregating those proposals. By definition, these approaches are limited by their particular definitions of objects. Indeed, they typically are not applicable to segmentation tasks where the regions of interest cannot be readily considered as objects, e.g., semantic segmentation where interest may include texture and colour defined regions.

\paragraph{Saliency based approaches.}
Other video segmentation approaches use saliency information as a prior or seed ~\cite{rahtu2010segmenting,papazoglou2013fast,faktor2014video,wang2015saliency,tsai2016semantic,hu2018unsupervised}. Typically, these approaches compute local salience based on feature contrast, with features being measurements of colour, texture and motion. Subsequently, local salience is aggregated across space and/or time in various ways, e.g., using Conditional Random Fields (CRFs) \cite{rahtu2010segmenting},  Non-Local Consensus voting (NLC) \cite {faktor2014video} or graph-based methods \cite{hu2018unsupervised,wang2015saliency}. Similar to object proposals, saliency approaches are limited by their definitions of salience and likely are inapplicable to cases where the regions of interest are not the most salient, e.g., small or low contrast regions.

\section{Overview of the transformer model}
The transformer model was originally proposed for machine translation as a fully attention-based deep neural  network~\cite{vaswani2017attention}. Here, attention refers to the model performing data associations between elements of its input, i.e., attending to those elements. The model is a convolution free alternative to recurrent models for machine translation~\cite{luong2015effective,wu2016google}. The transformer model works on a set of tokens with the representation of each token as a feature vector ($N\times C$) where $N$ is the length of the sets and $C$ is the feature dimension. The proposed model processes the entire input set as a whole, in contrast to the sequential processing of recurrent models. Processing the entire input at once allows better long range context aggregation in transformers compared to ConvNets. Subsequently, the transformer model was adapted to computer vision in the Vision Transformer (ViT) model \cite{dosovitskiy2020image}. In this adaptation, ViT transformed an image into a set of tokens by dividing it into patches and representing each patch as a single token. Further adaptation of transformers to videos operates by making spatiotemporal patches \cite{arnab2021vivit}. Following the success of ViT, there is a surge of research in the computer vision community for various image and video understanding tasks, largely outperforming the previously dominant ConvNet-based approaches, \eg \cite{selva2023video,ulhaq2022vision,tay2022efficient}.

Transformer models are constructed on self-attention and cross-attention mechanisms. Intuitively, self-attention is designed to capture the relationship among the different elements of the same token set, while cross-attention is designed to capture the relationship among the elements of two different sets of tokens. The transformer model follows a general encoder-decoder architecture. A transformer encoder uses self-attention and linear layers as building blocks. While, transformer decoder building blocks consist of self-attention, cross-attention and linear layers. Both the self-attention and cross-attention mechanisms use a multi-head attention mechanism. In the following, we provide a brief preliminary discussion on multi-head attention, self-attention and cross-attention.

\begin{figure*}[ht]
  \includegraphics[width=0.96\textwidth]{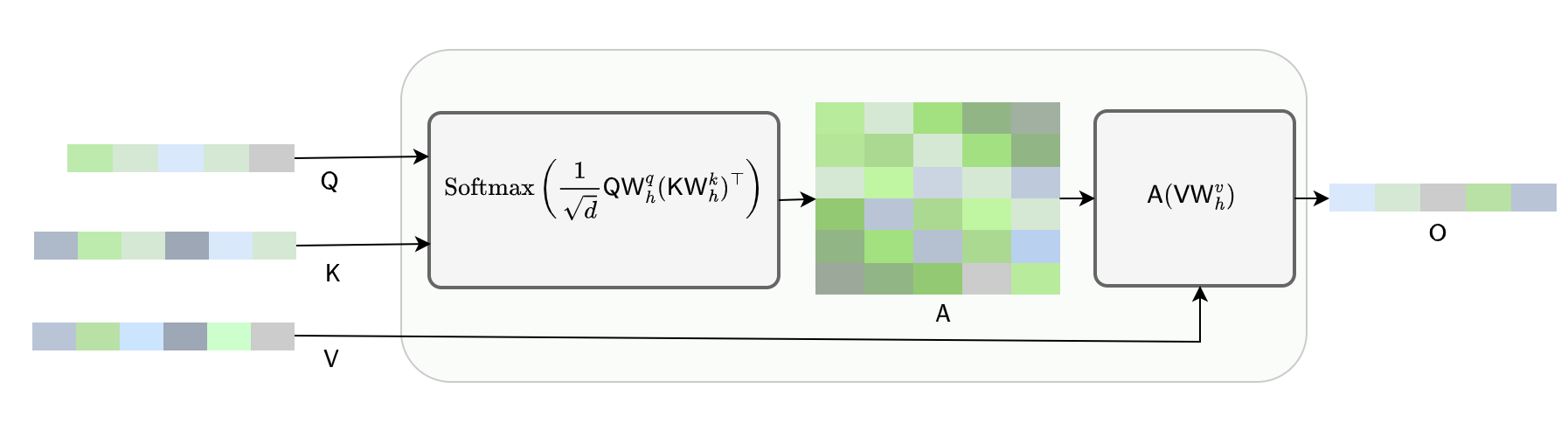}
  \caption[Single Attention Head Mechanism]{Single Attention Head Mechanism. Single attention head with query, $\mathsf{Q}$, key, $\mathsf{K}$, and value, $\mathsf{V}$. The $\mathsf{Q}$ and $\mathsf{K}$ (with their associated weights, $\mathsf{W}^q_h$, $\mathsf{W}^k_h$) are first used to produce attention affinity matrix, $\mathsf{A}$ and then the output, $\mathsf{O}$, is computed from the attention affinity matrix and value set. The colours in the set represent different tokens and the colours in the attention affinity matrix represent the different attention affinity scores between different pairs of tokens and do not correspond to any particular numerical values.}
  \label{fig:sha}
\end{figure*}

\begin{figure*}[ht]
  \includegraphics[width=0.96\textwidth]{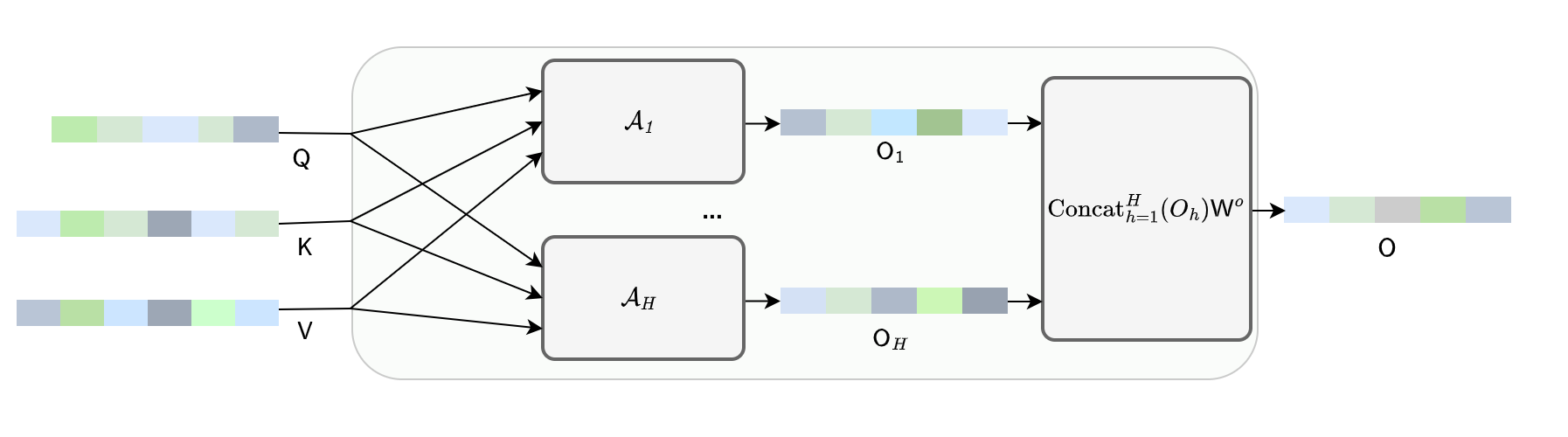}
  \caption[Multi-head Attention Mechanism]{Multi-head Attention Mechanism. Multi-head attention using $H$ attention heads is shown. The query, key and value are projected to each head, $\mathcal{A}_1, ..., \mathcal{A}_H$, and then outputs, $\mathsf{O}_1, ..., \mathsf{O}_H$, from these heads are concatenated and projected using output weights, $\mathsf{W^o}$, to produce final output, $\mathsf{O}$. The colours in the sets represent different tokens and do not correspond to any particular numerical values.}
  \label{fig:mha}
\end{figure*}

\begin{figure*}[ht]
  \includegraphics[width=0.96\textwidth]{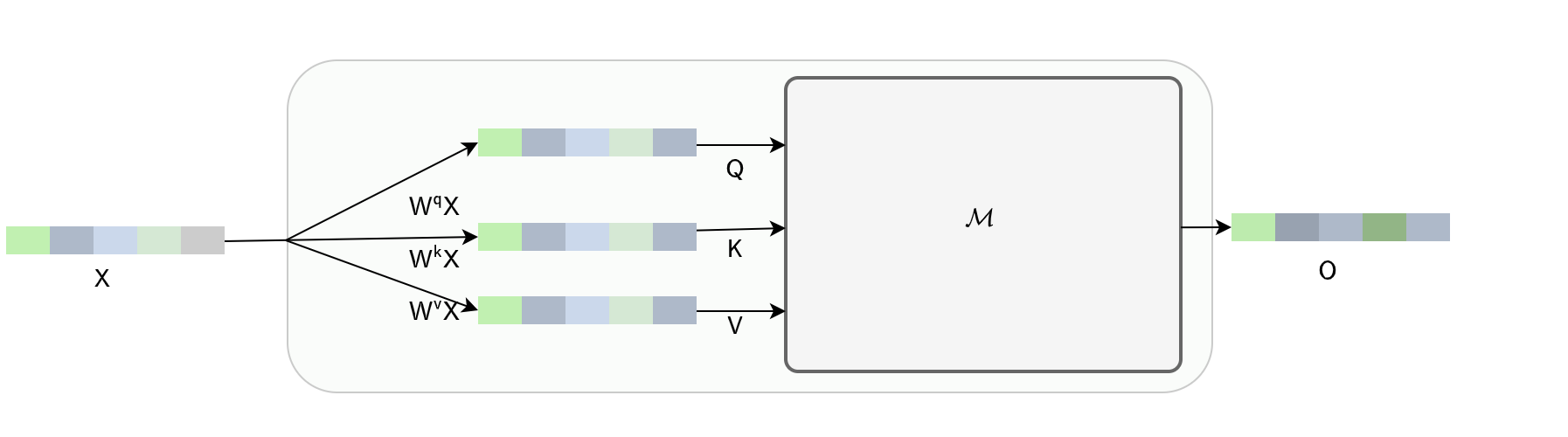}
  \caption[ Self-attention Mechanism]{Self-attention Mechanism. Self-attention mechanism explained in terms of multi-head attention, $\mathcal{M}$. The single input $X$ is projected to produce query, $\mathsf{Q}$, key, $\mathsf{K}$, and value, $\mathsf{V}$, sets for the multi-head attention to produce final output, $\mathsf{O}$. Position embeddings are not shown in favor of simplicity of visualization. The colours in the sets represent different tokens and do not correspond to any numerical values.}
  \label{fig:self-attn}
\end{figure*}

\begin{figure*}[ht]
  \includegraphics[width=0.96\textwidth]{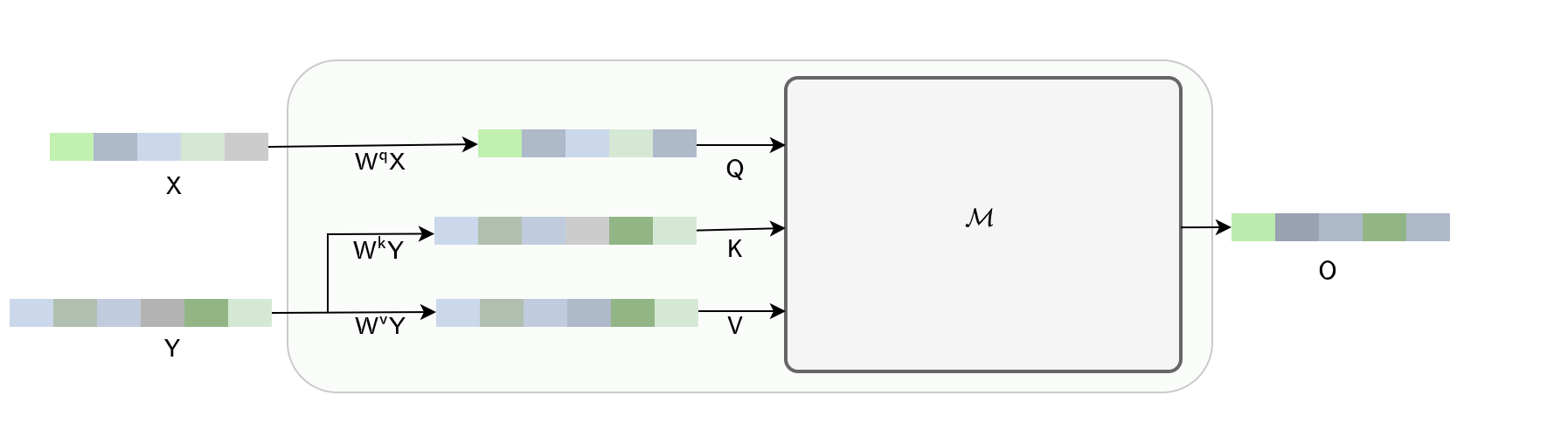}
  \caption[Cross-attention Mechanism]{Cross-attention Mechanism. Cross-attention mechanism explained in terms of multi-head attention, $\mathcal{M}$. One input, $\mathsf{X}$, is projected to produce query, $\mathsf{Q}$, while the other input, $\mathsf{Y}$, is projected to produce key, $\mathsf{K}$, and value, $\mathsf{V}$, sets for the multi-head attention to produce final output, $\mathsf{O}$. Position embeddings are not shown in favor of simplicity of visualization. The colours in the sets represent different tokens and do not correspond to any particular numerical values.}
  \label{fig:cross-attn}
\end{figure*}

\paragraph{Input format.}
Inputs to the attention function are represented as sets of tokens, also referred as vectors or sequences in some literature, especially in regards to natural language processing. A token is defined as a discrete unit of input with a numerical representation treated as a whole during processing. For example, a token can be a word, a sub-word or a character in language models and it can be a small spatial region or spatiotemporal volume in a vision transformer. Further details are discussed in \ref{sec:input_preprocessing}. For a single input, $X$, the set of $N$ tokens with $C$ features has the form $X^{N\times C}$. For batched data with $B$ elements in the batch, the token set has the form $X^{B\times N \times C}$.

\paragraph{Single-head attention.}
An attention function takes three sets of tokens as input and produces a set of tokens as its output, $\mathsf{O}$. The three inputs are called query, key and value. The terms query, key and value are used in analogy to their use in an information retrieval system, where data is retrieved based on an input ``query'', e.g., a text description, which is compared to a set of ``keys'', e.g., a set of video titles to produce ``values'', e.g., similarity scores, which are used to rank the videos with respect to the text. For transformers, the query, $\mathsf{Q}$, key, $\mathsf{K}$, and value, $\mathsf{V}$, are matrices whose rows are tokens. Single attention head, $\mathcal{A}_h$, then is defined in terms of query, key and value as 
\begin{equation}
   \hspace*{-0.2cm} \mathcal{A}_h(\mathsf{Q}, \mathsf{K}, \mathsf{V}) = \text{Softmax}\left(\frac{1}{\sqrt{d}}\mathsf{Q}\mathsf{W}^q_h(\mathsf{K}\mathsf{W}^k_h)^\top\right)\mathsf{V}\mathsf{W}^v_h,
    \label{eq:singlehead}
\end{equation} 
where $d$ is the feature dimension of $\mathsf{Q}$ and $\mathsf{K}$, while $\mathsf{W}^q_h, \mathsf{W}^k_h$ and $\mathsf{W}^v_h$ are the corresponding learned weight matrices for head $h$. Thus, $\mathsf{Q}\mathsf{W}^q_h(\mathsf{K}\mathsf{W}^k_h)^\top$ forms a matrix whose ${ij}^{th}$ element results from the inner product of row, $i$, of the (weighted) queries, $\mathsf{Q}$, and row, $j$, of the (weighted) keys, $\mathsf{K}$. Since by construction the rows of $\mathsf{Q}$ and $\mathsf{K}$ each correspond to a token, these operations systematically capture the relationships between the tokens, which subsequently are scaled and normalized by the token dimension, $d$, and the $\text{Softmax}$ to provide what is termed \textit{attention affinity}. In turn, the attention affinity is applied to the (weighted) values, $\mathsf{V}$, to capture its relationship and yield the final score. In short, an attention function first computes a global relationship score among all pairs of tokens in the input query and key sets in the form of attention affinity and then incorporates the tokens from the value set to emphasize this global relationship. These operations result in a representation that captures relevance or relationship among the tokens in the input sets. A visual depiction is shown in Figure~\ref{fig:sha}.

\paragraph{Multi-head attention.}
Multi-head attention uses multiple single-head attention mechanisms working in parallel and concatenates the outputs of all the heads. Thus, multi-head attention, $\mathcal{M}$, is defined  from~\ref{eq:singlehead} as
\begin{equation}
    \mathcal{M}(\mathsf{Q}, \mathsf{K}, \mathsf{V}) =
    \text{Concat}_{h=1}^{H} (\mathcal{A}_h(\mathsf{Q}, \mathsf{K}, \mathsf{V}))\mathsf{W}^o,
    \label{eq:multi-head}
\end{equation}
where $H$ is the number of attention heads in the multi-head attention and $\mathsf{W}^o$ is the weight matrix for the final multi-head output. By aggregating outputs from multiple attention head that operate on different projections of the inputs, the transformer model achieves increased representational capacity and enhanced generalization capability. As a result, the models can provide robust feature generation by capturing the diverse relationships among the tokens of input sets. A pictorial depiction of the operations involved is shown in Figure~\ref{fig:mha}.

\paragraph{Self-Attention.}
Self-attention, $\mathcal{S}$, is defined as an instance of attention with a single input, $\mathsf{X}$, and an optional positional embedding, $\mathsf{p}$. Commonly, it is instantiated in terms of multi-head attention~\eqref{eq:multi-head} as
\begin{equation}
    \mathcal{S}( \mathsf{X}, \mathsf{p})= \mathcal{M}(\mathsf{X}+\mathsf{p}, \mathsf{X}+\mathsf{p}, \mathsf{X}).
\end{equation} 
The positional embedding captures the position of tokens within a set to emulate that property of a sequence~\cite{dosovitskiy2020image}. It is used as an additional input to the attention mechanism to introduce locality, since attention itself is a permutation invariant operation. Self-attention relates different positions of a single set of tokens to compute a representation that captures global or long range context within the tokens of the input set. A visual depiction is shown in Figure~\ref{fig:self-attn}.

\paragraph{Cross-Attention}
Cross-attention, $\mathcal{C}$, is defined as an instance of attention with two inputs, $\mathsf{X,Y}$, and their corresponding positional embeddings, $\mathsf{p^X, p^Y}$. Commonly, it is instantiated in terms of multi-head attention~\eqref{eq:multi-head} as
\begin{equation}
    \mathcal{C}(\mathsf{X}, \mathsf{Y},
    \mathsf{p^X}, \mathsf{p}^{\mathsf{Y}})=\mathcal{M}(\mathsf{X} + \mathsf{p}^\mathsf{X},
    \mathsf{Y} + \mathsf{p}^\mathsf{Y},
    \mathsf{Y}).
\end{equation}
Hence, cross-attention is a four input operation used to mix or combine two different input sets and their positional embeddings into a single output. Here, one input serves as query and another input is used as both key and value. In this way, cross-attention allows a combination of information from two different modalities. A visual depiction is shown in Figure~\ref{fig:cross-attn}.

\section{Summary of related surveys}
Table~\ref{tab:related_surveys} provides a list of literature surveys that are complimentary to the current presentation.

\begin{table}[ht]
    \centering
    \begin{tabularx}{7in}{|X|c|X|}
        \toprule
        Title & Year & Remarks \\    
        \midrule
        Video Transformers: A Survey~\cite{selva2023video} & 2023 & transformers for video classification \\ \hline 
        Vision Transformers for Action Recognition: A Survey~\cite{ulhaq2022vision} & 2022 & transformers for action recognition\\ \hline
        A Survey on Deep Learning Technique for Video Segmentation~\cite{wang2021survey} & 2021 & approaches on video objects and video semantic segmentation\\ \hline
        Multi-task learning for dense prediction tasks: A survey~\cite{vandenhende2021multi} & 2021 & multitask learning for segmentation \\ \hline
        A Survey on Neural Network Interpretability~\cite{zhang2021interpretability} & 2021 &neural network interpretability \\\hline 
        Efficient Transformers: A Survey~\cite{tay2022efficient} & 2020 & transformers efficiency, e.g., factorized attention etc. \\ \hline 
        Video object segmentation and tracking: A survey~\cite{yao2020video} & 2020 & segmentation and tracking \\ \hline 
      \end{tabularx}
    \caption{Summary of literature surveys.}
    \label{tab:related_surveys}
\end{table}

\section{Outline}
This chapter has served to motivate and provide a broad overview of video segmentation, classical approaches and transformers. Chapter~\ref{CH2} will provide a detailed discussion of different categories of video segmentation and benchmark datasets. Chapter~\ref{CH3} will provide a component-wise review of application of transformers to video segmentation. That chapter also will present the level of performance achieved by these models on the video segmentation tasks itemized in Chapter~\ref{CH2}. Chapter~\ref{CH4} will provide a review of interpretability analysis of transformer models, with a special focus on video segmentation. Finally, Chapter \ref{CH5} will provide conclusions, including suggestions for future research directions.

\chapter{Video Segmentation}\label{CH2}
\section{Overview}\label{task_overview}
\begin{figure}[ht]
    \centering
    \begin{minipage}{.95\textwidth}
        \centering
        \includegraphics[width=1.0\textwidth]{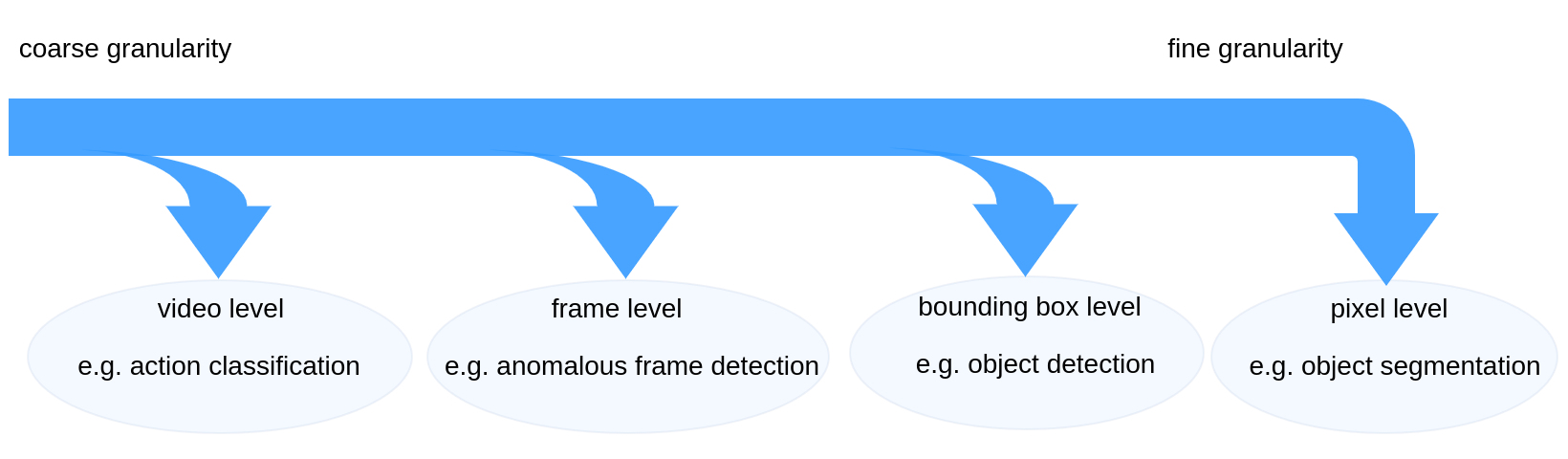}
    \end{minipage}
\caption[Video Understanding Tasks]{Video Understanding Tasks by Granularity.}
\label{fig:task_granularity}
\end{figure}

The primary goal of computer vision for an intelligent autonomous system is the perception or understanding of the surrounding world from captured images, both single images as well as multiple images acquired across time and/or space. The inferred knowledge can be used by other subsystems. These uses might include future prediction and path planning, among many others. Depending on the granularity of the perception, video understanding tasks can be broadly categorized into video level, frame level and pixel-level inference as shown in Figure ~\ref{fig:task_granularity}. Pixel level inference also can be of two levels of granularity: bounding box and per-pixel inference. Per-pixel inference is sometimes referred to as dense estimation or dense prediction. In this survey, our focus is on the family of video understanding tasks operating at pixel-level inference. 

Pixel level video understanding tasks are among the fundamental problems in computer vision with a wide range of applications, including robotics, autonomous driving, video editing and human-computer interaction. The task of pixel-level inference is to group pixels in images or video frames based on a given qualitative or quantitative criteria, such as an object, action, text reference, depth or motion. While all pixel level video understanding tasks work on the pixel level, there is a large variety of problem formulations. The diversity comes from the different possibilities of output space and objectives, such as motion, semantics, instances, action and activity of the objects. Video inference is again of two types: offline and online. In the offline setting, all the frames are available at once, while only the past frames are available to the model in online inference. In addition to task specific challenges, video segmentation tasks share some common challenges, \eg background motion, camera motion, object deformation and clutter.

Video segmentation, as a particular type of pixel-wise inference, is the focus of this chapter. In the following sections, we will present a brief taxonomy of the different types of video segmentation tasks and summarize the most recently used datasets in these tasks. More historical details of the datasets and approaches are available elsewhere~\cite{yao2020video, wang2021survey}. We organize this chapter as follows: Section ~\ref{sec:objseg} discusses object segmentation in videos; Section ~\ref{sec:actorseg} discusses actor-action segmentation; Section ~\ref{sec:sceneseg} discusses holistic scene segmentation; Section ~\ref{sec:mmseg} discusses multimodal video object segmentation.

\section{Objects}\label{sec:objseg}
Object-based video segmentation aims to map groups of pixels in video frames to object masks based on qualitative criteria regarding object properties in the videos. Object-based segmentation approaches are mainly of two types: Video Object Segmentation (VOS), which only segments foreground objects and Video Instance Segmentation (VIS), which segments all object instances along with tracking across frames. Depending on the required level of prior information or active human interaction, video object segmentation can be of three types: automatic VOS, semi-automatic VOS and interactive VOS. In the following, we discuss the problem formulations and currently popular datasets for each of these video object segmentation tasks.

\subsection{Automatic video object segmentation}
Automatic Video Object Segmentation (AVOS), often called primary object segmentation, is defined as grouping pixels into foreground object masks representing the primary object in a video sequence, where primary is distinguished as the object showing dominant motion or distinguishing appearances across the frames~\cite{yao2020video,wang2021survey}. For this reason, AVOS often has been termed foreground/background segmentation since an object with distinctive motion and/or appearance properties in a video sequence is generally the foreground object. Since no prior information is available, this kind of segmentation also is called unsupervised VOS or zero-shot VOS. In addition to the challenges common to most video segmentation task that were listed above, AVOS also faces the challenges of lack of prior information regarding the objects of interest.

\subsection{Semi-automatic video object segmentation}
Semi-automatic Video Object Segmentation (SVOS) is a mask-guided object segmentation where one or more annotated frames (commonly the first frame) are available as reference masks containing one or more objects as a target. The objective is to segment the target objects throughout the video. This task also is called semi-supervised VOS or one-shot VOS due to the availability of guidance or supervision. SVOS has its own challenges of mask propagation for target objects in video containing multiple objects, often from the same semantic category. Recently, a multimodal extension to this type of video object segmentation has become popular in the computer vision community where the reference is text or audio. We present a discussion of that variation in Section~\ref{sec:mmseg}.

\subsection{Interactive video object segmentation}
Interactive Video Object Segmentation (IVOS) is semi-automatic object segmentation where there is an interaction between the system and a human providing hints to the object to be segmented (e.g., scribbles)~\cite{nagaraja2015video,bai2009video, criminisi2010geodesic, zhong2012discontinuity,fan2015jumpcut}. This interaction happens in a loop with a series of interactions beginning with initial guidance and followed by feedback.  Other forms of interaction include contours~\cite{lu2016coherent} and points~\cite{wang2017selective}.  IVOS allows for applications requiring high-quality segmentation at the trade-off of human supervision. This approach also is used to generate datasets for automatic or other semi-automatic video segmentation tasks. IVOS faces task specific challenges from the complexity of user interaction (\eg target is given as a scribble, unlike a mask in semi-automatic VOS).

\subsection{Video instance segmentation}
Video instance segmentation (VIS) aims at the simultaneous detection, segmentation and tracking of instances in videos~\cite{yang2019video}. The problem also is referred to as Multi-Object Tracking and Segmentation (MOTS)~\cite{voigtlaender2019mots}. The related task, video object detection and tracking, aims at detecting (bounding boxes) and tracking objects, while video instance segmentation additionally generates pixel-wise object segmentation masks. Again, in contrast to instance segmentation in images, which involves detection and segmentation~\cite{hariharan2014simultaneous}, the requirement of tracking object instance masks across video frames makes it more challenging.  In short, VIS simultaneously predicts object bounding boxes and object masks where the bounding boxes are labeled with semantic categories and segmentation masks are based on object identities across frames. The tracking of object masks makes it especially relevant for the application of video editing, autonomous driving and augmented reality~\cite{yang2019video}.

\subsection{Benchmark datasets}
We now present a brief discussion of the large-scale video object segmentation benchmark datasets that are commonly used in recent literature. 

\paragraph{YouTube-Objects.} The YouTube-Objects~\cite{prest2012learning} dataset 
contains a total of 155 web videos collected from YouTube. There are objects of 10 semantic classes and each class has nine to 24 videos of duration between 30 seconds and three minutes at resolution 720p. These videos are split into 1,407 clips containing a total of approximately 570,000 frames. 

For video object segmentation, a subset of this dataset is used which has 126 shots with 20,647 frames covering all of the 10 semantic classes. A shot in this context is a short clip from a video containing the same objects. Generally, a shot is a continuous sequence from a video that was produced from a single capture of the camera and hence contains mostly the same objects with uniform camera settings. This subset provides pixel-level foreground-background annotations of every $10^{th}$ frame~\cite{jain2014supervoxel}. This subset is used to validate the generalization ability of automatic video object segmentation models and has no training set of its own. For this purpose, models trained on other datasets are validated on the pixel-level annotated subset of the YouTube-Objects dataset. For evaluation, a measure of region similarity between the estimate and ground truth is used. In particular, the region similarity is defined as the ratio of the intersection area to the union area of the prediction and ground truth. For this reason, this measure is often termed Intersection over Union (IoU), but also is known as the Jaccard index~\cite{shi2014jaccard}. This dataset initially computes the mean of IoU over all clips in each category and then over all categories, which is referred as mean IoU or mIoU.   


\paragraph{DAVIS 2016.}
The Pixel level Annotated VIdeo Segmentation (DAVIS) 2016~\cite{perazzi2016benchmark} dataset is an automatic foreground-background segmentation dataset containing 50 high-quality video sequences with a total of 3,455 annotated frames. There are 30 videos for training and 20 for testing. The videos cover multiple occurrences of common video object segmentation challenges, such as occlusions, motion blur and appearance changes. The video sequences contain a single object or two spatially connected objects annotated in each frame as the primary  foreground object with a distinctive motion. There also are video-level attributes describing challenges present in the sequence, such as occlusions, fast-motion, non-linear deformation and motion-blur to enable a deeper understanding of the results of algorithms on this dataset.

The DAVIS 2016 dataset proposed three complementary metrics to evaluate the performance of video object segmentation models. These metrics measure the spatial accuracy of the segmentation, the quality of the silhouette and the temporal coherence. Spatial accuracy is measured using region similarity (intersection-over-union or Jaccard index), like other VOS datasets. Silhouette quality is measured in terms of contour accuracy. To compute contour accuracy, the segmentation mask is considered as a set of closed contours, then the F-measure~\cite{chen2004statistical} is calculated from the contour based precision and recall between the segmentation mask and the ground truth. Temporal stability is measured in terms of the segmentation contours between frames. For this purpose, polygonal representations of the contours of adjacent frames are matched using dynamic time warping to minimize the shape context descriptor (SCD) distance~\cite{belongie2002shape}. The temporal stability is calculated as the mean cost per matched point.

\paragraph{DAVIS 2017.} The DAVIS 2017~\cite{pont20172017,caelles20182018} dataset provides annotations for automatic video object segmentation, semi-supervised video object segmentation and interactive video object segmentation. The DAVIS 2016 dataset was extended by adding more videos, annotations for multiple objects instances and semi-automatic video object segmentation (semi-supervised and interactive). There are 150 video sequences in total in this dataset. These are divided into train, validation, test-development and test-challenge splits with 60, 30, 30 and 30 videos, resp. The total number of annotated frames is 10,549 containing 376 objects. In addition to the distinctive motion of DAVIS 2016, the granularity of instances are defined by their semantics. As an example, people and animals are together segmented as a single foreground instance in DAVIS 2016. And in DAVIS 2017 they are two foreground object instances. Compared to DAVIS 2016, DAVIS 2017 contains additional challenges arising from more distractors, smaller objects, finer structures, more occlusions and faster motions.

To benchmark approaches for unsupervised video object segmentation, the metrics from DAVIS 2016 are adopted. To evaluate algorithms for the semi-supervised video object segmentation task, this dataset adopted the region similarity (IoU) and boundary accuracy (F-measure) measures from the DAVIS 2016 dataset. Temporal stability is no longer considered. The benchmark is to compute each metric per each object instance on each clip and take the average over all instances of the entire test set. The overall performance metric for ranking in the DAVIS 2017 challenge is defined as the average of the two metrics~\cite{pont20172017}.

To evaluate interactive video object segmentation, a test server is used to simulate human inputs and participants need to connect to a server to receive the scribbles. There are a maximum of eight interactions to iteratively refine the interactive segmentation result. To benchmark the algorithms, first the region similarity~\cite{caelles20182018} or the mean of region similarity and contour accuracy~\cite{caelles20192019} is plotted against the interaction time and then the Area Under the Curve (AUC) of the plot is computed.

\paragraph{MoCA.} Moving Camouflaged Animals (MoCA) is an especially challenging video object segmentation dataset containing camouflaged animals~\cite{lamdouar2020betrayed}. Notably, in many cases the camouflage is so strong that the target animal is indistinguishable from the background in a single frame; however, it is revealed in the motion exhibited across frames. The dataset contains 141 video sequences collected from YouTube depicting various camouflaged animals of 67 categories moving in natural scenes. Similar to YouTube Objects dataset, models trained on another dataset are tested on this dataset to validate generalization ability. The annotation provides a bounding box in each frame, as well as a motion label. The motion level can be one of locomotion, deformation or static. 

The evaluation protocol for segmentation on this dataset computes IoU between the ground truth box and the minimum box that includes the estimated segmentation mask. In addition, the evaluation protocol on MoCA includes success rate at various thresholds. The evaluation in this dataset includes only the segmentation of the animals under locomotion or deformation and excludes static frames.

\paragraph{YouTube-VOS.} 
The YouTube Video Object Segmentation dataset (YouTube-VOS)~\cite{xu2018youtube} is a large-scale benchmark for video object segmentation. The dataset contains 3,252 YouTube video clips. These are divided into 2,796 videos for training, 134 videos for validation and 322 videos for testing. The test set is further split into test-seen and test-unseen subsets to evaluate the generalization ability of segmentation models on unseen categories. They provide annotations for 94 object categories including common objects and human activities. To benchmark on this dataset, region similarity and contour accuracy are adopted, following DAVIS 2016~\cite{perazzi2016benchmark}.

\paragraph{YouTube-VIS.} YouTube-VIS~\cite{yang2019video,vis2021} is a video instance segmentation dataset with a subset of sequences from YouTube-VOS and instance-level annotation. The dataset contains 2,883 high resolution YouTube videos. This dataset provides annotations for 131,000 high-quality instance masks covering 4,883 unique objects from 40 object categories selected from the 94 categories of YouTube-VOS~\cite{xu2018youtube}.

The benchmark metrics employed are Average Precision (AP) and Average Recall (AR), used in image segmentation with the extension of a different region similarity (IoU) measure to account for tracking~\cite{yang2019video}. These metrics are first evaluated per category and then averaged over the category set. The AP is defined as the area under the precision-recall curve of the confidence score. Following the COCO benchmark~\cite{lin2014microsoft}, AP is averaged over multiple intersection-over-union (IoU) thresholds from $50\%$ to $95\%$ with steps of $5\%$. The maximum recall at some given fixed number of segmented instances per video is used as AR. 

\paragraph{UVO.} Unidentified Video Objects (UVO) is a nontrivially large class-agnostic object segmentation dataset that shifts the problem to the open-world setting with an extremely large number of object annotations~\cite{wang2021unidentified}. The open-world setting is aimed at detecting, segmenting and tracking all objects exhaustively in a video, which may contain unfamiliar objects. The dataset provides 5,641 videos for training, 2,708 videos for validation and 2,879 videos for testing. The videos are adopted from the Kinetics dataset, a popular action recognition benchmark~\cite{carreira2017quo}. To evaluate video instance segmentation, this dataset used Average Precision (AP) at different Intersection-over-Union (IoU) thresholds and Average Recall (AR), similar to YouTube-VIS~\cite{yang2019video}.

\paragraph{OVIS.} Occluded Video Instance Segmentation (OVIS)~\cite{qi2022occluded} is a large-scale video instance segmentation dataset consisting of very challenging scenarios of objects under various levels of occlusion. There are 901 videos with conditions of severe object occlusions covering 5,223 unique instances from 25 semantic categories.

To evaluate video instance segmentation, this dataset also used Average Precision (AP) at different Intersection-over-Union (IoU) thresholds and Average Recall (AR), similar to YouTube-VIS~\cite{yang2019video}. In addition, to evaluate algorithms at various levels of occlusion, AP is calculated under different occlusion degrees. To facilitate this evaluation, occlusion level information is included for each object in the annotation.

\paragraph{MOSE.} coMplex video Object SEgmentation (MOSE)~\cite{ding2023mose} is a large scale video object segmentation dataset designed for semi-supervised VOS under crowded and occluded scenarios. The dataset contains 2,149 video clips and 5,200 objects from 36 categories. The target objects in most of the videos typically are occluded by other objects and disappear in some frames. Region similarity (IoU) and contour accuracy (F-measure) are used for evaluation on this dataset, similar to other evaluations~\cite{perazzi2016benchmark, pont20172017}.

\paragraph{VOST.} Video Object Segmentation under Transformations (VOST)~\cite{tokmakov2023breaking} is a benchmark for semi-supervised video object segmentation, emphasizing intricate object transformations. In contrast to existing datasets, VOST includes objects that are fractured, broken, torn and reshaped, leading to nontrivial changes in their overall appearance. A meticulous, multi-step approach is employed to ensure that these videos capture complex transformations comprehensively throughout their temporal duration. The dataset comprises over 700 high-resolution videos, captured in diverse settings, with an average duration of 20 seconds, and is densely labelled with instance masks. There are 51 different types of transformations and 154 object categories in the dataset. There are a total of 713 videos split into 572 for training, 70 for validation and 71 for testing. Two evaluation metrics are used in this dataset, IoU after
transformation and the overall IoU.

\paragraph{TCOW.} Tracking through Containers and Occluders in the Wild (TCOW)~\cite{van2023tracking} is a semi-supervised video object segmentation dataset under challenging multi-object scenarios focused on segmenting target objects as well as marking the containers and occluders when the object goes inside a container or becomes occluded by another object. The dataset includes both synthetic and real world data with masks annotated for object, container of the object and occluder of the object. The evaluation metrics uses mean IoU for target object, container and occluder masks separately.

\section{Actors and actions}\label{sec:actorseg}
\subsection{Actor-action segmentation}
Actor-action segmentation is defined as delineating regions from a given video with the corresponding label of an actor-action tuple~\cite{xu2015can}. Actors are objects that are participants in some actions. This task has been investigated far less intensively than object segmentation. Due to the exclusiveness of certain actor-action tuples, jointly segmenting actor-action has been found to result in better modeling than independently segmenting them in the context of action recognition. As an example, a car can not eat and a cat can not fly. The exclusiveness of these actor-action tuples makes it less likely for the model to learn to predict such tuples. Conversely, independent inference of actor and action can predict car as actor and climb as action simultaneously for a particular segment, even though the inference of actor-action tuple car-climb is unrealistic. This exclusiveness of actor-action tuples is evident from the quantitative experiments presented in~\cite{xu2015can}. 

\subsection{Benchmark datasets}
\paragraph{A2D.} The A2D~\cite{xu2015can} dataset provides actor-action segmentation with annotation for seven actor classes (adult, baby, ball, bird, car, cat and dog), eight action classes (climb, crawl, eat, fly, jump, roll, run and walk) and the no-action class. This dataset facilitates three specific problems including actor-action recognition with single-label, multiple-label and  actor-action segmentation. The A2D dataset has 3,782 unconstrained “in-the-wild” videos with varying characteristics collected from YouTube and provides pixel-level annotations. There are 43 valid actor-action tuples with at least 99 instances per tuple for joint actor-action segmentation tasks. The training split has 3,036 videos and test split has 746 videos divided evenly over all actor-action tuples. Both pixel level and bounding box annotations are provided per actor-action pair.

The evaluation of  actor-action segmentation on this dataset includes global pixel accuracy, mean per-class accuracy and mean pixel Intersection-over-Union (mIoU) for both separate actor, action segmentation and joint actor-action segmentation, following prior work~\cite{xu2016actor}. The global pixel accuracy measures the overall percent of pixels with correct label. The mean per-class accuracy measures the percent of pixel level accuracy initially per class and then averaged across classes. The mean pixel Intersection-over-Union (mIoU) evaluates the region similarity using Intersection over Union (IoU), aggregated similarly to per-class accuracy. Recent work considers mIoU as the most representative metric for correct pixel prediction over all classes~\cite{rana2021we}.

\section{Scenes}\label{sec:sceneseg}
Scene segmentation aims to segment all the pixels in a scene with labels that assign them to a particular category or instance of a category. In this context, scene categories are defined either as ``thing'' or ``stuff''~\cite{lin2014microsoft}. In general, ``things'' are countable objects that typically have a well defined shape, whereas ``stuff'' is an uncountable entity with shape typically not being a defining attribute. For example: People and cars are countable objects, with shape being an important attribute; in contrast, sky and forest are not countable and shape is not important in their definition. Scene segmentation also is referred to as scene parsing in some contexts, since it aims to parse all the contents of a scene according to categories or instances. Both video semantic segmentation and video panoptic segmentation aim to label all the pixels in a scene, including both things and stuff. The difference is that semantic segmentation does not consider instances, while panoptic segmentation does.

\subsection{Video semantic segmentation}
Video Semantic Segmentation (VSS), also referred to as Video Scene Parsing (VSP)~\cite{yan2020video}, aims to label all the pixels in a scene according to semantic categories of thing and stuff classes in an instance agnostic fashion~\cite{wang2021survey}. Similar to VOS, VSS also can be automatic or semi-automatic. Automatic VSS segments all the frames without using any key-frame, while semi-automatic VSS uses annotated key-frames to segment other frames. Due to the ability of VSS to describe the high-level semantics of the physical environment, it can serve as upstream perception for many applications of autonomous systems, e.g., autonomous driving, robot navigation and human-machine interaction.

\subsection{Video panoptic segmentation}
Panoptic segmentation of images was first introduced as a holistic segmentation of all foreground instances and background regions in a scene~\cite{kirillov2019panoptic}, which was later extended to video~\cite{kim2020video}. Video Panoptic Segmentation (VPS) unifies video instance segmentation for thing classes and video semantic segmentation for stuff classes~\cite{kim2020video}. The difference between VPS with VSS is that in VSS things are segmented into semantic categories without tracking instances. Here, it also is worth noting that the difference between VPS and VIS is that the latter considers all stuff classes simply as background, while VPS distinguishes different classes of stuff. So, this is an instance-aware semantic segmentation with tracking of object IDs. Overall, VPS jointly provides estimates of object classes, bounding boxes, masks, instance ID tracking and semantic segmentation in video frames.

\subsection{Depth-aware video panoptic segmentation} 
Depth-aware Video Panoptic Segmentation (DVPS) is defined as an inverse projection problem that restores the point clouds from a video and simultaneously provides each point with instance-level semantic interpretations~\cite{qiao2021vip}. This task also can be considered as an extension to VPS that aims to jointly provide estimates of panoptic segmentation and depth in a video.

\subsection{Panoramic video panoptic segmentation} 
Panoramic Video Panoptic Segmentation (PVPS) extends VPS to panoramic videos generated simultaneously from multiple cameras. In addition to the multi-view setting, this task has another added challenge of tracking instance IDs across multiple cameras at each time. PVPS is an important requirement for autonomous driving, where there are multiple cameras on the vehicle to capture videos of the surrounding environment.

\subsection{Benchmark datasets}

\paragraph{VSPW.} Video Scene Parsing in the Wild (VSPW)~\cite{miao2021vspw} is a large-scale video semantic segmentation dataset containing 3,536 videos, each approximately five seconds long. The dataset provides pixel-level annotations for 124 categories. There are 2,806 videos in the training split, 343 videos in the validation split and 387 videos in the test split. The dataset covers a wide range of real-world scenarios, including both indoor and outdoor environments. Annotations are provided for both thing and stuff classes. 

The evaluation metrics for this dataset includes mean IoU, weighted IoU and Temporal Consistency (TC). The mean IoU is the region similarity measure using intersection-over-union between the estimated and ground truth pixels averaged over all the classes. The weighted IoU also calculates the region similarity but is weighted by the total pixel ratio of each class. The Temporal Consistency (TC) is computed as the intersection-over-union between the estimated mask of frame t and the warped estimated mask from frame t-1 to frame t.  Optical flow-based warping is used for this purpose.

\paragraph{VIPSeg.} VIdeo Panoptic Segmentation
in the wild (VIPSeg)~\cite{miao2022large} is a large-scale video panoptic segmentation dataset that aims to assign semantic classes and track identities for all pixels in a video. The dataset contains 3,536 videos. There are a total of 84,750 pixel-wise annotated frames that only focus on street view scenes. There are total of 232 scenarios with 124 categories, including 58 thing classes and 66 stuff classes.

The evaluation metrics for this dataset are Video Panoptic Quality (VPQ) as well as Segmentation and Tracking Quality (STQ). VPQ is extended for videos from Panoptic Quality (PQ) used in panoptic image segmentation. VPQ computes the average quality  using tube IoU matching across a small span of frames. STQ is the geometric mean of Association Quality (AQ) and Segmentation Quality (SQ) that evaluates instance agnostic segmentation quality using semantic classes. Here, AQ is a measure of the pixel-level association across an entire video and SQ is a measure of semantic segmentation quality by class-level IoU.

\paragraph{VIPER.} The VIPER dataset~\cite{richter2017playing} is a synthetic dataset collected from the GTA-V video game~\cite{richter2016playing} and supports video semantic segmentation, video instance segmentation and video panoptic segmentation. The dataset was originally proposed for video semantic segmentation and video instance segmentation. Subsequently, it was extended by adding COCO-style metadata to support video panoptic segmentation~\cite{kim2020video}. The dataset contains ego-centric scenes of driving, riding and walking a total of 184 kilometers in diverse ambient conditions in a realistic virtual world. There are 254,064 annotated frames with 10 thing classes and 13 stuff classes. The evaluation metric for this dataset is Video Panoptic Quality (VPQ), as with VIPSeg~\cite{miao2022large}. 


\paragraph{Cityscapes-VPS.} The Cityscapes-VPS~\cite{kim2020video} dataset is developed for video panoptic segmentation by extending the annotations in the popular image panoptic segmentation dataset Cityscapes~\cite{cordts2016cityscapes}. Provided are pixel-level panoptic annotations for 3,000 frames with 19 classes and temporally consistent instance IDs for objects within videos. Cityscapes-VPS also uses the video panoptic quality (VPQ) metric as an evaluation measure.

\paragraph{KITTI-STEP.} KITTI-STEP is a challenging video panoptic segmentation dataset containing long videos recorded by a camera mounted on a driving car~\cite{weber2021step}. This dataset provides 12 videos for training, 9 videos for validation and 29 videos for testing. The videos contain challenging scenarios of regularly (re-)appearing and disappearing objects, occlusions, lighting condition changes and scenes of pedestrian crowds. This dataset adopts evaluation metrics similar to VIPSeg~\cite{miao2022large}, Video Panoptic Quality (VPQ) and Segmentation and Tracking Quality (STQ). 

 \paragraph{Cityscapes-DVPS.} ViP-DeepLab~\cite{qiao2021vip} designed the Cityscapes-DVPS dataset by adding depth annotation to Cityscapes-VPS~\cite{kim2020video}. A new evaluation metric for DVPS, Depth aware Video Panoptic Quality (DVPQ), is introduced for this dataset extending VPQ by additionally considering depth estimation. Two parameters, $k$ and $\lambda$, are used with DVPQ to define $DVPQ^k_\lambda$ for this purpose. Here, $\lambda$ refers to the threshold of relative depth error used for calculating VPQ, while $k$ refers to the number of frames in the video clip used to calculate the VPQ. $DVPQ^k_\lambda$ is calculated for thing and stuff classes separately and jointly.

\paragraph{SemKITTI-DVPS.} SemKITTI-DVPS~\cite{qiao2021vip} extends the SemanticKITTI dataset~\cite{behley2019semantickitti} by adding depth annotation. The dataset has a total of 22 sequences of which 11 are for training and 11 are for testing. One of the training sequences also is used for validation. The original dataset has perspective images and panoptic-labeled 3D point clouds with annotation for 8 thing and 11 stuff  classes. The dataset is converted to DVPS by projecting the 3D point clouds into the image plane followed by some additional post-processing to remove sampled points that exhibit large relative errors due to sensor alignment issues. Similar to Cityscapes-DVPS, $DVPQ^k_\lambda$ is used as the evaluation metric for the SemanticKITTI-DVPS dataset.

\paragraph{WOD-PVPS.} The Waymo Open Dataset for Panoramic Video Panoptic Segmentation 
(WOD-PVPS)~\cite{mei2022waymo} is a very large scale panoramic video panoptic segmentation dataset of driving scenarios. The dataset contains 2,860 temporal sequences of driving scenarios in three different geographical locations, as captured by five cameras mounted on autonomous vehicles. The provided annotations are consistent over time and across multiple  camera views to facilitate full panoramic scene understanding. Overall, there are total of 100,000 labeled camera images with 28 semantic categories. 

The PVPS evaluation metric for this dataset is weighted Segmentation and Tracking Quality (wSTQ), adopted from the Segmentation and Tracking Quality (STQ) metric for VPS. In particular, each pixel is weighted by the camera coverage fraction before calculating association quality and segmentation quality. Camera coverage fraction for a pixel is calculated as the ratio of the number of cameras where that pixel is present to the total number of cameras.

\section{Multimodal video segmentation}\label{sec:mmseg}
Multimodal Video Object Segmentation, often termed Referring Expression Object Segmentation, uses input from another modality (e.g., text or audio) to indicate the segmentation target. This formulation is similar to semi-supervised VOS, where reference is given as a mask annotation of the first frame. In contrast, in multimodal video segmentation, reference to the object is given by text or audio. 

\subsection{Text-guided video object segmentation}
Text-Guided video object segmentation is defined as automatically separating foreground objects from the background in a video sequence based on referring text expressions. Since text is the most common reference source for video object segmentation, it simply is called Referring Video Object Segmentation (RVOS). There are two types of annotations used for text-guided video object segmentation. The first is based on reference to an object description based on the first frame, which is very similar to semi-supervised VOS except the reference is text. This kind of reference text mostly describes the object appearance, e.g., `A man wearing a white cap'. The second type of referring text expression is based on the full video clip and also includes temporal dynamics of the object in the referring expression, e.g., `A person showing his skateboard skills on the road'.

\subsection{Audio-guided video object segmentation}
Audio-Guided Video Object Segmentation (AGVOS) is defined as automatically separating foreground objects from the background in a video sequence based on referring audio expressions comprised of spoken descriptions of target objects~\cite{pan2022wnet}. The authors of ~\cite{pan2022wnet} refer to this task as AVOS, while we use AGVOS to avoid confusion with automatic VOS. The motivation for audio-guided VOS is that audio often contains additional information compared to text, e.g., the accent, emotion and speed.

In addition to segmentation based on speech-based referring audio expressions, an audio guided formulation known as Audio-Visual Segmentation (AVS)~\cite{zhou2022audio} aims to segment objects as sound sources in the video. In contrast to the audio referring or audio guided video segmentation where the audio comes from a speaker asking or describing a target object, the audio of this setting comes from the scene captured in the video. The problem is to segment the objects which are producing the sounds.

\subsection{Benchmark datasets}
\paragraph{A2D Sentence~\cite{gavrilyuk2018actor}.} This dataset is designed for pixel-level segmentation of actors and their actions in videos from a natural language query sentence. This dataset extended actor and action dataset A2D~\cite{xu2015can} with natural language descriptions. The natural language description contains information about what each actor is doing in the videos. An example of a referring expression is `red ball is rolling on a bowling floor'. There are 3,036 training videos and 746 testing videos.

The standard evaluation protocol in A2D Sentence includes overall IoU, mean IoU,  precision at 5 different intersections from 0.5 to 0.9 and the mean average precision at intersection over thresholds from 0.50 to 0.95 at interval of 0.05. The overall IoU is the ratio of total intersection area of all test data and the total union area. Since this measure of overall IoU over the whole test data favors large segmented regions, the mean IoU (mIoU) is used as an additional metric to treat large and small regions similarly. The precision at threshold K also called Precision@K is the percentage of test samples with IoU above the threshold K.

\paragraph{J-HMDB Sentence~\cite{gavrilyuk2018actor}.} This dataset is designed by extending the J-HMDB action recognition dataset~\cite{jhuang2013towards} with natural language expressions describing various actions. There are 928 video clips of 21 different actions in the original J-HMDB dataset. The annotations are available with a 2D articulated human puppet that provides scale, pose, segmentation and a coarse viewpoint for the humans involved in each action. The added natural language description contains information about what the target object is doing in each video. An example text expression is `boy in gray shirt and black shorts swinging baseball bat'. The evaluation protocol for this dataset is the same as for A2D Sentence~\cite{gavrilyuk2018actor}.

\paragraph{Ref-DAVIS.}  Referring expressions for DAVIS 2016 and 2017 (Ref-DAVIS)~\cite{khoreva2019video} is an extension of the DAVIS 2016 and DAVIS 2017 datasets by adding referring text expressions. The referring expression is generated to describe the object in the first frame mask of the semi-supervised setting. All objects annotated with mask labels in DAVIS 2016 and DAVIS 2017 are augmented with non-ambiguous referring expressions. Overall, the augmented dataset contains over a thousand referring expressions for more than 400 objects on 150 videos with approximately 10,000 frames. The evaluation metrics for this dataset are region similarity, contour accuracy and their average value.

\paragraph{Ref-YouTube-VOS.} Refer-YouTube-VOS, also known as RVOS~\cite{seo2020urvos}, is a large scale referring video object segmentation dataset with referring expressions added to the YouTube-VOS~\cite{xu2018youtube} dataset. The dataset provided two kinds of annotations to describe the highlighted object: full-video expression and  first-frame expression. The evaluation metric for this dataset is the same as for Ref-DAVIS, i.e., region similarity, contour accuracy and their average.

\paragraph{AGVOS.} Audio Guided VOS (AGVOS)~\cite{pan2022wnet} is a large-scale audio-guided VOS dataset developed by extending the Ref-YouTube-VOS~\cite{seo2020urvos}, A2D~\cite{xu2015can} and J-HMDB~\cite{jhuang2013towards} datasets. Similar to the author's use of confusing terminology for this task (see above), they also use the acronym AVOS for the dataset. To generate the audio, 36 speakers read descriptive sentences. The sampling rate is 44,100 kHz or above, the sampling number is 16 bits and the speaking speed is 100-150 words per minute. The average length of each recording is 5 to 6 seconds, approximately 28 hours in total. The train, test and validation split ratio is 75 : 15 : 10. Notably, the audio guidance is not in terms of sounds of the target objects, but rather audio recordings of text descriptors.

\paragraph{AVSBench.} The AVSBench~\cite{zhou2022audio} dataset is a benchmark for audio-visual segmentation, aiming to segment sound sources in video frames according to the audio signal. This formulation of multimodal video segmentation is different from the speech referring video object segmentation as the audio is a sound generated from some objects in the video rather than a target reference given by a speaker. The dataset provides pixel-level annotations for the sounding objects in audible videos. It explores two settings: semi-supervised audio-visual Single Sound Source Segmentation (SS4) and fully-supervised audio-visual segmentation with Multiple Sound Sources (MS3). The videos are split into train, valid and test sets. There also are two subsets of data, single-source and multi-source audio. The subset focused on a single source comprises 4,932 videos spanning 23 categories, encompassing various sounds produced by humans, animals, vehicles, and musical instruments. The multi-source subset has 424 videos in total with train, valid and test splits of 296, 64 and 64 videos, respectively. Similar to other object segmentation tasks, the evaluation metrics used are mean IoU and F-measure.

\section{Summary}
Table \ref{tab:pixel level_perception_tasks} provides a summary of the video segmentation tasks presented in this chapter, as well as a recapitulation of all their acronyms. This chapter also overviewed the currently most widely used datasets for training and evaluation on these tasks. Discussion of state-of-the-art performance on these tasks, with an emphasis on transformer-based approaches, will be presented in Chapter~\ref{CH3}. 

\begin{landscape}
\begin{table}[t]
    \centering
    \begin{tabularx}{9in}{|c|X|X|c|c|X|}
        \toprule
        Category & Task & Target & Instances & Tracking & Datasets \\    
        \midrule
        \multirow{4}{*}{Objects} & Automatic Video Object Segmentation (AVOS) & Primary moving object & - & - & DAVIS'2016~\cite{perazzi2016benchmark}, MoCA~\cite{lamdouar2020betrayed}, YouTube-VOS~\cite{xu2018youtube}, YouTube-Objects~\cite{prest2012learning} \\   
             & Semi-automatic VOS (SVOS) & Mask-guided object & - & - & DAVIS'2017~\cite{pont20172017,caelles20182018} \\    
             & Interactive VOS (IVOS) & Scribble-guided object & - & - &  DAVIS'2017~\cite{pont20172017,caelles20182018} \\    
             & Video Instance Segmentation (VIS) & All Objects & \checkmark & \checkmark & YouTube-VIS~\cite{yang2019video}, OVIS~\cite{qi2022occluded} \\    
        \midrule
        Actor-action & Actor-action segmentation & Primary Object related to actions & - & - & A2D~\cite{xu2015can} \\
        \midrule
        \multirow{4}{*}{Scene} & Video Semantic Segmentation/ Video Scene Parsing (VSS/ VSP) & All thing and stuff classes &  - & - & VIPER~\cite{richter2017playing}, VSPW~\cite{miao2021vspw} \\   
             & Video Panoptic Segmentation (VPS) & All thing and stuff classes & \checkmark & \checkmark &  Cityscapes-VPS~\cite{kim2020video}, VIPER~\cite{richter2017playing}, VIPSeg~\cite{miao2022large} \\
             & Depth-aware Video Panoptic Segmentation (DVPS) & All thing and stuff classes and depth & \checkmark & \checkmark & Cityscapes-DVPS~\cite{qiao2021vip}, SemanticKITTI-DVPS~\cite{qiao2021vip}  \\
             & Panoramic Video Panoptic Segmentation (PVPS) & All thing and stuff classes & \checkmark & \checkmark & WOD:PVPS~\cite{mei2022waymo} \\ 
        \midrule
         \multirow{3}{*}{Multimodal} & Text guided VOS/Referring-VOS (RVOS) & Text reference guided object &  - & - & A2D-Sentence~\cite{gavrilyuk2018actor}, J-HMDB Sentence~\cite{gavrilyuk2018actor}, Ref-DAVIS~\cite{khoreva2019video}, RVOS~\cite{seo2020urvos} \\  
             & Audio Guided VOS (AGVOS) & Audio reference guided object & - & - & AVOS~\cite{pan2022wnet} \\   
             & Sound Source Segmentation & Sound producing objects & - & - & AVSBench~\cite{zhou2022audio}\\ 
        \bottomrule
      \end{tabularx}
    \caption{Summary of Video Segmentation Tasks.}
    \label{tab:pixel level_perception_tasks}
\end{table}
\end{landscape}

\pagebreak

\chapter{Transformers for Video Segmentation}\label{CH3}
\section{Overview}
In this chapter, we present a component-wise discussion of the most prominent transformer-based models proposed for various categories of video segmentation. The purpose of this chapter is to analyze the functional components of transformer-based video segmentation models across different segmentation tasks to develop a unified understanding of transformers as applied to video segmentation. We begin the presentation with a functional abstraction of components. Next, we present various design choices  for each component as found in recent literature. We also discuss task related specialization at relevant points of the presentation. Finally, we present the level of performance achieved by state-of-the-art transformer models on the segmentation tasks where they have been applied.

\subsection{Video segmentation model components}
\begin{figure}
    \centering
        \centering
        \includegraphics[width=1.0\textwidth]{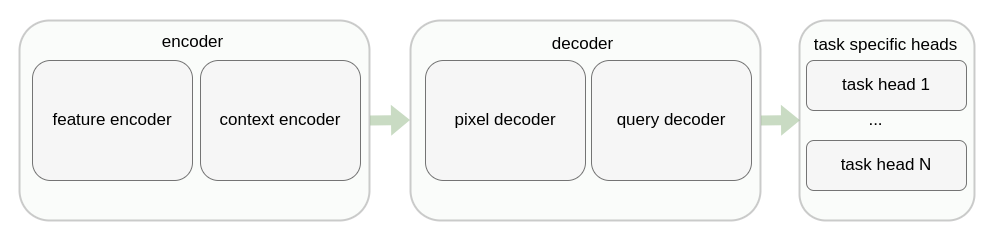}
\caption{Overview of Video Segmentation Transformer Model Functional Components.}
\label{fig:model_overview}
\end{figure}

The high level design of the functional components of a video segmentation transformer model is presented in Figure~\ref{fig:model_overview}. This diagram shows an abstraction  of the encoder-decoder model found in recent transformer-based solutions. The model components are subdivided based on the functional purposes they serve. Since the models evolve over time through research efforts with incremental improvements, some are instantiated with a subset of these components. More recent models contain most of these components with diversity in design, typically driven by performance and/or efficiency concerns. Different segmentation tasks follow the general architecture, with task specialization embodied in a final task specific head. Typically, the task head itself is a learned network of variable design. 

Despite the variety in segmentation tasks, video segmentation models typically use a common encoder-decoder approach. They begin by computing high resolution features with rich semantics from input image data and ultimately use a task-specific head to yield the desired segmentation. Rich semantics in this context means a representation containing high level information such as edges, contours, shapes and objects, in contrast to low level representation of pixel colour intensity. The encoder has a feature encoder to generate a semantically rich feature representation from the pixel space representation of the input video. The decoder generates high resolution semantic features and pixel-level masks. The final task head is a custom component for the specific segmentation objective under consideration, e.g., semantic or instance segmentation. 

The processing of an input video begins with a feature encoder, often referred as the feature backbone or extractor. The feature encoder takes input from a low dimensional colour space and maps to a semantically richer high dimensional feature space. For this reason, the recent trend is to use models trained on coarse granularity tasks (\eg classification) for the feature encoder and use transfer learning for application to other tasks of interest, \eg \cite{szegedy2015going,he2016deep,liu2022video}. The main reasons that models trained on classification tasks are good candidates for transfer learning with rich semantics are the availability of large-scale datasets for classification tasks and the manageable objective functions used in training that arise from the coarse granularity of classification. Recent video segmentation models generally use ConvNet models trained on ImageNet or transformer models trained on ImageNet~\cite{russakovsky2015imagenet} or Kinetics~\cite{carreira2017quo} as the backbone feature encoder, \eg \cite{szegedy2015going,he2016deep,liu2022video}. The context encoder, also known as the relation encoder, is designed to encode long range contextual relationships either spatially or spatiotemporally over the encoder features. This context encoder generally is equipped with a self-attention mechanism over a low resolution feature representation, \eg \cite{carion2020end,wang2021end,duke2021sstvos}. Since attention is a permutation invariant operation, locality is induced on the features via the use of a positional embedding. 

The spatial decoder (also called the pixel decoder) is developed for pixel level prediction tasks to generate high resolution features by propagating semantics from low resolution feature maps over a multiresolution feature space. Conventionally, this component is instantiated with the Feature Propagation Network (FPN)~\cite{lin2017feature} or its successors ~\cite{kirillov2019panoptic,xu2019auto,tan2020efficientdet,zhu2020deformable}. The query decoder (also called the mask decoder) generates pixel-level mask proposals corresponding to salient features or object instances. The query decoder consists of self-attention, cross-attention and a feed-forward network. 

The task specific head is the most heterogeneous component across different segmentation tasks. This component can be based on convolutional or linear layers. Depending on the task, it may also contain additional computation, \eg Hungarian matching~\cite{wang2021end} for video instance segmentation.

\subsection{Input preprocessing}\label{sec:input_preprocessing}
Since the multi-head attention mechanism works on sets of tokens, the input video requires preprocessing steps to represent it as a set of tokens. A generalized discussion of preprocessing steps for video is presented elsewhere~\cite{selva2023video}. In this section, we provide details on the input preprocessing steps specific to video segmentation, some of which also may be applied to other tasks (e.g., tokenization essentially is needed for any application of transformers to raw pixel data to keep dimensionality manageable). In particular, for transformer models of video segmentation, the preprocessing steps involve tokenization, token-embedding and positional embedding. 

\paragraph{Tokenization.} 
In vision transformers, tokenization refers to dividing an input image or video into smaller units or patches called tokens~\cite{dosovitskiy2020image}. All the tokens produced from an image or video are then serialized to produce a set that is transformed into token embeddings and then used as input to a transformer-based model. The patches are generally produced from non-overlapping spatial regions of various shapes for images, \eg $(8\times 8)$ or $(16\times 16)$. For videos, the patches can be spatiotemporal volumes for classification and spatial regions for pixel level estimation (due to the requirement of preserving the temporal dimension). This set of tokens is then projected to a token embedding using an embedding layer. Alternatively, in a hybrid model where transformers are used on the output of a ConvNet-based feature extractor, tokenization is done on the ConvNet features to produce a set of tokens from the feature maps. Since the ConvNet features are already in a lower spatial resolution, tokenization of these features may involve a small region, \eg one pixel. In place of token embedding, these hybrid models use a linear transform to project high dimensional ConvNet features (\eg 2048) to a lower dimension (\eg 384).

\paragraph{Token embedding.} The token embedding transforms the spatial region or spatiotemporal volume of the token with multiple pixels to a single feature vector known as the token embedding. When using patch-embedding on high resolution data, such as RGB values or early layer features of a ConvNet, a spatial patch of size greater than a single pixel is used as the token. In that case, feature aggregation of all the elements under this window is used to transform it into a single feature vector. This operation is performed by a combination of simple averaging, concatenating the pixel features, convolution and linear layers. The convolutional or linear layer can additionally serve to transform the feature dimension to a specific dimension to be used in the subsequent transformer model. Alternatively, when using tokenization on the low resolution features of a ConvNet network, often a single pixel feature is transformed into a single token requiring no further processing for spatial dimension reduction. Even so, a feature projection still can be used to transform the feature dimension of the token-embedding. The components involved here are often referred to as a token-embedding network.      

\paragraph{Positional embedding.} Positional embedding is used to augment the token embedding to capture the positional dependency of the tokens in the set. This addition is important because the attention mechanism is permutation invariant, while position can be important in many vision tasks, including segmentation. Different forms of position embedding are used in transformer-based models to encode the position of elements in an input set. The positional embedding can be fixed or learned depending on whether they are analytically defined or learned during the training step. Further, it can be  absolute or relative depending on how the position embedding is aggregated with the token-embedding. Details on the different forms of positional embeddings can be found elsewhere ~\cite{dosovitskiy2020image}.

\subsection{Transfer learning}
Video segmentation aims to provide pixel-wise labels. Objective functions for learning at such fine granularity are difficult to train from scratch \cite{van2014transfer}. Moreover, generation of large-scale datasets that have been annotated with this level of granularity is not practical. For these reasons, video segmentation models use transfer learning from video-level or image-level granularity tasks. As one specific example, video object segmentation generally uses models trained on image or video classification tasks~\cite{russakovsky2015imagenet,kay2017kinetics} to initialize the feature extractor backbone. In particular, many video segmentation models use ConvNet models trained for ImageNet~\cite{russakovsky2015imagenet} classification as a backbone feature extractor, \eg \cite{szegedy2015going,he2016deep}. These models use pre-trained weights from the classification tasks to initialize the segmentation model for transfer learning. VGG~\cite{simonyan2015very} and ResNet~\cite{he2016deep} models are the dominant backbones in those models. Recent video segmentation models with a transformer-based feature backbone use transformer models trained on the Kinetics action classification dataset~\cite{kay2017kinetics}. It also is found that first training the classification model for object detection (\eg on COCO~\cite{lin2014microsoft}) and subsequently using those trained weights to initialize the segmentation model leads to further improved learning~\cite{karim2020distributed,zhong2020squeeze,wang2021unidentified}. 



\section{Transformer encoder}
Typically, the transformer encoder operates in two steps: feature encoding and context encoding. Feature encoding serves to extract an initial representation of the video from colour input that is suitable for delineating items of segmentation interest. The context encoding serves to capture relationships between the extracted features across both space and time and to do so over large extents to enrich the initial features. The remainder of this subsection describes these two steps in detail.

 
\subsection{Feature encoder}
In general, the first step of encoding in video segmentation models involves extracting high dimensional feature representations from the input data. A large number of transformer-based models for video segmentation (\eg \cite{paul2020efficient,wang2021end,cheng2020panoptic,hwang2021video,yang2021crossover, rana2021we,wang2021temporal,paul2021local,hu2020real,qiao2021vip,cheng2021mask2formervid,li2021video,qiao2021vip,wu2022language,pan2022wnet,wu2022seqformer,kim2022tubeformer,zhou2022slot,yuan2022polyphonicformer,wu2022language,pan2022wnet}) use a ConvNet (\eg \cite{simonyan2015very,he2016deep,chollet2017xception,romera2017erfnet,zagoruyko2016wide}) as feature extractor with a transformer in subsequent processing steps. Another trending approach (\eg ~\cite{yang2022temporally,sun2022coarse,kim2022tubeformer,zhou2022slot,yuan2022polyphonicformer}) is to use an image transformer (\eg \cite{dosovitskiy2020image,xie2021segformer,wang2021pyramid,liu2021swin}) to extract features of individual frames and then aggregate to get video features. Most recently, models have appeared (\eg \cite{botach2022end,wu2022language,karim2023med}) that use spatiotemporal transformers (\eg \cite{liu2022video}) as their feature extractor. The input data to some video segmentation tasks include information from other modalities in addition to images, \eg text or audio for referring video object segmentation. In cases of multimodal segmentation, there is an additional feature extractor for the additional input modalities and the cross-modal features are later aggregated. Recent models on multimodal video segmentation employ a transformer-based text encoder to extract text features~\cite{vaswani2017attention,devlin2018bert, liu2019roberta} and Mel Frequency Cepstral Coefficients (MFCCs)-based  models to extract audio features~\cite{bouchakour2018mfccs}. 

The ConvNet-based feature extractors follow a similar hierarchical architecture using convolution, rectification, normalization and pooling layers to generate a feature hierarchy of successively decreasing resolution and increasing semantics \cite{simonyan2015very,he2016deep,chollet2017xception,romera2017erfnet,zagoruyko2016wide}. Different ConvNet models have their own characteristics in their internal architecture, such as VGG~\cite{simonyan2015very} presented an architecture using a moderately deep architecture with small filters, while ResNet~\cite{he2016deep} used residual connections to facilitate the training of a very deep network. These architectures have been widely in use as feature extractors for video segmentation due to their ability to generate hierarchical features from input images. A detailed discussion of these ConvNet models is available elsewhere~\cite{hadji2018we}.

Given the recent trend of using spatial transformers \cite{dosovitskiy2020image,xie2021segformer,wang2021pyramid,liu2021swin} and spatiotemporal transformers~\cite{liu2022video} as backbone feature extractors in video segmentation, in the following we provide a discussion of transformer-based feature extractors. Early image transformers (e.g., ViT \cite{dosovitskiy2020image}) operate at a single scale, which limits applicability to segmentation where multiscale features are useful in distinguishing segmentation targets. Multiscale features are useful for segmentation as the distinguishing aspects of targets typically are manifest across a range of scales. In response, image transformers with multiscale features have been developed \cite{chu2021twins,xie2021segformer,wang2021pyramid,liu2021swin}, somewhat akin to multiscale ConvNet features. Notably, these features are captured on a framewise basis and thereby leave all temporal processing to subsequent stages. 



Pyramid Vision Transformer (PVT) was the first model to include hierarchical multiresolution features in a transformer backbone~\cite{wang2021pyramid}. PVT extracts features from images that are of successively decreasing resolution. In particular, the PVT architecture is composed of four stages with progressive shrinking of output resolution by successively increasing the strides. Each stage is comprised of a patch embedding layer and several transformer encoders with multi-head self-attention. The resulting model benefits from both the hierarchical structure of ConvNets as well as the long range data association of transformers. Results show improvement over comparable ConvNets on object detection and semantic segmentation \cite{wang2021pyramid}. A subsequent extension also adapted the model to video instance segmentation \cite{yang2022temporally}.


An efficient image transformer backbone used in video segmentation tasks is the Swin transformer~\cite{liu2021swin}. The Swin transformer achieves improved efficiency by using a shifted window mechanism in a hierarchical transformer backbone. The model is presented as a general purpose transformer backbone offering linear computational complexity in image size. The architecture involves four stages with patch merging inbetween stages. Each stage consists of alternating blocks with Windowed Multi-head Self-Attention (W-MSA) and Shifted-Windowed Multi-head Self-Attention (SW-MSA). The shifting window mechanism allows for cross-window attention. The model achieves linear complexity by limiting the self-attention locally within the non-overlapping windows of fixed size patches that partition an input spatially.  The hierarchical feature representation is constructed by beginning with small-sized patches and gradually merging neighboring patches in deeper stages. The Swin transformer has been adopted in many recent video segmentation tasks~\cite{cheng2021mask2formervid,wu2022seqformer,kim2022tubeformer,zhou2022slot,yuan2022polyphonicformer,wu2022language}. 


SegFormer introduced another notable hierarchical image transformer encoder named the Mix Transformer (MiT) encoder~\cite{xie2021segformer}. MiT is designed on top of PVT with a number of novel features, including overlapped patch merging and a positional-encoding-free design. For efficient self-attention, a sequence reduction process is adopted, following previous work~\cite{wang2021pyramid}. The positional encoding free design allows MiT to not depend on interpolation of positional codes during testing with a different resolution input. Several variants also were presented using the same architecture with different sizes, named MiT-B0 to MiT-B5. The MiT encoder has been used as a feature extractor for transformer-based video semantic segmentation~\cite{sun2022coarse}. Recently, Hierarchical Feature Alignment Network (HFAN)~\cite{pei2022hierarchical} also used the MiT backbone for automatic video object segmentation. 



Video transformers trained for video level granularity (e.g., action recognition \cite{kay2017kinetics}) have been used as the basis of feature extraction for segmentation. Early video transformer models trained on action classification used a transformer model on top of a 2D feature encoder backbone. Video Transformer Network (VTN)~\cite{neimark2021video} used LongFormer~\cite{beltagy2020longformer} on top of ResNet~\cite{he2016deep} or DeiT~\cite{touvron2021training}, and augmented the features with an additional token for classification. Despite being efficient for long videos, VTN did not encode temporal information in the feature backbone.  Video Vision Transformer (ViViT)~\cite{arnab2021vivit} is a purely transformer-based model that encodes spatiotemporal tokens extracted  from an input video using a transformer encoder consisting of a series of 3D spatiotemporal self-attention layers. To handle long input sets, a factorized spatiotemporal encoder was employed. Both VTN and ViViT models generate features at a single resolution making them less attractive for segmentation where fine localization from multiscale features is an important design principle.

The Multiscale Vision Transformer (MViT) is a transformer model that progressively grows the feature dimension with simultaneous reduction of the spatiotemporal resolution~\cite{fan2021multiscale}. MViT achieves these hierarchical feature stages by using multi-head pooling attention blocks to reduce spacetime resolution at successive stages. In contrast to shrinking~\cite{wang2021pyramid} or patch-merging~\cite{xie2021segformer}, pooling operations in query, key and value are used to reduce the feature resolution (2D or 3D). Temporal dynamics modeling analysis by shuffling the frame orders show that this models presents strong temporal modeling capabilities~\cite{fan2021multiscale}. MViT has been extended with improved embedding and pooling mechanisms~\cite{li2022mvitv2} as well as via memory augmentation~\cite{wu2022memvit}; however, these versions have yet to be employed in video segmentation.


The Video Swin Transformer~\cite{liu2022video}, an extension to the Swin Transformer~\cite{liu2021swin}, proposed a transformer model with shifted windows that combines the locality of ConvNets using windowing and the long range data association of transformers through attention mechanisms. The extension from the Swin transformer to the Video Swin Transformer is done by extending the scope of local attention computation from only the spatial domain to the spatiotemporal domain. Moreover, the model is very fast due to the windowing mechanism. Further, the model allows bridging between the windows of two successive layers by means of the shifted windows. By associating information at window boundaries, the modeling power is enhanced, resulting in performance improvement on various tasks. Due to the benefits of both speed and performance in encoding video data, the Video Swin transformer is recently gaining popularity as a feature backbone in recent video segmentation tasks~\cite{botach2022end,wu2022language}.

Another method, SIFA-Transformer~\cite{long2022stand}, uses a form of deformable attention to rescale the offset estimates by the difference between two frames. Deformable attention uses dynamic sampling to adaptively generate a sparse attention matrix~\cite{zhu2020deformable}. This method achieved superior results on video action recognition compared to MViT and Video Swin, suggesting a possible stronger feature encoding in the model~\cite{wang2022deformable}. While not yet applied to video segmentation tasks, SIFA-Transformer has potential in this area.


\subsection{Context encoder}\label{sec:context_encoder}
The vast majority of transformer-based video segmentation approaches use ConvNets as their feature encoder and thereby require a global context aggregation in the feature maps. Somewhat similarly, many transformer-based feature encoders use attention in a local window or pooling with large strides in attention. Moreover, in multimodal video segmentation, there is a need for long range relationship aggregation on the combined features generated from multiple input modalities for multimodal video segmentation. To address these challenges, transformer-based models typically use a context encoder, also referred to as a relation encoder, equipped with global self-attention to capture long-range relationships between tokens. More generally, any feature encoder can achieve the added advantage of enriched features by global context encoding. The context encoder helps the model capture the spatial and temporal dependencies between elements in the video frames, allowing it to better understand the semantic content of the scene. This representation can lead to more accurate segmentation results, especially in complex scenes with multiple interacting objects.

The general architecture of the context encoder, as proposed in DETR for object detection~\cite{carion2020end}, begins with feature dimension reduction and tokenization to become suitable for global self-attention processing. Next, the feature maps are tokenized to get a token embedding. Subsequently, a set of transformer encoder layers process the tokens. Each transformer encoder layer has a standard multi-head self-attention and a feed forward network (FFN). To incorporate position information in the permutation invariant transformer architecture, DETR added a fixed positional encoding~\cite{parmar2018image,bello2019attention} with the tokens.

Approaches to video instance segmentation that have employed context encoding include the following. Video instance segmentation with Transformers (VisTR)~\cite{wang2021end} adapted the context encoder from DETR to model the similarities among all the pixel level features in a clip extracted by a ConvNet backbone. It used the context encoder on the deepest layer features of the ConvNet backbone to model all pairwise similarities between elements in the token sets. Inter-frame Communication transformers (IFC)~\cite{hwang2021video} used additional tokens to store compact context information of each frame of a scene in memory. The inter-frame correlation is achieved by aggregating relationships between the memory tokens. Specifically, two transformers are used to yield this encoder: one to encode the frame context summary and the other to exchange information between frames. Thus, the memory encoder works on each frame individually and the communication encoder works between the frames. The Sequential Transformer (SeqFormer)~\cite{wu2022seqformer} used a transformer-based context encoder in each frame independently. SeqFormer argued for handling the 2D space domain and 1D time domain differently based on the view that the spatial and temporal domains are fundamentally distinct~\cite{zhao2018trajectory}. The framewise encoding employed deformable attention~\cite{zhu2020deformable}. Subsequently, information is aggregated across all frames via region sampling. Ablation studies were presented to support the design of distinct spatial and temporal processing. A recent multiscale approach to video instance segmentation, MS-STS~\cite{thawakar2022video}, uses multiscale spatiotemporal split attention in its context encoder to effectively capture spatiotemporal feature relationships across frames in a video. It uses intra- and inter-scale temporal attention to generate enriched features during encoding. Additionally, it uses an adversarial loss to enhance foreground-background separability.

Various types of context encoder also are used in semi-supervised VOS. TransVOS~\cite{mei2021transvos} used a context encoder to model the spatiotemporal relationships among pixel-level features of the input token set. This approach used ConvNet features and adopted a context encoder architecture similar to DETR. Another approach to semi-supervised VOS, Sparse Spatiotemporal Transformers (SST)~\cite{duke2021sstvos} used a context encoder with sparse-attention~\cite{child2019generating} for computational efficiency. The SST encoder used sparse multi-head attention, in contrast to global attention. In particular, a sparse connectivity between the key and value was used in the multi-head attention, based on the video spacetime coordinates. 

Curiously, it appears that only a single approach to automatic VOS has employed a transformer-based context encoder. In particular, MED-VT, used a transformer-based multiscale context encoder for unsupervised video object segmentation (as well as actor-action segmentation)~\cite{karim2023med}. This approach used within-scale and between-scale transformer encoding on the multiscale features from the feature extractor. The within-scale attention works on each feature scale and the between-scale attention works between two different feature scales. While it appears that no other AVOS approach has used transformer-based context encoding, probably the closest ones using the attention mechanism for AVOS are Hierarchical Feature Alignment Network (HFAN)~\cite{pei2022hierarchical} and Prototype Memory Network (PMN)~\cite{lee2023unsupervised}. HFAN used a mechanism similar to self-attention to fuse appearance and motion features at intermediate stages of encoding, while PMN used an attention mechanism to adaptively score and select the most useful superpixel-based component prototypes in each frame as candidates to a memory bank.

Finally, approaches have appeared that address multimodal segmentation using transformers. Multimodal Tracking Transformer (MTTR) for text Referring Video Object Segmentation (RVOS) presented an end-to-end transformer-based approach~\cite{botach2022end}. This approach employed a single multimodal transformer for context aggregation jointly across the two modalities. MTTR initially extracts features separately from the input text and video frames. Next, the features from the two modalities are flattened and concatenated to form a multimodal feature set. The context encoder then uses self-attention on the multimodal feature set to exchange information between the modalities. Another approach, ReferFormer, also used context encoding to fuse text and video features~\cite{wu2022language}. This approach adopted deformable transformer~\cite{zhu2020deformable} as its transformer model for efficient global attention. The deformable transformer enhances the standard attention mechanism by allowing the model to selectively focus on different positions in the input with varying degrees of importance. By introducing learnable offsets, the deformable transformer adapts its attention window dynamically, capturing more fine-grained dependencies and improving performance on tasks requiring long-range context understanding. The approach projects multiscale visual features to lower dimension and multiplicatively combines the projection with text features. Finally, a context encoder operates on the fused features. Yet another multimodal video segmentation approach, Wnet, for audio-guided video object segmentation, used a transformer encoder to learn a cross-modal representation of the video and audio features~\cite{pan2022wnet}. In contrast to concatenating or multiplying, a wavelet-based cross-modal module is used to fuse the two modalities. In this context encoder, cross-attention is used with visual features as queries and audio features as keys and values. This cross-modal context encoder uses a wavelet based component in addition to the multi-head attention and feed forward network.

\section{Transformer decoder}
\begin{figure}[h!]
    \centering
    \begin{subfigure}[b]{0.9\textwidth}
    \centering
    \includegraphics[width=0.8\textwidth]{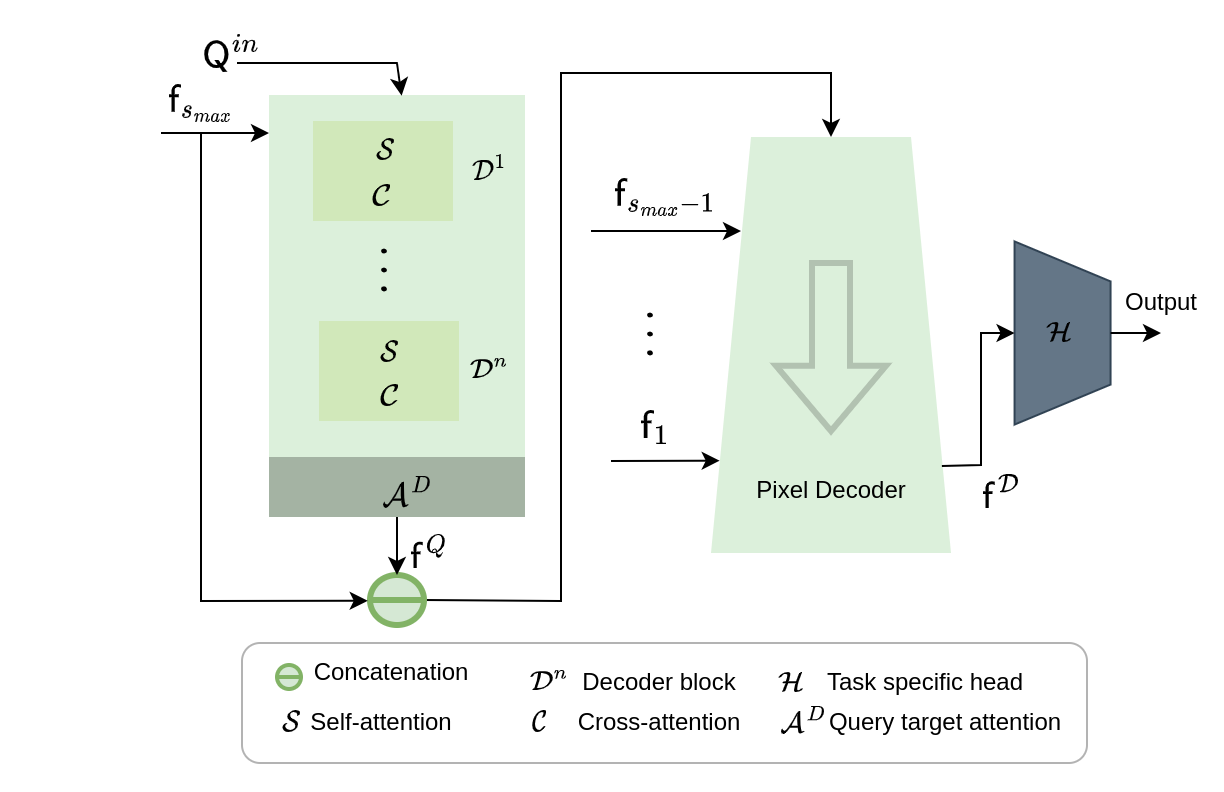}
    \caption{A query decoder generally is used before a pixel decoder in single scale decoding models. Here, $\mathsf{f}_1, ..., \mathsf{f}_{s_{max}}$ are the feature maps of successively deeper layers; $\mathsf{Q}^{in}$ is query initialization; $\mathcal{D}^1, ..., \mathcal{D}^n$ are query decoder layers each containing self-attention, $\mathcal{S}$, and cross-attention, $\mathcal{C}$; $\mathcal{A}^D$ is a query attention module that follows the query decoder; and $\mathsf{f}^Q$ is output feature from the query attention module. In this type of architecture (\eg VisTR~\cite{wang2021end}), generally  the high resolution feature maps from pixel decoder, $\mathsf{f}^\mathcal{D}$, are used as input to a task specific head, $\mathcal{H}$.}
    \end{subfigure}  
    \begin{subfigure}[b]{0.9\textwidth}
    \centering
    \includegraphics[width=0.8\textwidth]{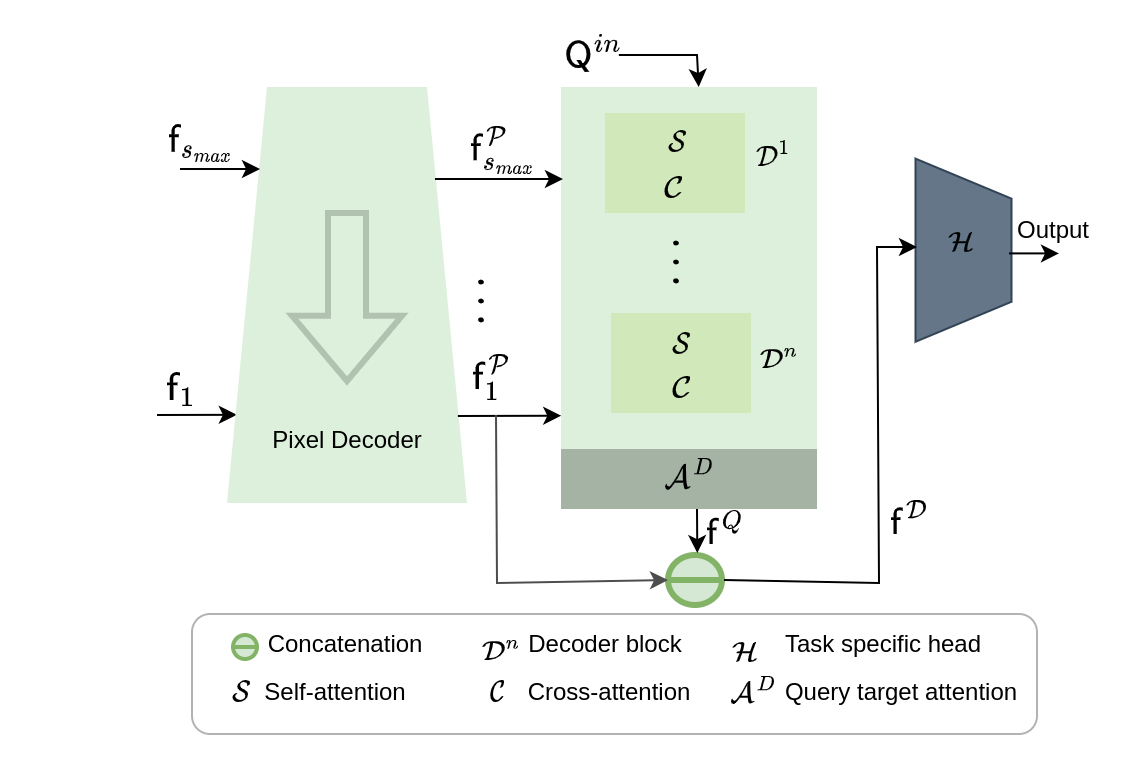}
    \caption{A pixel decoder is used before query decoding in multiscale decoding. Here, $\mathsf{f}_1, ..., \mathsf{f}_{s_{max}}$ are the feature maps of successively deeper layers; $\mathsf{f}^\mathcal{P}_1, ..., \mathsf{f}^\mathcal{P}_{s_{max}}$ are the multiscale feature maps from pixel decoder; $\mathcal{D}^1, ..., \mathcal{D}^n$ are query decoder layers, each containing self-attention, $\mathcal{S}$, and cross-attention, $\mathcal{C}$; $\mathcal{A}^D$ is the query attention module that follows the query decoder; and $\mathsf{f}^Q$ is the output feature from the query attention module. In this type of architecture (\eg MED-VT~\cite{karim2023med}, Mask2Former~\cite{cheng2022masked,cheng2021mask2formervid}) the output feature of the query attention module, $\mathsf{f}^q$, and highest resolution feature from pixel decoder, $\mathsf{f}^\mathcal{P}_1$, are concatenated to produce decoder feature, $\mathsf{f}^\mathcal{D}$, which subsequently is used as input to a task specific head, with $\mathcal{H}$ the task specific head.}
    \end{subfigure}  
  \caption[Pixel Decoder and Query Decoder]{Pixel Decoder and Query Decoder in Pixel Level Estimation Model. In single scale designs, query decoding typically precedes pixel decoding (a); in multiscale designs, the opposite is true (b).}
  \label{fig:transformer_decoder}
\end{figure}

The decoding phase of video segmentation tasks includes generating high resolution semantic features and mask proposals, followed by the final segmentation. Decoding typically operates in two steps, while the order of the steps depends on the design choices of the particular architecture considered. Some architectures use a query decoder initially to learn dynamic queries to represent a set of mask proposals, followed by a pixel decoder for generating high resolution semantic features. Other variants initially use a pixel decoder to produce multiresolution semantic features and then a query decoder to learn the dynamic queries from those features. In this section we will discuss various forms of spatial decoder and query decoder found in the literature. We present a brief visual overview of the general trends in the combination of pixel decoder and query decoder for video segmentation in Figure ~\ref{fig:transformer_decoder}.


\subsection{Pixel decoder}
In the context of pixel level video understanding, we define a pixel decoder (also known as a spatial decoder) as a model component used to produce high resolution features with rich semantic abstraction. In general, a pixel decoder takes as input a set of multiscale features of different resolutions and different levels of abstraction. The multiscale features result from several intermediate stages of transformer encoding  and may also be augmented with features from the backbone feature extractor. This multiscale feature set thereby contains features of successively lower spatial dimension, albeit higher in channel dimension and richer in semantics. In this regard, the pixel decoder is employed to propagate the rich semantics from deeper layer low resolution features to the early layer high resolution features. Thus, the multiscale feature output from the pixel decoder has features that all are rich in semantic information. Depending on the design choice, the subsequent modules of the pixel level prediction model can use any or all of the higher resolution features for task specific mask generation. 

Pixel decoding addresses three important aspects of pixel level image/video understanding. First, small objects in a scene may have lost their detailed shape or appearance in the deep layer features due to successive striding operations. Indeed, in practical scenarios, objects can appear at various scales in an image or video frame. For example, the output features of a traditional ConvNet are generally 32 times smaller in spatial resolution compared to the input. A small object in a scene may have lost the original shape at this scale. Second, pixel level prediction requires precise localization at the granularity of a pixel. However, masks generated from such low resolution features may lose precise localization when upsampled to the input size. Third, early layer features, while precise in localization, may lack in semantics (e.g., awareness of objects and context) to unambiguously detect or label on a local basis. The pixel decoder addresses these issues by producing high resolution features with semantics from deep layer features.

\begin{figure}[h!]
    \centering
        \begin{subfigure}[b]{0.35\textwidth}
        \includegraphics[width=0.95\textwidth]{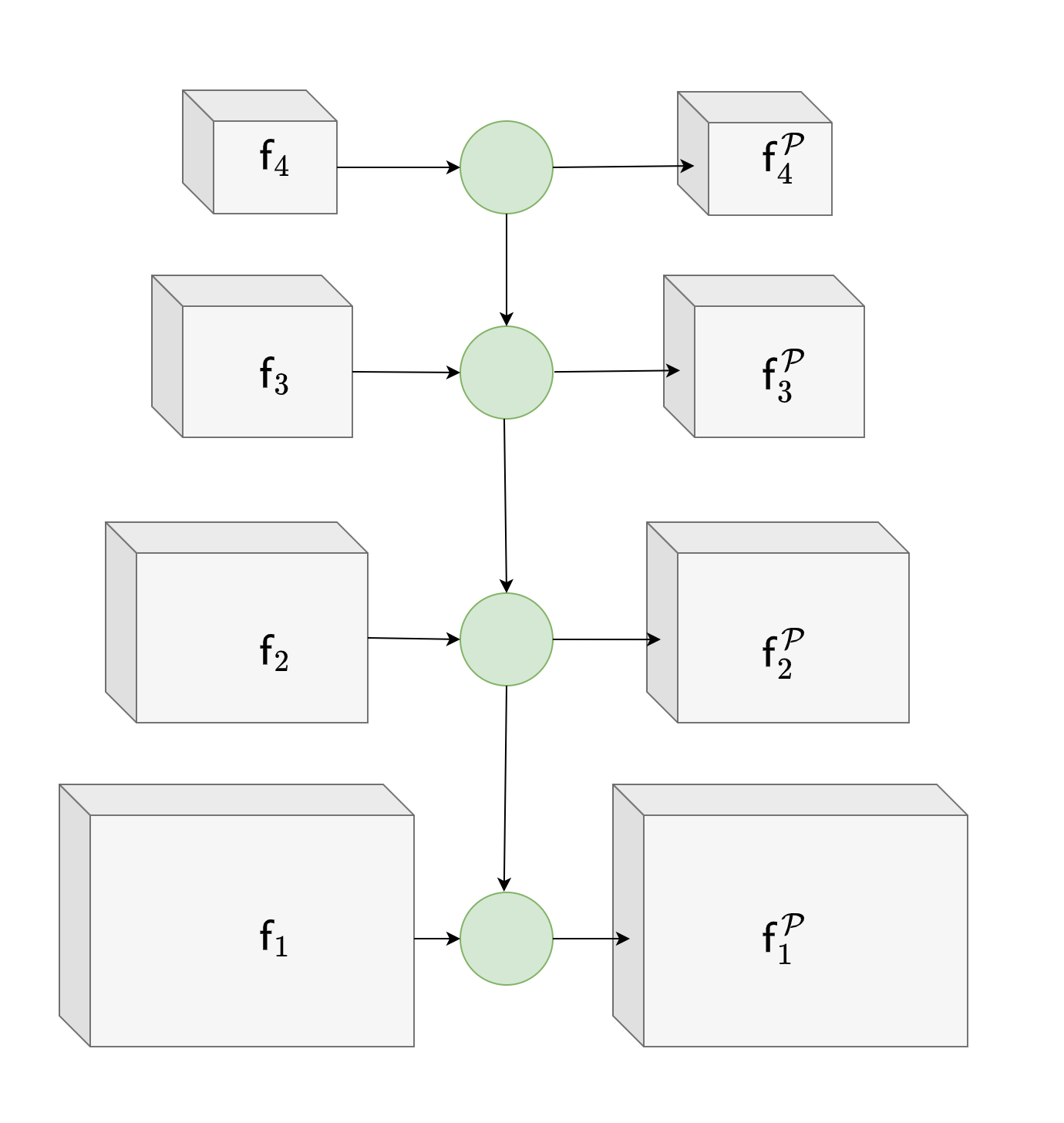}
        \caption{FPN~\cite{lin2017feature} with a single top-down pathway.               }
        \end{subfigure}  
        \begin{subfigure}[b]{0.41\textwidth}
        \includegraphics[width=0.95\textwidth]{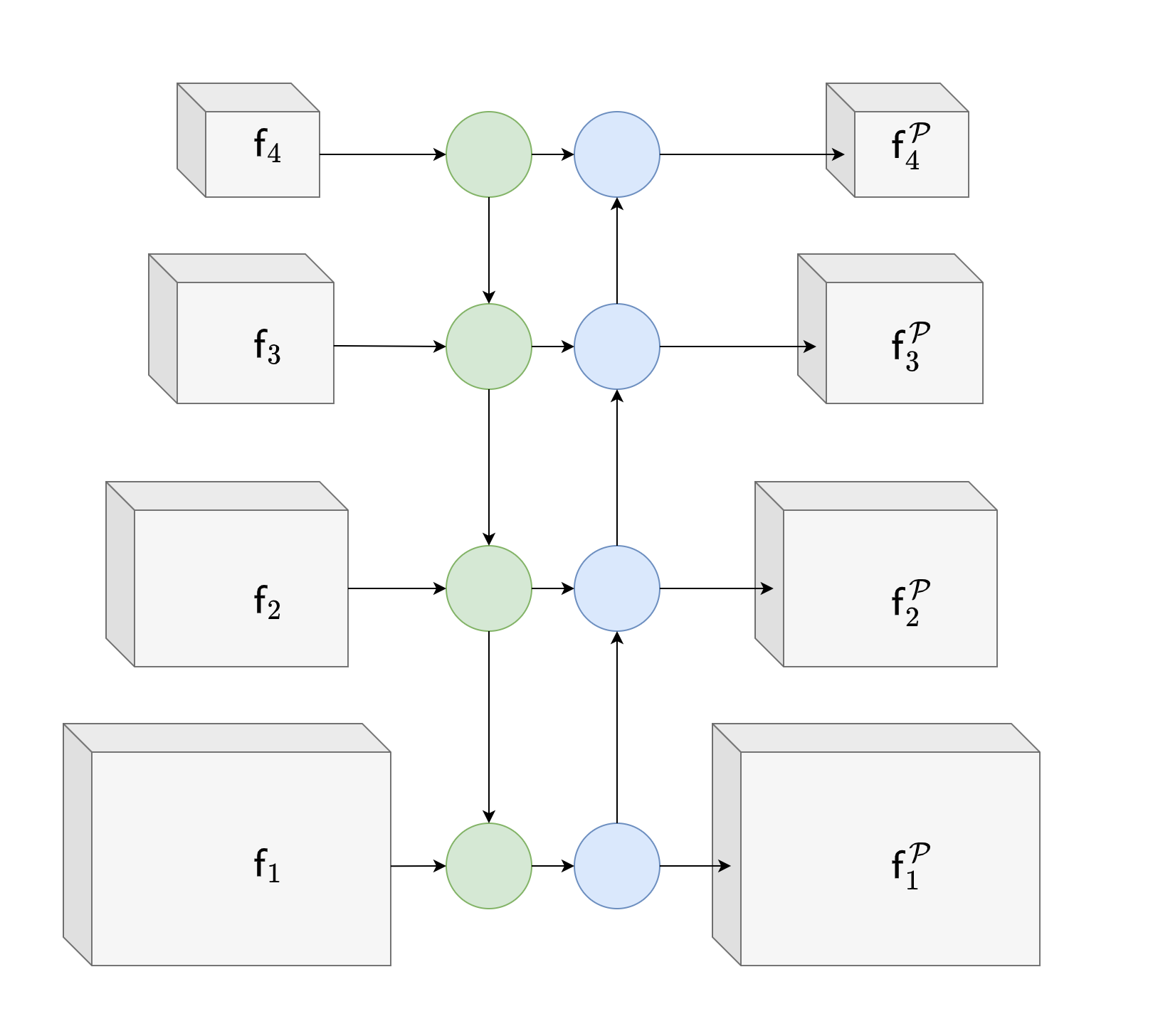}
        \caption{Panoptic-FPN~\cite{kirillov2019panoptic} with a single top-down and an additional bottom-up pathway.}
        \end{subfigure}  
        \begin{subfigure}[b]{0.7\textwidth}
        \includegraphics[width=0.80\textwidth]{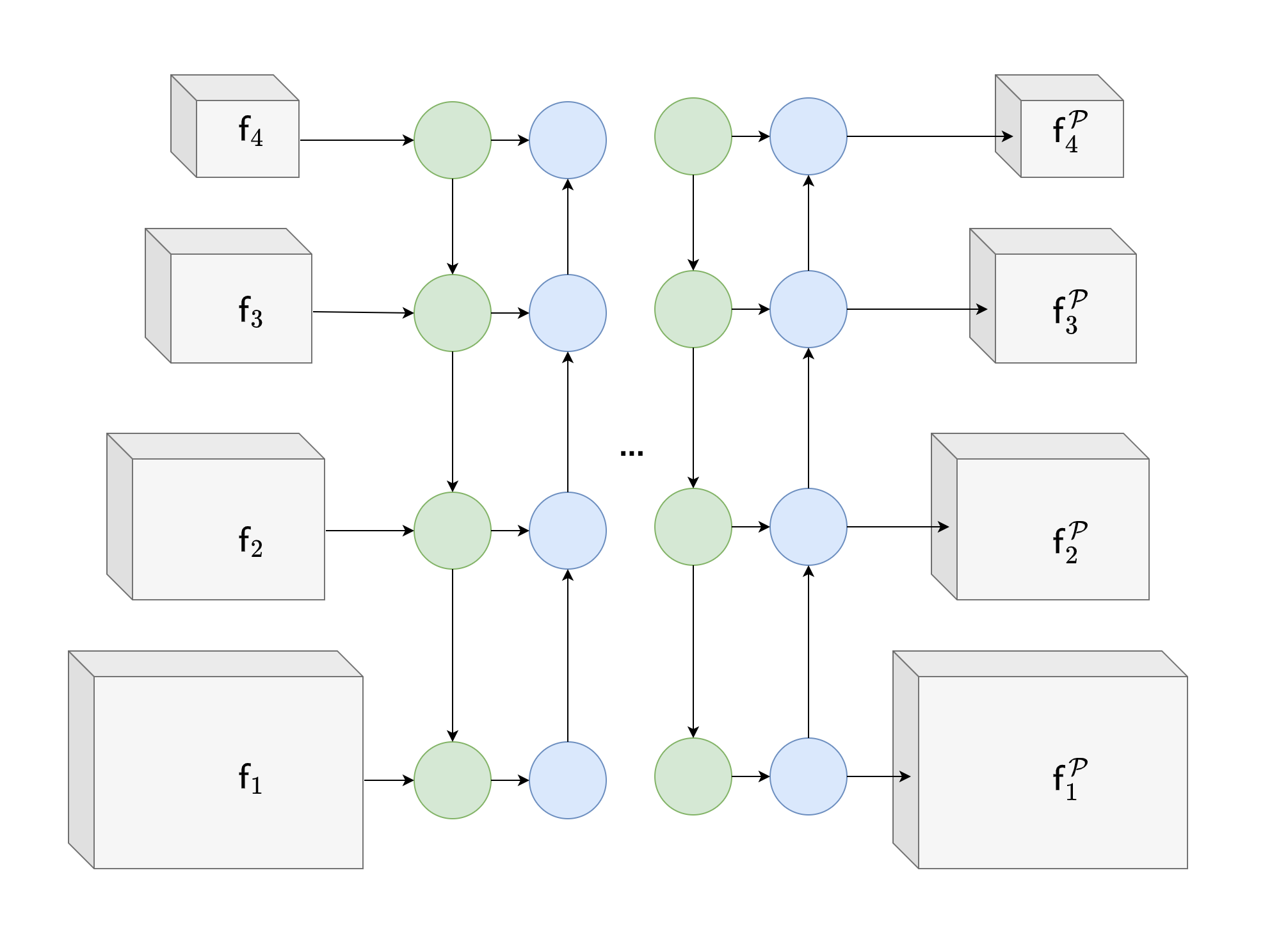}
        \caption{BiFPN~\cite{tan2020efficientdet} with multiple alternative top-down and bottom-up pathways.}
        \end{subfigure}      
    \caption[Recent Trends in Pixel Decoder Design]{Recent Trends in Pixel Decoder Design. A schematic diagram of recent trends in the design of pixel decoder showing a) FPN~\cite{lin2017feature}, b) Panoptic-FPN~\cite{kirillov2019panoptic} and c) BiFPN~\cite{tan2020efficientdet}. In all three figures, $\mathsf{f}_1$, ..., $\mathsf{f}_4$ are four input features to the pixel decoder, $\mathsf{f}^\mathcal{P}_1$, ..., $\mathsf{f}^\mathcal{P}_4$ are four output features from the pixel decoder, green circles are top-down pathways and blue circles are bottom-up pathways. Four features are shown for a concise depiction and comparison, the number of features can vary across different implementations. The circles of top-down and bottom-up pathways generally contain convolution, up-sample and arithmetic addition operations.} 
  \label{fig:pixel_decoder_sota}
\end{figure}

Recent developments in the design of pixel decoders are shown in Figure ~\ref{fig:pixel_decoder_sota}. A widely popular architecture for pixel decoding is the Feature Pyramid Network (FPN)~\cite{lin2017feature}, initially proposed for object detection. This architecture produces high-level semantic feature maps at all scales using lateral connections on the feature encoder in a top-down manner. The design operates by creating a feature pyramid based on hierarchical ConvNet features. Initially, each feature map of the pyramid is projected to a common embedding dimension (\eg 256 channels). The feature embedding from the deepest layer is then up-sampled and added to the next higher resolution feature embedding. This feature embedding is then further up-sampled and added to the next highest resolution. This process is continued until the highest resolution feature embedding. Finally, all these new feature embeddings are input to common mask heads to generate estimates at multiple resolutions.  

The Feature Pyramid Network was presented as a generic solution for building feature pyramids from the feature hierarchies of backbone feature extractors, \eg ConvNets. The model was originally proposed for object detection and segmentation from images. For object detection, the model was adopted in the Region Proposal Network (RPN)~\cite{ren2015faster} and Fast R-ConvNet ~\cite{girshick2015fast}. Additionally, it also was demonstrated for segmentation proposal generation. The original image based approach is now widely used for videos.

Panoptic-FPN~\cite{kirillov2019panoptic} added a FPN to Mask R-CNN~\cite{he2017mask} for panoptic segmentation. Two task heads were employed, with one for instance segmentation to capture things and the other for semantic segmentation to capture stuff. As part of the innovation, the architecture modified the original FPN by up-sampling all intermediate features to the highest resolution features. The resulting approach was found to perform better for semantic segmentation compared to Mask R-CNN. Several other approaches have followed this general path. PANet~\cite{liu2018path} adds an additional bottom-up pathway on top of a FPN for instance segmentation. PAN~\cite{lipyramid} added a global attention up-sample module to the FPN for the task of semantic segmentation.  NAS-FPN~\cite{ghiasi2019fpn} and Auto-FPN~\cite{xu2019auto} presented an improved FPN architecture by using a Neural Architecture Search (NAS) approach. NAS is a method used to automatically search for optimal network architectures by exploring and evaluating a large space of possible architectures for a given task. EfficientDet \cite{tan2020efficientdet} presented a weighted Bidirectional Feature Pyramid Network (BiFPN) for fast multiscale feature fusion. It extended the idea of PANet by adding repeated blocks of top-down and bottom-up feature aggregation. Additionally, it used a learnable weighting method for each feature during aggregation to emphasis their different contributions. The approach performed better on object detection and instance-level segmentation compared to using FPN. Another approach, FaPN, is designed to contextually align up-sampled higher-level features~\cite{huang2021fapn}. For this purpose, a feature selection module is used to emphasize deeper features with richer semantics. These features are upsampled and spatially registered to higher resolution features via a learnable alignment module. The approach is demonstrated across four pixel level prediction tasks: object detection, semantic, instance and panoptic segmentation.

Recently, attention-based pixel decoding has been found to provide superior performance over ConvNet-based pixel decoding. Deformable-DETR~\cite{zhu2020deformable} used an attention-based pixel decoder for object detection, which later was adopted for segmentation. This pixel decoder also is known as a deformable attention pixel decoder. The decoding combines the sparse spatial sampling of the deformable convolution operation and the relation modeling of the transformer to yield a multiscale deformable self-attention module. Subsequently, Mask2former~\cite{cheng2022masked} adopted this deformable attention pixel decoder for segmentation tasks. This multiscale attention module works on top of a multiscale feature pyramid and aggregates the transformer-based context encoder and FPN in a single module. Due to the quadratic complexity of self-attention, transformer-based context encoding is not practical for high resolution features from the feature encoder. The use of sparse sampling allowed the self-attention to be computationally feasible on high resolution feature maps. This pixel decoding approach was found to perform consistently better than other ConvNet-based pixel decoding approaches across various segmentation tasks.

\subsection{Query decoder}\label{sec:mask_decoder}
The query decoder used in video segmentation takes as input the feature maps from a pixel decoder. This module also is known as mask decoder or bounding box decoder. The query decoder generates segmentation masks using successive iterations of self-attention and cross-attention mechanisms. The query decoder design is inspired by DETR~\cite{carion2020end}, which used learned queries to formulate object detection as a set prediction problem. In particular, learned queries are used to capture the centre, height and width of bounding box estimates. Subsequently, the approach has been extended to pixel level segmentation for mask prediction \cite{wang2021end,cheng2022masked}.   

There are several benefits of using a query decoder for pixel level segmentation in images or videos. First, the query learned from a spacetime volume of video features can generate pixel level masks for the volume in an adaptive fashion. Second, the mask decoding approach can unify multiple segmentation tasks in a single framework~\cite{cheng2022masked}. Third, the resulting model is end-to-end trainable without requiring an additional proposal generation network for videos. Fourth, the model is able to capture long range relationships between video targets (e.g., objects).

For video, the query decoder takes two inputs: flattened spacetime feature volumes and a randomly initialized set of queries. The flattened spacetime feature volumes also are referred to as a key-value memory entries in this context. The query decoder processing steps include self-attention, cross-attention and a Multi-Layer Perceptron (MLP). Initially, self-attention is performed on the query input. Next, cross-attention is performed between query and spacetime features using the spacetime feature as key and value. Finally, the query is processed with a MLP. After multiple layers of such operations, the output queries operate on the memory features to generate attention maps. These attention maps can be used directly as object saliency or object binary masks for the segmentation. For video instance segmentation, an additional mask classification head is used along with the Hungarian algorithm~\cite{kuhn1955hungarian} for bipartite matching between the sets of masks and classes~\cite{wang2021end}.

Several approaches have emerged that follow the general paradigm that was sketched in the previous paragraph. Video instance segmentation with Transformers (VisTR) presented a transformer-based query decoder aimed at decoding pixel features to represent object instances in each frame, which were termed instance level features~\cite{wang2021end}. Following a similar approach to DETR, it used instance queries learned with a transformer decoder operating on top layer features. These instance queries are then used with the top layer features to produce instance mask features at the pixel level. The instance mask features are concatenated with the top layer features and further processed with a pixel decoder to produce high resolution mask features by aggregating high resolution encoder features. Finally, the instance mask is generated from these instance features using a 3D ConvNet. Another approach, Inter-Frame Communication Transformers (IFC), used decoding instance masks with two modules: a spatial decoder and a transformer decoder~\cite{hwang2021video}. The transformer decoder uses the encoder output to learn a fixed set of query embeddings corresponding to a set of objects. The number of queries is constant per clip, while in VisTR the number of queries grows linearly with the number of frames; this innovation enables more efficient decoding compared to VisTR. Two output heads work on top of the transformer decoder: a class head and a segmentation head. The segmentation head uses a FPN-like~\cite{lin2017feature} pixel decoder to generate high resolution features before mask prediction. Another recent approach for video instance segmentation, SeqFormer, presented a query decompose transformer decoder that breaks the instance queries into frame-level box queries~\cite{wu2022seqformer}. This query decomposition mechanism allows for restricting attention to focus on the same instance across frames. A fixed number of learnable instance queries for the same instance from each frame was employed, in contrast to the frame-level instances in VisTR. The decomposed transformer decoder learns one instance query for the whole clip and one box query for each frame. SeqFormer used deformable attention in its transformer decoder~\cite{zhu2020deformable}. Since the deformable attention only attends to a small set of key sampling points, this design results in efficient and less noisy attention. On top of the decoder, it used three task specific heads: mask head, box head and class head. Yet another transformer-based universal image and video segmentation model, Mask2Former, has shown state-of-the art results on video instance segmentation by trivially generalizing the architecture for video segmentation~\cite{cheng2022masked,cheng2021mask2formervid}. In contrast to other transformer decoding used for video instance segmentation, it learned instance queries from a set of multiscale features. Operations proceed through initial use of a FPN to enrich lower layer features with higher layer semantics, followed by a transformer decoder that learns instance queries from the enriched multiscale features. The decoder building blocks follow an architecture similar to VisTR, with the modification of alternating multiple resolution feature sets in the decoder layers.

TransVOS~\cite{mei2021transvos} presented a transformer-based query decoder similar to VisTR~\cite{wang2021end} that also produces pixel mask features. It uses the query decoder to predict spatial positions of target objects in query frames for the task of semi-automatic (query frame mask guided) video object segmentation. A single object query is sufficient because semi-automatic video object segmentation aims to segment one given object in a query frame throughout the other frames in the video. Subsequently, a Target Attention Block (TAB) is used to get an attention map of the object at the query frame. This target object attention map is concatenated to the features of the frames to be segmented. A segmentation head is used to segment the target object in the current frame from these concatenated feature maps. Another approach for semi-automatic VOS, SST, used a generic convolutional segmentation network in the decoder that works on current frame features and a truncated history embedding instead of using a transformer decoder~\cite{duke2021sstvos}. An approach for automatic VOS, MED-VT, used a multiscale transformer decoder to generate object queries from three feature resolutions and applied the query on the highest resolution features to generate initial object saliency~\cite{karim2023med}. This initial proposal is used to guide final segmentation by a task head that also considers the finest features.

The multimodal transformer decoders that were introduced in the previous section \cite{botach2022end,pan2022wnet} also have corresponding decoders. Multimodal Tracking Transformer (MTTR) for text-referring video object segmentation used a transformer decoder on top of the multimodal encoder to learn object query embeddings~\cite{botach2022end}. Multiple object queries per input frame were used in this model with the decoder memory features from the encoder's multimodal features. In this approach, corresponding queries of different frames share the same weights and each query represents an embedding specific to some instance. By using instance specific shared queries across frames, this decoder achieves natural tracking of object instances. For each of the instance query outputs from the transformer decoder, a feed forward network was used to generate conditional convolution kernels~\cite{tian2020conditional,wang2021max}. Again, from the visual feature component of the multimodal output of the encoder, high resolution, semantically rich features were generated using an FPN-like~\cite{lin2017feature} spatial decoder. Then, a sequence of segmentation masks is generated by convolving the generated convolution kernels with the corresponding frame features. Finally, a text association scoring is used to determine which of the object query sets has the strongest association with the object described in the given text. Another approach for the same problem, ReferFormer, uses a transformer decoder with object queries for the referring objects conditioned on the language expression~\cite{wu2022language}. This approach of using conditional object queries simplifies the segmentation by directly decoding only the referred objects. Hence, this approach does not require object and text association on all of the objects in the scene. The transformer decoder architecture is similar to VisTR~\cite{wang2021end} with the difference of shared queries across frames. The use of shared queries allows support for videos of variable lengths. In addition, the text features are jointly fed to the decoder as input to make the query learning conditioned on the text features. Finally, three task heads were used to work on the instance embeddings and cross-modal segmentation features: class head, box head and mask head. A Cross-Modal Feature Pyramid Network (CM-FPN), adapted from the FPN~\cite{lin2017feature}, was used to generate high resolution cross-modal features. For this purpose, they added multi-head self-attention to the FPN with visual features as queries and text-features as keys and values. The mask head used the instance embeddings to generate dynamic convolution kernels~\cite{tian2020conditional} and convolve with the encoder features.

Multimodal video segmentation approach, Wnet~\cite{pan2022wnet} for audio-guided video object segmentation, uses a self-attention free transformer decoder to learn instance queries, following Fnet~\cite{lee2021fnet}. This paper argues that the joint representation of audio and video is more suitable for processing in the frequency domain. For this reason, it replaced the self-attention in the transformer decoder layer with a Fourier sublayer. A 2D Discrete Fourier Transform (DFT) is applied to the embedded input, with one dimension of the transform operating along the token dimension and the other operating along the hidden dimension. Subsequently, only the real part of the transform is used as instance queries to generate object segmentation masks.

\section{Universal video segmentation transformers}
Recently, video segmentation research has focused on designing universal segmentation models capable of handling multiple segmentation tasks in a single model. This trend is motivated by the evolution of various segmentation models toward having similar components with a difference in customized task heads and training objectives. These models typically use a  Multi-Task Learning (MTL) approach~\cite{caruana1993multitask,caruana1997multitask,zhang2021survey,vandenhende2021multi}.  MTL in the context of deep learning aims to design a universal network architecture capable of learning generalization across multiple tasks. MTL has been gaining interest in the computer vision community to improve generalization, reduce latency and allow multiple tasks to benefit from sharing complementary information. The recent advancement in artificial intelligence using transformer-based models has led to increased interest in unified modeling approaches~\cite{jaegle2021perceiver,jaegle2021perceiverio,kolesnikov2022uvim,reed2022generalist}. In particular, there has been an increased interest in unified modeling of multiple video segmentation tasks.

The recently presented TubeFormer-DeepLab~\cite{kim2022tubeformer} is a pioneering effort aimed at providing a unified model for multiple video segmentation tasks. Specifically, a model is presented to solve video semantic segmentation, instance segmentation and panoptic segmentation with a possibility for  extension to Depth-aware Video Panoptic Segmentation (DVPS). This approach formulated video segmentation as partitioning video frames into tubes with various semantic labels. In this formulation, multiple video segmentation tasks are generalized to tube labeling. The model architecture is developed by extending mask transformer~\cite{wang2021max}. The processing begins by generating a set of pairs of tube embeddings and class predictions. Next, the tube prediction is obtained from the tube embedding and pixel level video features. The generalization comes from the new formulation for directly predicting class labeled tubes. This universal architecture achieved better or competitive results across several video segmentation tasks on different benchmarks compared to specialized single or ensemble models.

Another recent approach, TarViS, presented a unified model for object detection and segmentation~\cite{athar2023tarvis}. This model is trained jointly on multiple video segmentation tasks. The model is able to `hot-swap' between tasks during inference by specifying a segmentation target without requiring task-specific retraining. The model is demonstrated on four different tasks including Video Instance Segmentation (VIS), Video Panoptic Segmentation (VPS), Video Object Segmentation (VOS) and Point Exemplar-guided Tracking (PET). In contrast to TubeFormer-DeepLab, TarViS used multiple sets of output queries instead of multiple task heads, following PerceiverIO~\cite{jaegle2021perceiver,jaegle2021perceiverio}. The main idea of TarViS is to encode a set of task specific targets as queries and then produce pixel level masks for each target using a transformer-based model. Swapping between the tasks is achieved by providing task specific targets during the inference. The segmentation masks are generated by the inner product of learned output queries and video features. In comparison to TubeFormer-DeepLab, TarViS reported competitive results on CityscapesVPS ~\cite{kim2020video} and better results on VIPSeg dataset~\cite{miao2022large}.

\section{Miscellaneous video transformers}
In addition to the above mentioned common model components and transformer-based models, we include some miscellaneous approaches in the following. 

There have been several research efforts on improving the efficiency of video transformers. One notable approach, TimeSFormer, explored various methods of spacetime separable attention~\cite{bertasius2021space}. In particular, TimeSFormer compares spacetime attention, divided spacetime attention and axial attention. In the context of video action recognition, the divided spacetime attention, where temporal attention and spatial attention are applied separately, is found to perform better compared to others. Another approach to computational overhead, Spacetime Mixing Attention, uses the full spatial region of video frames, but a temporal window that grows with layer depth, somewhat reminiscent of ConvNet design~\cite{bulat2021space}. This approach allows for the computational complexity to be independent of the temporal length of the video, because temporal window size remains constant. Yet another approach, Temporally Efficient Vision Transformer (TeViT)~\cite{yang2022temporally} for video instance segmentation, uses a messenger token shift and spatiotemporal query interaction for efficient to computation. The messenger token shift is a nearly parameter-free mechanism used for early temporal context fusion. TeViT obtains state-of-the-art results on VIS with an efficient computation model. 

Other research on transformer-based video understanding models explored various methods for dynamically sampled attention maps, rather than using global attention. A prominent method in this direction is the deformable attention mechanism~\cite{zhu2020deformable}, which was introduced in Section~\ref{sec:context_encoder}. Another transformer-based approach to video captioning, SwinBERT~\cite{lin2022swinbert}, also uses adaptive sampling to learn a sparse attention mask. It used this learnable sparse attention mask as a regularizer to improve long-range video sequence modeling with less noisy attention. To achieve this goal, the model imposes a sparsity constraint overlaid on the attention masks of the visual features. The model yielded state-of-the-art results on multiple datasets for video captioning. Although not yet applied to video segmentation, this approach of sparse attention has the potential for adaptation to address computational issues for segmenting long duration videos.


\section{Performance}\label{sec:sota_all}
In this section we document current state-of-the-art transformer performance on the various segmentation tasks identified in the previous chapter. In some cases, approaches that were described above in a component-wise fashion are reviewed here as complete systems for ease of reference. 


\subsection{Objects}\label{sec:objseg_sota}

\subsubsection{Automatic video object segmentation}
Dominant approaches to Automatic Video Object Segmentation (AVOS) rely on optical flow as an additional input along with colour images~\cite{zhou2020motion,ren2021reciprocal,siam2019video,jain2017fusionseg}. These models generally use a two-stream architecture for extracting appearance features from colour images and motion features from optical flow. There are different feature-fusing approaches available in the literature to aggregate these features, e.g., the commonly used late fusion~\cite{jain2017fusionseg} vs. recently presented multistage fusion~\cite{zhou2020motion,pei2022hierarchical}. In contrast, Treating Motion as Option TMO~\cite{cho2023treating} used optical flow but reduced the dependency on flow features by treating it as optional information and fusing it to appearance features only during decoding, if it is available. Another family of approaches considers attention to capture recurring objects without requiring optical flow features in a video via simple mechanisms, e.g., co-attention~\cite{wang2019zero,lu2019see}. Among other recent approaches, Iterative Mask Prediction (IMP) applied a Semi-automatic Video Object Segmentation (SVOS) model to AVOS~\cite{lee2022iteratively}. This approach used a Salient Object Detection (SOD) model to generate an initial pixel-wise estimate for all frames and then sampled the easy frames using a separately trained model. Finally, it propagated these sampled frame estimates to all other frames using an SVOS model. Memory based approaches also have been in use for AVOS. One such recent approach is Prototype Memory Network (PMN) that scores and samples superpixel based aggregated features as memories to be used in subsequent frames~\cite{lee2023unsupervised}. Recently, Multiscale Encoder-Decoder Video Transformer (MED-VT) presented a multiscale transformer encoder-decoder without using optical flow that used higher resolution features with high frequency information as well as lower resolution features in encoding and decoding throughout~\cite{karim2023med}. Yet another recent approach, Isomerous Transformer (Isomer)~\cite{yuan2023isomer}, proposed to use two variants of transformer: Context-Sharing Transformer (CST) for global-shared contextual information and Semantic Gathering-Scattering Transformer (SGST) for semantic correlation modeling for automatic video object segmentation. The accuracy achieved by the currently strongest performers on AVOS is summarized in Table~\ref{tab:sota_avos}. 


\begin{table*}[!ht]
  \centering
  \aboverulesep=0ex
   \belowrulesep=0ex
   \begin{adjustbox}{max width=0.75\textwidth}
    \begin{tabular}{@{}l|c|ccc@{}}
    \toprule
    \multicolumn{1}{c|}{\multirow{2}{*}{Methods}} & \multicolumn{1}{c|}{\multirow{2}{*}{Transformer-based}} &\multicolumn{3}{c}{Datasets} \\
    \multicolumn{1}{c|}{} & \multicolumn{1}{c|}{} & Davis 2016~\cite{perazzi2016benchmark}  &   YouTube Objects~\cite{prest2012learning}  & \multicolumn{1}{c}{MoCA~\cite{lamdouar2020betrayed}}     \\ 
    \midrule
    MATNet~\cite{zhou2020motion}                & -          & 82.4 & 69.0 & 64.2 \\
    RTNet~\cite{ren2021reciprocal}              & -          & 85.6 & 71.0 & 60.7 \\
    COSNet~\cite{lu2019see}                     & -          & 80.5 & 70.5 & 50.7 \\
    IMP~\cite{lee2022iteratively}               &    -       & 84.5 & -    &  -   \\    
    HFAN~\cite{pei2022hierarchical}             & \checkmark & 88.0 & 73.4 &  -   \\    
    TMO~\cite{cho2023treating}                  &  -         & 85.6 & 71.5 &  -   \\    
    PMN~\cite{lee2023unsupervised}              & \checkmark & 85.6 & 71.8 &  -   \\    
    MED-VT~\cite{karim2023med}                  & \checkmark & 85.9 & \textbf{78.5} & \textbf{77.9} \\
    Isomer~\cite{yuan2023isomer}                & \checkmark & \textbf{88.8} & - & - \\
     \bottomrule
    \end{tabular}
    \end{adjustbox}
    \caption[State of the art on automatic video object segmentation]{State of the art on automatic video object segmentation reporting mean Intersection over Union (mIoU). Best performance on each dataset is highlighted with bold font.}
    \label{tab:sota_avos}
\end{table*}

\subsubsection{Semi-automatic video object segmentation}
In general, Semi-automatic Video Object Segmentation (SVOS) approaches employ feature alignment or attention-guided mechanisms to propagate the mask of target objects (also called query objects in some contexts) from a given annotated reference frame to the entire video sequence. Matching-based approaches learn to match features of the target frames with the reference frame~\cite{hu2018videomatch}. In some approaches, the matching also uses estimates from several previous frames to segment the current frame~\cite{oh2018fast,voigtlaender2019feelvos,yang2020collaborative,yang2021collaborative}. Another research direction leverages a memory network to remember an embedding of previous predictions and uses a memory-guided attention mechanism to segment the current frame~\cite{lu2020video,oh2019video,seong2020kernelized}. One such notable approach, Hierarchical Memory Matching Network (HMMN) uses a memory-based method with two advanced memory read modules to perform memory reading at multiple scales and exploit temporal smoothness~\cite{seong2021hierarchical}. Conversely, Space-Time Correspondence Network (STCN) uses direct image-to-image correspondence for efficient and robust matching purposes~\cite{cheng2021rethinking}. Alternatively, Per-Clip Video Object Segmentation (PCVOS) presented a clip-wise mask propagation approach using an intra-clip refinement and a progressive memory matching module~\cite{park2022per}. Recent approaches also use attention-based mechanisms for SVOS to propagate masks~\cite{duke2021sstvos,lin2019agss,yang2021associating}. To further improve the performance of challenging multi-object scenarios, an attention-guided decoder also has been used~\cite{lin2019agss}. Another line of research uses a long short-term transformer for constructing hierarchical matching and propagation to associate multiple targets into a high-dimensional embedding space for simultaneous multi-object segmentation~\cite{yang2021associating}. Table~\ref{tab:sota_savos} summarizes the accuracy achieved by these current state-of-the-art approaches.

\begin{table*}[!ht]
  \centering
  \aboverulesep=0ex
   \belowrulesep=0ex
   \begin{adjustbox}{width=\textwidth}
    \begin{tabular}{l|c|ccc}
    \toprule
    \multicolumn{1}{c|}{\multirow{2}{*}{Methods}} &\multicolumn{1}{c|}{\multirow{2}{*}{Transformer-based}}& \multicolumn{3}{c}{Datasets} \\
    \multicolumn{1}{c|}{} &\multicolumn{1}{c|}{} & \multicolumn{1}{c}{ 
     YouTube-VOS 2018~\cite{xu2018youtube}(unseen)}  &   \multicolumn{1}{c}{YouTube-VOS 2019~\cite{xu2018youtube} (unseen)}  & \multicolumn{1}{c}{DAVIS 2017~\cite{pont20172017,caelles20182018} (test split)}     \\ 
    \midrule
    KMN~\cite{seong2020kernelized}      &    -    & 75.3      &   -       &  74.1    \\
    HMMN~\cite{seong2021hierarchical}   &     -    & 76.8      &   77.3    & 81.9 \\
    STCN~\cite{cheng2021rethinking}     &     -    & 77.9      &   78.2    & 82.2 \\ 
    SST~\cite{duke2021sstvos}           &\checkmark& 76.0      &   76.6    &  -    \\
    CFBI+~\cite{yang2021collaborative}  &\checkmark& 77.1      &   77.1    & 74.4 \\
    AOT~\cite{yang2021associating}      &\checkmark& 77.9      &   78.4    & 77.3 \\
    PCVOS~\cite{park2022per}            &\checkmark& \textbf{79.6} & \textbf{80.0} & \textbf{83.0} \\
     \bottomrule
    \end{tabular}
    \end{adjustbox}
    \caption[State of the art on semi-automatic video object segmentation]{State of the art on semi-automatic video object segmentation reporting mean Intersection over Union (mIoU). Best performance on each dataset is highlighted with bold font.}
    \label{tab:sota_savos}
\end{table*}

\subsubsection{Interactive video object segmentation}
Recent Interactive Video Object Segmentation (IVOS) approaches typically work by following an interaction-propagation approach~\cite{heo2020interactive, miao2020memory, heo2021guided, cheng2021modular}. Interaction-and-Propagation Networks (IPNet) introduced an approach with two internally connected networks jointly trained for user interaction processing and transferring the annotations to neighboring frames~\cite{oh2019fast}. Another approach, the Memory Aggregation Networks (MANet), used integrated interaction and propagation in a single network to improve efficiency~\cite{miao2020memory}. An effective memory aggregation mechanism was used in this method to record the informative knowledge from the previous interaction rounds. Later, Annotation Transfer Network (AT-Net), used an Annotation Network (A-Net) and a Transfer Network (T-Net) that alternate to generate the segmentation results~\cite{heo2020interactive}. The A-Net first generates a segmentation result from user scribbles on a frame and then the T-Net propagates the segmentation result bidirectionally to the other frames by employing global and local transfer modules. Yet another approach, Guided Interactive Segmentation (GIS), presented a reliability-based attention module and intersection-aware propagation module~\cite{heo2021guided}. Conversely, Modular interactive Video Object Segmentation (MiVOS), decouples the interaction-propagation as interaction-to-mask and mask propagation, where an interaction module converts user interactions to an object mask and then the mask is temporally propagated~\cite{cheng2021modular}. For efficient temporal mask propagation, a spacetime memory model is employed in this method with a novel top-k filtering strategy to read the memory. Table ~\ref{tab:sota_iavos} summarizes the accuracy achieved by these current state-of-the-art approaches to interactive VOS.

\begin{table*}[!ht]
  \centering
  \aboverulesep=0ex
   \belowrulesep=0ex
   \begin{adjustbox}{max width=0.70\textwidth}
    \begin{tabular}{@{}l|c|c@{}}
    \toprule
    \multicolumn{1}{c|}{\multirow{2}{*}{Methods}} &\multicolumn{1}{c|}{\multirow{2}{*}{Transformer-based}}& \multicolumn{1}{c}{Datasets} \\
    \multicolumn{1}{c|}{}&\multicolumn{1}{c|}{} & \multicolumn{1}{c}{DAVIS 2017(interactive/val)~\cite{pont20172017,caelles20182018,caelles20192019}} \\ 
    \midrule
    IPNet~\cite{oh2019fast}            & - & 73.4 \\
    MANet~\cite{miao2020memory}        & - & 76.1 \\
    ATNet~\cite{heo2020interactive}    & - & 79.0 \\
    GIS~\cite{heo2021guided}           & - & 82.9 \\ 
    MiVOS~\cite{cheng2021modular}      & - & \textbf{85.4} \\
     \bottomrule
    \end{tabular}
    \end{adjustbox}
    \caption[State of the art on interactive video object segmentation]{State of the art on interactive video object segmentation reporting mean intersection over union ($J@60$ score). To date, no transformer models have appeared for this task. Best performance on each dataset is highlighted with bold font.}
    \label{tab:sota_iavos}
\end{table*}

\subsubsection{Video instance segmentation}
Recent dominant approaches for Video Instance Segmentation (VIS) follow a simultaneous segmentation and tracking approach using transformer-based attention mechanisms~\cite{wang2021end,hwang2021video,wu2022seqformer}. 
Crossover learning for Video Instance Segmentation (CrossVIS), used a crossover learning scheme for online VIS that localizes an instance in other frames from the instance feature from the current frame~\cite{yang2021crossover}. This approach also used an additional branch, called global balanced instance embedding, for a stable online instance association. Another approach, Video instance segmentation with Transformers (VisTR), used a transformer-based encoder-decoder approach followed by multiple task-heads for detection, segmentation and matching-based tracking to work on an entire video sequence in a single inference~\cite{wang2021end}. This formulation allows for a simple yet efficient video instance segmentation model. Yet another approach, Inter-Frame Communication Transformers (IFC), used a concise memory token representation as an efficient encoding of the clip context~\cite{hwang2021video}. The inter-frame tracking is achieved via an exchange of information between these memory tokens. Conversely, Sequence Mask R-CNN (Seq Mask R-CNN), presented a propose-reduce paradigm to generate prediction of a whole clip in a single step~\cite{lin2021video}.

Recently, Efficient Video Instance Segmentation (EfficientVIS), presented a fully end-to-end solution that achieves whole video instance segmentation in a single pass without requiring additional data association and post-processing~\cite{wu2022efficient}. To address computational complexity, Temporally efficient Vision Transformer (TeViT), uses an efficient message passing based frame-level temporal context and query-based instance-level temporal context to achieve strong temporal consistent modeling capacity on the VIS task~\cite{yang2022temporally}. Another approach addressing computational complexity, Sequential transFormer (SeqFormer)~\cite{wu2022seqformer}, used deformable attention~\cite{zhu2020deformable} in the decoder to first locate the object bounding box queries in each frame and then aggregated to the instance queries to predict dynamic mask head parameters. The tracking is achieved by convolving the encoded feature maps by using a mask head. Another interesting approach, Tubelet Transformer DeepLab (TubeFormer-DeepLab) demonstrated a transformer-based unified model for multiple video segmentation tasks including video semantic, instance and panoptic segmentation~\cite{kim2022tubeformer}. Similarly, video Kernel Network (K-Net), also used a video panoptic segmentation model equipped with a dynamic kernel for video instance segmentation as a subtask~\cite{li2022video}.

Another research direction formulated the problem not to use annotation during training; in particular, FreeSOLO~\cite{wang2022freesolo} presented a simple yet effective solution to learn instance segmentation following a contrastive learning approach~\cite{wang2021dense}. Conversely, motivated by online problem settings, In Defense of OnLine models (IDOL) is a transformer-based approach to VIS that used contrastive learning to learn more discriminative and robust instance embeddings for frame-to-frame association~\cite{wu2022defense}. Another approach, Video Mask Transfiner (VMT), introduced a method to leverage the high resolution temporal features for temporal mask refinement by using an efficient video transformer~\cite{ke2022video}. Following similar motivation, Multi-Scale SpatioTemporal Split attention transformer (MS-STS) introduced a multi-scale spatiotemporal split attention module to capture spatiotemporal feature relationships across multiple scales for video instance segmentation~\cite{omkar2022vis}.

Very recently, Instance Motion for object-centric video segmentation (InstMove) introduced a method to leverage  instance-level motion information by using a memory network that can be integrated into existing models to boost the performance of various video segmentation tasks including video instance segmentation~\cite{liu2023instmove}. Conversely, Generalized Video Instance Segmentation (GenVIS) introduced a method for unified video label assignment using a framework to sequentially associate clip-wise predictions \cite{heo2023generalized}.  Another approach, Context-Aware Relative Object Queries (CAROQ) introduced an approach to continuously propagated object queries frame-by-frame to track objects without post-processing~\cite{choudhuri2023context}.  Yet another approach in the open-world setting for segmenting unfamiliar objects, Cut-and-LEaRn (CutLER), used a cut-and-learn pipeline to discover objects~\cite{wang2023cut}. Another unified video segmentation model, TarViS~\cite{athar2023tarvis}  utilized transformer based queries for video instance segmentation. Conversely, Decoupled Video Instance Segmentation Framework (DVIS)~\cite{zhang2023dvis} proposed a decoupling strategy for VIS by dividing it into three independent sub-tasks of segmentation, tracking and refinement. Another recent approach, Temporally Consistent Online Video Instance Segmentation (TCOVIS)~\cite{li2023tcovis}, presented a novel online strategy for segmenting video instances, emphasizing the optimal use of temporal data within video clips. The approach prioritizes a global strategy for assigning instances and a module for enhancing spatiotemporal properties to improve the coherence of temporal features. Yet another recent approach, Consistent Training for Online Video Instance Segmentation (CTVIS)~\cite{ying2023ctvis}, presented a simple yet effective training strategy termed consistent training, which is based on aligning the training and inference pipelines to construct contrastive items. Another universal video segmentation approach, Tube-Link~\cite{li2023tube}, also presented video instance segmentation along with several other video segmentation tasks following a temporal contrastive learning approach to build instance-wise discriminative features for tube level association. Table ~\ref{tab:sota_vis} summarizes the accuracy achieved by these currently state-of-the-art approaches.


\begin{table*}[!ht]
  \centering
  \aboverulesep=0ex
   \belowrulesep=0ex
   \begin{adjustbox}{width=\textwidth}
    \begin{tabular}{@{}l|c|cccc@{}}
    \toprule
    \multicolumn{1}{c|}{\multirow{2}{*}{Methods}}&\multicolumn{1}{c|}{\multirow{2}{*}{Transformer-based}} & \multicolumn{3}{c}{Datasets} \\
    \multicolumn{1}{c|}{}&\multicolumn{1}{c|}{}& \multicolumn{1}{c}{YouTube-VIS-2019~\cite{yang2019video}}  &  \multicolumn{1}{c}{YouTube-VIS-2021~\cite{vis2021}}  & \multicolumn{1}{c}{OVIS~\cite{qi2022occluded}} & \multicolumn{1}{c}{UVO~\cite{wang2021unidentified}}    \\ 
    \midrule
    CrossVIS~\cite{yang2021crossover}           &      -     & 36.6 & 34.2 & 14.9 &   -  \\
    VisTR~\cite{wang2021end}                    & \checkmark & 40.1 &   -  &   -  &   -  \\
    IFC~\cite{hwang2021video}                   & \checkmark & 42.6 & 35.2 &   -  &   -  \\
    Seq Mask R-CNN~\cite{lin2021video}          &      -     & 47.6 &  -   &   -  &   -  \\  
    EfficientVIS~\cite{wu2022efficient}         & \checkmark & 39.8 &   -  &   -  &   -  \\
    TeViT~\cite{yang2022temporally}             & \checkmark & 46.6 & 37.9 & 17.4 &   -  \\
    SeqFormer~\cite{wu2022seqformer}            & \checkmark & 59.3 &   -  &   -  &   -  \\
    TubeFormer-DeepLab~\cite{kim2022tubeformer} & \checkmark & 47.5 & 41.2 &   -  &   -  \\
    Video K-Net~\cite{li2022video}              & \checkmark & 51.4 &  -   &   -  &   -  \\ 
    FreeSOLO~\cite{wang2022freesolo}            &     -      &  -   &   -  &   -  &  4.8 \\    
    IDOL~\cite{wu2022defense}                   & \checkmark & 62.2 & 56.1 & 42.6 &   -  \\
    VMT~\cite{ke2022video}                      & \checkmark & 59.7 &  -   & 19.8 &   -  \\
    MS-STS VIS~\cite{omkar2022vis}              & \checkmark & 61.0 &  -   &   -  &   -  \\
    InstMove~\cite{liu2023instmove}             & \checkmark &  -   &  -   & 30.7 &   -  \\ 
    GenVIS~\cite{heo2023generalized}            & \checkmark & 64.0 & 59.6 & 45.4 &   -  \\
    CAROQ~\cite{choudhuri2023context}           & \checkmark & 61.4 & 54.5 & 38.2 &   -  \\
    CutLER~\cite{wang2023cut}                   & \checkmark &   -  &   -  &   -  & \textbf{10.1} \\ 
    TarViS~\cite{athar2023tarvis}               & \checkmark &   -  & 60.2 & 43.2 &   -  \\ 
    DVIS~\cite{zhang2023dvis}                   & \checkmark & 64.9 & 60.1 & \textbf{49.9} &  -   \\ 
    TCOVIS~\cite{li2023tcovis}                  & \checkmark & 64.1 & \textbf{61.3} & 46.7 & - \\
    CTVIS~\cite{ying2023ctvis}                  & \checkmark & \textbf{65.6} & 61.2 & 46.9 & - \\
    Tube-Link~\cite{li2023tube}                 & \checkmark & 64.6 & 58.4   & - &    -    \\       
    \bottomrule
    \end{tabular}
    \end{adjustbox}
    \caption[State of the art on video instance segmentation]{State of the art on video instance segmentation reporting average precision (AP). Best performance on each dataset is highlighted with bold font.}
    \label{tab:sota_vis}
\end{table*}

\subsection{Actors and Actions}\label{sec:actorseg_sota}
\subsubsection{Actor-action segmentation}
Trending research on actor-action segmentation typically include 2D or 3D ConvNets as feature backbones and use optical flow as an extra input along with colour (RGB) frames to localize object motion as a probable actor~\cite{ji2018end,dang2018actor,rana2021we}. Some approaches formulate the model as simultaneous multi-task learning by adding multiple branches of task heads on top of a feature backbone~\cite{ji2018end}. Specifically, separate prediction branches are used for mask segmentation, actor classification and action classification on top of fused features from RGB frames and optical flow inputs. Another notable approach uses a region of interest-based approach with region masks from an instance segmentation algorithm followed by non-maximum suppression and a classification head~\cite {dang2018actor}. Recently, Single Shot Actor-Action Detection (SSA2D) introduced a simple yet effective proposal-free model using 3D convolutions to perform pixel-level joint actor-action detection in a single shot~\cite{rana2021we}. Yet another recent approach, Multiscale Encoder-Decoder Video Transformer (MED-VT), achieved state-of-the-art results on actor-action segmentation using a multiscale transformer encoder-decoder model without relying on optical flow~\cite{karim2023med}. Table ~\ref{tab:sota_aaseg} summarizes the accuracy achieved by these approaches.

\begin{table*}[!ht]
  \centering
  \aboverulesep=0ex
   \belowrulesep=0ex
   \begin{adjustbox}{width=0.55\textwidth}
    \begin{tabular}{@{}l|c|c@{}}
    \toprule
    \multicolumn{1}{c|}{\multirow{2}{*}{Methods}}&\multicolumn{1}{c|}{\multirow{2}{*}{Transformer-based}} & \multicolumn{1}{c}{Datasets} \\
    \multicolumn{1}{c|}{}&\multicolumn{1}{c|}{}& \multicolumn{1}{c}{ Actor-Action Dataset (A2D)~\cite{xu2015can} }   \\ 
    \midrule
      Ji et al.~\cite{ji2018end}        &      -     & 36.9 \\
      Dang et al.~\cite{dang2018actor}  &      -     & 38.6 \\
      SSA2D~\cite{rana2021we}           &      -     & 39.5 \\
      MED-VT~\cite{karim2023med}        & \checkmark & \textbf{52.6} \\ 
    \bottomrule
    \end{tabular}
    \end{adjustbox}
    \caption[State of the art on video actor-action segmentation]{State of the art on video actor-action segmentation reporting mean intersection over union on joint actor-action segmentation. Best performance is highlighted with bold font.}
    \label{tab:sota_aaseg}
\end{table*}

\subsection{Scenes}\label{sec:sceneseg_sota}

\subsubsection{Video semantic segmentation}
Most common approaches in Video Semantic Segmentation (VSS), also referred to as Video Scene Parsing (VSP), follow propagation based  segmentation from a key-frame (semi-automatic) or previous frames (automatic) to other frames. Typically, these approaches rely on optical flow based warping to propagate their labels~\cite{gadde2017semantic,liu2017surveillance,ding2020every,jain2019accel}. However, trending research uses inter-frame attention~\cite{paul2020efficient,wang2021temporal,li2021video} to capture the temporal relations between consecutive frames without depending on optical flow. For example, Temporal Memory Attention Network (TMANet) adaptively integrates the long-range temporal relations over a video sequence with a temporal memory attention module to capture the relation between the current frame and a memory generated from several past frames~\cite{wang2021temporal}. 
To further address efficiency, Sparse Temporal Transformer (STT) used sparse inter-frame cross-attention on a query and key selection strategy to reduce the time complexity of attentional processing~\cite{li2021video}. Another approach, Temporal Context Blending (TCB), introduced a model to effectively harness long-range contextual information across video frames using a spatiotemporal pyramid pooling mechanism~\cite{miao2021vspw}. Conversely, SegFormer, presented a hierarchically structured transformer encoder, named Mix Transformer (MiT), that operates without using positional embedding coupled with a simple decoder~\cite{xie2021segformer}. Another approach, video Kernel Network (K-Net), used a video panoptic segmentation model equipped with a dynamic kernel for video semantic segmentation as a subtask~\cite{li2022video}. As yet another approach, Coarse-to-Fine Feature Mining (CFFM) used inter-frame cross-attention with coarse features on past frames for efficient cross-attention~\cite{sun2022coarse}. An interesting unified segmentation model, Tubelet Transformer DeepLab (TubeFormer-DeepLab), also demonstrated its unified video segmentation model on video semantic segmentation~\cite{kim2022tubeformer}. Recently, another universal video segmentation approach, Tube-Link~\cite{li2023tube}, also presented video semantic segmentation following a temporal  cross-tube matching approach. Table ~\ref{tab:sota_vss} summarizes the accuracy achieved by these approaches. 


\begin{table*}[!ht]
  \centering
  \aboverulesep=0ex
   \belowrulesep=0ex
   \begin{adjustbox}{width=0.90\textwidth}
    \begin{tabular}{@{}l|c|ccc@{}}
    \toprule
    \multicolumn{1}{c|}{\multirow{2}{*}{Methods}}&\multicolumn{1}{c|}{\multirow{2}{*}{Transformer-based}} & \multicolumn{3}{c}{Datasets} \\
    \multicolumn{1}{c|}{}&\multicolumn{1}{c|}{}& \multicolumn{1}{c}{ VSPW~\cite{miao2021vspw} }  &  \multicolumn{1}{c}{ Cityscapes~\cite{kim2020video} }  & \multicolumn{1}{c}{Camvid~\cite{brostow2009semantic}}     \\ 
    \midrule
     TMANet~\cite{wang2021temporal}             & \checkmark &  -    & 80.3  & 76.5 \\
     STT~\cite{li2021video}                     & \checkmark &  -    & 82.5  & \textbf{80.2} \\
     TCB~\cite{miao2021vspw}                    &      -     & 37.8  &    -  & - \\
     SegFormer~\cite{xie2021segformer}          & \checkmark & 48.2  & \textbf{84.0}  & - \\
     Video K-Net~\cite{li2022video}             & \checkmark & 38.0  &   -   & - \\   
     CFFM~\cite{sun2022coarse}                  & \checkmark & 49.3  & 75.1  & -  \\
     TubeFormer-DeepLab~\cite{kim2022tubeformer} & \checkmark & \textbf{63.2} &   -   & - \\
     Tube-Link~\cite{li2023tube}                 & \checkmark & 59.7 &  -   & - \\
    \bottomrule
    \end{tabular}
    \end{adjustbox}
    \caption[State of the art on video semantic segmentation]{State of the art on video semantic segmentation reporting mean intersection over union (mIoU). Best performance on each dataset is highlighted with bold font.}
    \label{tab:sota_vss}
\end{table*}

\subsubsection{Video panoptic segmentation}
Recent efforts on Video Panoptic Segmentation (VPS) use multiple task head branches for detection, segmentation and tracking. The Video Panoptic Segmentation Network (VPSNet) initially aggregates warped features of past frames to the current frame, performs spatiotemporal attention followed by multi-task branches and then explicitly models cross-frame instance association specifically for tracking~\cite{kim2020video}. Another approach, Siamese model with Tracking head (SiamTrack), used a similar architecture and jointly learned coarse segment-level matching and fine pixel-level matching for temporal associations using a supervised contrastive learning approach~\cite{woo2021learning}. Yet another approach, video Kernel Network (K-Net), introduced a content-aware dynamic kernel with a bipartite matching strategy~\cite{li2022video}. Video inverse Projection DeepLab (ViP-DeepLab) for depth-aware video panoptic segmentation jointly performs video panoptic segmentation and monocular depth estimation as two subtasks of the main problem using a unified model~\cite{qiao2021vip}. This model achieved state-of-the-art results on the task of video panoptic segmentation as the first subtask of depth-aware video panoptic segmentation. Another unified multitask model, Tubelet Transformer DeepLab (TubeFormer-DeepLab), presented a mask transformer-based approach to partition a video clip into class-labeled tubes that yielded a general solution to Video Semantic Segmentation (VSS), Video Instance Segmentation (VIS) and Video Panoptic Segmentation (VPS)~\cite{kim2022tubeformer}. 

Yet another approach, panoptic Slots for VPS (Slot-VPS), presented an end-to-end framework with an object-centric representation learning approach for video panoptic segmentation~\cite{zhou2022slot}. It encoded all panoptic entities in a video with a unified representation, called panoptic slots, that capture their visual appearance and spatial information across time. It used a video panoptic retriever to retrieve and encode coherent spatiotemporal object information into the panoptic slots. The retriever is a three-stage pipeline that includes slot assignment, slot prediction and slot refinement. The approach used three prediction heads to produce classification, mask and object ID from the panoptic slots. A recent video instance segmentation approach, Context-Aware Relative Object Queries (CAROQ), also demonstrated the effectiveness of object query propagation for panoptic video segmentation~\cite{choudhuri2023context}. Table ~\ref{tab:sota_vps} summarizes the accuracy achieved by these currently state-of-the-art approaches.

\begin{table*}[!ht]
  \centering
  \aboverulesep=0ex
   \belowrulesep=0ex
   \begin{adjustbox}{width=\textwidth}
    \begin{tabular}{@{}l|c|ccc@{}}
    \toprule
    \multicolumn{1}{c|}{\multirow{2}{*}{Methods}} &\multicolumn{1}{c|}{\multirow{2}{*}{Transformer-based}}& \multicolumn{3}{c}{Datasets} \\
    \multicolumn{1}{c|}{}&\multicolumn{1}{c|}{}& \multicolumn{1}{c}{ Cityscapes-VPS~\cite{kim2020video} (val) }  &  \multicolumn{1}{c}{ VIPER~\cite{richter2017playing} (val) }  &   \multicolumn{1}{c}{KITTI-STEP~\cite{weber2021step}}\\ 
    \midrule
    VPSNet~\cite{kim2020video}                  &     -      & 56.1 & 51.9 &   -   \\
    SiamTrack~\cite{woo2021learning}            &     -      & 57.3 & 50.2 &  -    \\
    Video K-Net~\cite{li2022video}              & \checkmark & 62.2 &   -  & 63.0  \\ 
    ViP-Deeplab~\cite{qiao2021vip}              &      -     & 63.1 &   -  &    -  \\
    TubeFormer-DeepLab~\cite{kim2022tubeformer} & \checkmark &   -  &   -  & \textbf{65.25} \\
    Slot-VPS~\cite{zhou2022slot}                & \checkmark & \textbf{63.7} & \textbf{56.2} &    -  \\
    CAROQ~\cite{choudhuri2023context}           & \checkmark & 63.0 &   -  &    -  \\
    TarViS~\cite{athar2023tarvis}               & \checkmark & 58.9 &   -  &   72.0  \\
    \bottomrule
    \end{tabular}
    \end{adjustbox}
    \caption[State of the art on video panoptic segmentation]{State of the art on video panoptic segmentation reporting Video Panoptic Quality (VPQ) on Cityscapes-VPS (validation set)~\cite{kim2020video}, VPQ on VIPER~\cite{richter2017playing} and (Segmentation and Tracking Quality) STQ on KITTI-STEP (test set)~\cite{weber2021step}. Best performance on each dataset is highlighted with bold font.}
    \label{tab:sota_vps}
\end{table*}

\subsubsection{Depth-aware video panoptic segmentation} 
Recent efforts on Depth-aware Video Panoptic Segmentation (DVPS) mostly extended non-depth aware panoptic segmentation approaches with an extra depth estimation head. One notable approach, ViP-DeepLab was the first presented model for DVPS that used an additional prediction head for monocular depth estimation~\cite{qiao2021vip}. Subsequently, TubeFormer-DeepLab presented a unified model to combine multiple video segmentation tasks in a single model, which thereby went beyond just DVPS~\cite{kim2022tubeformer}. The approach begins by partitioning a video clip into tubes with class prediction and tube embedding vector pairs that later are used to generate a tube prediction. For this purpose, hierarchical structure is employed, with two types of dual-path transformer blocks using message passing between video features and a latent memory or a global memory. Yet another recent approach, PolyphonicFormer presented a transformer-based query learning approach for DVPS that unifies both panoptic segmentation and depth estimation~\cite{yuan2022polyphonicformer}. This unified modeling enables interaction between panoptic context and depth information allowing them each to benefit from the other. Overall, the model has three components. First, a feature extractor generates features for panoptic segmentation and depth estimation. Second, a polyphonic head learns the queries for these tasks. Third, a tracking head is used to learn the feature embedding among the frames. Another recent approach, Bi-directional Guidance Learning (BGL)~\cite{he2023towards}, presented a framework that integrates depth information into panoptic segmentation by simultaneously conducting segmentation and depth estimation for each segment using shared object queries. It employs a geometric query enhancement method and bi-directional guidance learning to enhance the relationship between the tasks. Table ~\ref{tab:sota_dvpq} summarizes the accuracy achieved by these currently state-of-the-art approaches.

\begin{table*}[!ht]
  \centering
  \aboverulesep=0ex
   \belowrulesep=0ex
   \begin{adjustbox}{max width=0.90\textwidth}
    \begin{tabular}{@{}l|c|ccc@{}}
    \toprule
    \multicolumn{1}{c|}{\multirow{2}{*}{Methods}}&\multicolumn{1}{c|}{\multirow{2}{*}{Transformer-based}} & \multicolumn{2}{c}{Datasets} \\
    \multicolumn{1}{c|}{}&\multicolumn{1}{c|}{}& \multicolumn{1}{c}{  Cityscapes-DVPS~\cite{kim2020video,qiao2021vip} }  &  \multicolumn{1}{c}{  SemKITTI-DVPS~\cite{behley2019semantickitti,qiao2021vip} }    \\ 
    \midrule
        ViP-DeepLab~\cite{qiao2021vip}                   &       -    &      55.1       &  45.6/63.36 \\
        TubeFormer-DeepLab~\cite{kim2022tubeformer}      & \checkmark &     -           & -/\textbf{67.0} \\
        PolyphonicFormer~\cite{yuan2022polyphonicformer} & \checkmark &     55.4        & 46.4/-   \\
        BGL~\cite{he2023towards}                         & \checkmark & \textbf{64.3}   & \textbf{47.9}/-      \\
    \bottomrule
    \end{tabular}
    \end{adjustbox}
    \caption[State of the art on depth-aware video panoptic segmentation]{State of the art on depth-aware video panoptic segmentation using Depth-aware Video Panoptic Quality (DVPQ) and Depth-aware Segmentation and Tracking Quality (DSTQ) metrics.  Results are reported for DVPQ on Cityscapes-DVPS~\cite{kim2020video,qiao2021vip} and DVPQ/DSTQ on SemKITTI-DVPS~\cite{behley2019semantickitti,qiao2021vip}. Best performance on each dataset is highlighted with bold font.}
    \label{tab:sota_dvpq}
\end{table*}

\subsubsection{Panoramic video panoptic segmentation} 
Currently, a single approach to Panoramic Video Panoptic Segmentation (PVPS) has been appeared~\cite{mei2022waymo}. It operates by augmenting the ViP-DeepLab video panoptic segmentation algorithm to accept multi-camera input~\cite{qiao2021vip}. For this purpose, it trains separate models for each camera view and an additional model on a panorama generated from all views. To track instance IDs across camera views, a similar method to that used in ViP-DeepLab to track instance IDs across time is followed. This instance ID tracking method is based on comparing IoU between region pairs, e.g., two masks with high IoU overlap are re-assigned with the same instance ID. Table ~\ref{tab:sota_pvps} summarizes the accuracy achieved by this approach.

\begin{table*}[!ht]
  \centering
  \aboverulesep=0ex
   \belowrulesep=0ex
   \begin{adjustbox}{width=0.65\textwidth}
    \begin{tabular}{@{}l|c|ccc@{}}
    \toprule
    \multicolumn{1}{c|}{\multirow{2}{*}{Methods}}&\multicolumn{1}{c|}{\multirow{2}{*}{Transformer-based}} & \multicolumn{1}{c}{Datasets} \\
    \multicolumn{1}{c|}{}&\multicolumn{1}{c|}{}& \multicolumn{1}{c}{ WOD:PVPS~\cite{mei2022waymo} }     \\ 
    \midrule
    ViP-DeepLab+~\cite{cheng2020panoptic,qiao2021vip, mei2022waymo}   & - &  \textbf{39.78}  \\
    \bottomrule
    \end{tabular}
    \end{adjustbox}
    \caption[State of the art on panoramic video panoptic segmentation]{State of the art on panoramic video panoptic segmentation reporting weighted Segmentation and Tracking Quality (wSTQ). To date, no transformer models have appeared for this task. Indeed, only a single approach has appeared and its accuracy is highlighted in bold.}
    \label{tab:sota_pvps}
\end{table*}

\subsection{Multimodal video segmentation}\label{sec:mmseg_sota}

\subsubsection{Text-guided video object segmentation}
Early efforts on Text Guided VOS (TGVOS)/Referring-VOS (RVOS) generally follow a two stage pipeline: First, a referring bounding box is tracked across a video; second, the box is segmented to indicate the object of interest on a pixel-wise basis~\cite{khoreva2018video}. A notable recent approach, Unified Referring Video Object Segmentation (URVOS), presented an efficient and unified model to initially segment some masks guided by the referring expressions, followed by a propagation based semi-supervised VOS to produce the final results~\cite{seo2020urvos}. Another approach, You Only inFer Once (YOFO), presented a multiscale cross-modal feature mining structure and an augmented memory based meta-transfer mechanism that learns target specific features~\cite{li2022you}. Recently, transformer-based models have been introduced for RVOS that use an encoder-decoder architecture. One such approach, Multimodal Tracking TransformeR (MTTR), initially extracts features from the text and video for subsequent concatenated multimodal feature sequences~\cite{botach2022end}. Then, a  multimodal transformer encoder is used as feature relation encoder followed by a transformer decoder to decode instance-level queries. Finally, corresponding mask and reference labels are generated. Another transformer-based approach, ReferFormer, presented a simple and unified end-to-end transformer-based framework that uses object queries conditioned on language expressions~\cite{wu2022language}. Overall, the model has four components. First, a backbone feature extractor with a visual encoder and a text encoder are used to extract visual features and text features. Second, a transformer encoder-decoder is applied: The encoder operates on the visual features to encode their relations; the decoder uses both visual and text features to generate shared object queries that are converted into dynamic kernels. Third, high resolution visual features are generated by a cross-modal feature pyramid network~\cite{ghiasi2019fpn} that facilitates multi-scale vision-language fusion. Fourth, the segmentation component uses the dynamic kernels from the transformer decoder as convolution kernels to extract the segmentation masks from the high resolution feature maps. The tracking of objects across frames is achieved by linking the object queries across frames. Another recent work, Hybrid Temporal-scale Multimodal Learning (HTML)~\cite{han2023html} framework employs a multi-scale strategy to synchronize language and visual features for recognizing crucial object meanings in videos. This goal is achieved through hierarchical learning of multimodal interactions across various timeframes, supported by an innovative inter-scale multimodal perception module. The primary objective of this module is to facilitate dynamic communication between language queries and visual features, thereby minimizing object confusion by integrating video context across different scales. Conversely, Temporal Collection and Distribution (TempCD)~\cite{tang2023temporal}, presented a referent-guided dynamic temporal decoding for text-referring video object segmentation. For this purpose, a global referent token and a sequence of object queries are used to capture language guided video-level referent and high quality object localization, respectively. The model also introduced a temporal collection-distribution mechanism to capture object motions and spatial-temporal cross-modal reasoning for efficient reasoning between the referent sequence and object queries in every frame. Yet another approach, Spectrum-guided Multigranularity (SgMg)~\cite{miao2023spectrum}, proposed segmenting encoded features and leveraging visual details to enhance mask optimization. Additionally, this approach used a Spectrum-guided Cross-modal Fusion (SCF) to foster effective multimodal representation by enabling global interactions within frames in the spectral domain. This approach was extended to multi-object R-VOS, facilitating simultaneous segmentation of multiple referred objects in a video. Table ~\ref{tab:sota_tgvos} summarizes the accuracy achieved by these currently state-of-the-art approaches.

\begin{table*}[!ht]
  \centering
  \aboverulesep=0ex
   \belowrulesep=0ex
   \begin{adjustbox}{width=0.90\textwidth}
    \begin{tabular}{@{}l|c|ccc@{}}
    \toprule
    \multicolumn{1}{c|}{\multirow{2}{*}{Methods}}&\multicolumn{1}{c|}{\multirow{2}{*}{Transformer-based}} & \multicolumn{3}{c}{Datasets} \\
    \multicolumn{1}{c|}{}&\multicolumn{1}{c|}{}& \multicolumn{1}{c}{ Ref-Youtube-VOS~\cite{seo2020urvos} (val)}  &  \multicolumn{1}{c}{ Ref-DAVIS17~\cite{khoreva2019video} (val)}  & \multicolumn{1}{c}{A2D-Sentences~\cite{gavrilyuk2018actor}  }     \\ 
    \midrule
    URVOS~\cite{seo2020urvos}         & \checkmark &  45.3 & 47.3 &  - \\
    YOFO~\cite{li2022you}             & \checkmark &  49.7 & 52.2 & - \\
    MTTR~\cite{botach2022end}         & \checkmark &  54.0 &   -  & 64.0  \\
    ReferFormer~\cite{wu2022language} & \checkmark &  62.8 & 58.1 & 70.3  \\
    HTML~\cite{han2023html}           & \checkmark &  61.5 & 59.2 & 71.2  \\
    TempCD~\cite{tang2023temporal}    & \checkmark &  63.6 & \textbf{61.6} & \textbf{76.6}  \\
    SgMg~\cite{miao2023spectrum}      & \checkmark &  \textbf{63.9} & 60.6 & 72.0  \\
    \bottomrule
    \end{tabular}
    \end{adjustbox}
    \caption[State of the art on multimodal text-guided video object segmentation]{State of the art on multimodal text-guided video object segmentation reporting mean intersection over union ($J$). Best performance on each dataset is highlighted with bold font.}
    \label{tab:sota_tgvos}
\end{table*}

\subsubsection{Audio-guided video object segmentation}
Currently, it appears that there is only a single effort that has considered  Audio Guided VOS (AGVOS). In particular, Wavelet-based cross-modal denoising network (Wnet), used a cross-modal denoising approach to clean the audio signal and a wavelet-based encoder network to learn the cross-modal representations of the target video and query audio~\cite{pan2022wnet}. Overall, the model is composed of five modules: visual encoder, audio encoder, transformer encoder, transformer decoder and segmentation module. The visual encoder is a ConvNet (ResNet-50~\cite{he2016deep}) that is used to extract features from the input video clip. The audio encoder uses a pretrained acoustic feature extractor network ~\cite{bouchakour2018mfccs}, followed by convolution. Following feature extraction, a transformer encoder is used to encode the cross-modal features. Here, a wavelet based cross-modal module is used to fuse the multimodal features in a denoised joint representation. Next, a transformer decoder is used to learn object instance queries. Finally, the queries are used together with the cross-modal features to segment instances with a segmentation head. Since there was no directly comparable method, the authors extended three video only models to also encompass audio for the sake of comparison~\cite{pan2022wnet}. Table ~\ref{tab:sota_agvos} summarizes the accuracy of Wnet and the comparison algorithms.

\begin{table*}[!ht]
  \centering
  \aboverulesep=0ex
   \belowrulesep=0ex
   \begin{adjustbox}{max width=0.50\textwidth}
    \begin{tabular}{@{}l|c|c@{}}
    \toprule
    \multicolumn{1}{c|}{\multirow{2}{*}{Methods}}&\multicolumn{1}{c|}{\multirow{2}{*}{Transformer-based}}  & \multicolumn{1}{c}{Datasets} \\
    \multicolumn{1}{c|}{}&\multicolumn{1}{c|}{}& \multicolumn{1}{c}{ AGVOS~\cite{pan2022wnet} } \\ 
    \midrule
     URVOS+\textdagger~\cite{seo2020urvos}    &  \checkmark   & 37.1 \\
     PAM+~\textdagger~\cite{ning2020polar}    &  \checkmark   & 38.6 \\
     VisTR+\textdagger~\cite{wang2021end}     &  \checkmark   & 38.0 \\
     Wnet~\cite{pan2022wnet}                  &  \checkmark   & \textbf{43.0} \\
    \bottomrule
    \end{tabular}
    \end{adjustbox}
    \caption[State of the art on audio-guided video object segmentation.]{State of the art on audio-guided video object segmentation reporting mean intersection over union ($J$). Best performance is highlighted with bold font. \textdagger: Extensions from video only methods done by Wnet for the sake of comparison~\cite{pan2022wnet}. }
    \label{tab:sota_agvos}
\end{table*}

In addition to the above audio-guided video object segmentation based on spoken descriptions, there is another formulation of audio-visual segmentation (AVS) aiming to segment sound sources in video frames from audio. Similar to the audio guided video segmentation, AVS is also critically under explored. AVSBench~\cite{zhou2022audio} presented a baseline approach that employs a module facilitating pixel-level audio-visual interaction over time, integrating audio cues to guide the visual segmentation. Another recent approach, Explicit Conditional Multimodal Variational Auto-Encoder (ECMVAE)~\cite{mao2023multimodal}, addressed audio-visual segmentation (AVS) by explicitly modeling the contribution of each modality while emphasizing the importance of cross-modal shared representation learning. Table ~\ref{tab:sota_avs} summarizes the accuracy of the only two approaches to audio-visual sound source-based segmentation.

\begin{table*}[!ht]
  \centering
  \aboverulesep=0ex
   \belowrulesep=0ex
   \begin{adjustbox}{max width=0.50\textwidth}
    \begin{tabular}{@{}l|c|c@{}}
    \toprule
    \multicolumn{1}{c|}{\multirow{2}{*}{Methods}}&\multicolumn{1}{c|}{\multirow{2}{*}{Transformer-based}}  & \multicolumn{1}{c}{Datasets} \\
    \multicolumn{1}{c|}{}&\multicolumn{1}{c|}{}& \multicolumn{1}{c}{AVSBench~\cite{zhou2022audio}} (S4/MS3)    \\ 
    \midrule
     AVSBench-Baseline~\cite{zhou2022audio}   & \checkmark   & 78.74/54.0  \\
     ECMVAE~\cite{mao2023multimodal}          &  \checkmark   & \textbf{81.74}/\textbf{57.84} \\
    \bottomrule
    \end{tabular}
    \end{adjustbox}
    \caption[State of the art on audio-visual segmentation.]{State of the art on audio-visual segmentation of sound sources reporting mean intersection over union ($J$). S4 and MS3 refers to two different benchmark settings: the semi-supervised Single Sound Source Segmentation (S4) and the fully supervised Multiple Sound Source Segmentation (MS3). Best performance is highlighted with bold font.}
    \label{tab:sota_avs}
\end{table*}

\subsection{Training and testing efficiency}
Currently, computational efficiency is not consistently reported for video segmentation models. Indeed, it is unusual for anything other than inference speed to be discussed and even that is rare. Furthermore, the computational platforms on which effiiciency is reported varies; although, in all cases contemporary middle to high end GPU based processors are employed. In the following, a few examples are given to provide an indication of the current state of the art regarding inference efficiency in video segmentation. For video object segmentation, MED-VT~\cite{karim2023med} is reported to be an efficient AVOS method achieving 0.6s amortized inference time per frame compared to 0.9s for MATNet~\cite{zhou2020motion}, 1.3s for COSNet~\cite{lu2019see} and 1.9s for RTNet~\cite{ren2021reciprocal}. Among VIS methods, CrossVIS~\cite{yang2021crossover} reported processing more than 35 frames per second (FPS) in YouTube-VIS-2019~\cite{yang2019video} while, also introducing a faster variant of the model able to process over 48 frames per second with a slightly reduced accuracy. EfficientVIS~\cite{wu2022efficient} reported achieving inference efficiency of 32-36 FPS. In contrast, TeViT~\cite{yang2022temporally} reported an inference efficiency of over 68 FPS on the same dataset. Among the most recent approaches to VIS, CAROQ~\cite{choudhuri2023context} reported an inference speed of 20 FPS for their best model using the Swin-L~\cite{liu2021swin} backbone and 40 FPS when using the ResNet-50~\cite{he2016deep} backbone. For the VSS task, SegFormer~\cite{xie2021segformer} reported over 2 FPS inference speed on the Cityscapes~\cite{kim2020video} dataset for their highest accuracy variant, while the efficient variant can achieve over 47 FPS speed with slightly lower accuracy. For the VPS task, VPSNet~\cite{kim2020video} has roughly 1 FPS using a ResNet-50~\cite{he2016deep} backbone. Slot-VPS~\cite{zhou2022slot} reported an inference speed of 4.2 and 2.0 FPS when using ResNet-50~\cite{he2016deep} and Swin-L~\cite{liu2021swin} backbones, respectively. Conversely, SiamTrack~\cite{woo2021learning} reported an inference speed of 5.1 FPS using ResNet-50~\cite{he2016deep}, but the accuracy is lower than Slot-VPS~\cite{zhou2022slot}. Among the models for the RVOS task, YOFO~\cite{li2022you} reported an inference speed of 10 FPS per object.

\section{Summary}
In this chapter,  we presented a component-wise analysis of transformer-based models for video segmentation tasks in general. We also discussed task-related specialization of components at relevant points of the presentation. In particular, the presentation has considered feature extraction, transformer encoding and transformer decoding for video segmentation. Following the presentation of the components, the performance state of the art in applying these models to video segmentation tasks also was presented.

\pagebreak

\chapter{Transformer Interpretability}\label{CH4}
\section{Overview}
Classical methods in machine learning, e.g., support vector machines and decision trees are largely interpretable by design. However, the recently dominant black box models based on convolutional neural networks and transformers are hard to interpret~\cite{hiley2019explainable}. Due to their strong performance, these models are being deployed in various applications, e.g., medical image processing and autonomous driving~\cite{litjens2017survey,kuutti2020survey}. The interpretability of such models is of utmost importance to researchers and developers in understanding the inner workings and further improving the models. The benefits from interpretability can include accuracy gains, identification of undesirable behaviors of neural network models~\cite{sundararajan2017axiomatic} and distillation of complex models into simpler ones [209,239]. Moreover, it serves in explaining decision-making for users and stakeholders. Indeed, there is a growing concern about explainability in applications impacting humans~\cite{tan2018learning, wu2018beyond}. These scientific and social requirements have led to a notable amount of research on the interpretability of deep networks.


Most of the interpretability research in computer vision has focused on convolutional network (ConvNet) models because they were ubiquitous in the recent decade. There also has been a recent increase in model agnostic interpretability methods in the literature. Given the recent rise in transformer models, there has been a corresponding rise in attention-specific interpretability methods. Interpretability falls into two basic categories based on their integration in any given model, post-hoc and ante-hoc methods. The post-hoc interpretability approaches are external methods used to analyse the learned representations and provide explanations for the outputs after the model has been trained. In contrast, the ante-hoc interpretability methods generally are integrated as a component of the model before training. Additionally, there are two categories of post-hoc interpretability, local and global. Local interpretability methods focus on understanding the model's decision-making process for a specific input or a small subset of instances to explain the model's behavior in particular cases. In contrast, global interpretability methods aim to provide holistic insights into the model's decision-making process to analyze and understand the model's biases, generalization capabilities and feature importance in general. In this chapter, we mainly focus our discussion on methods for the interpretability of transformer-based models for video understanding tasks. For the sake of completeness, we also extend our discussion beyond transformer models and video understanding to include relevant methods used for ConvNets and a broader range of computer vision tasks.


\section{Interpretability of deep networks}
In this section, we briefly discuss the most commonly used interpretability methodologies for deep networks. Once the basic methodologies have been introduced, the following section covers what has been learned through their application to vision transformers. We mainly focus on model-agnostic approaches since they can be applied to transformer models. However, we also include some prominent methods mostly used with ConvNets for relevant context.

\subsection{Understanding internal representation}
Various methods have been used in post-hoc interpretability to understand the internal representation of deep learning models. These methods range from learned filter understanding by kernel visualization to quantitative analysis of feature maps using statistical tools. The most commonly used approaches to understanding internal representation learning for deep networks are listed below.

\subsubsection{Kernel visualization}
Kernel visualization techniques work by visualizing the filter kernels, feature maps or activation maps. Typically, these methods operate by either optimizing input images to maximally activate one single filter of a model or directly visualize the numerical values of the learned kernels (e.g., \cite{erhan2009visualizing,simonyan2013deep,xiao2019gradient,zeiler2014visualizing,mahendran2015understanding,hadji2020convolutional}). Feature map visualization involves applying the learned filters to a subset of input data and visualizing the resulting feature maps. These visualizations help understand what kind of features the filters are detecting. Kernel visualization methods can be insightful for understanding the behavior of ConvNets, identifying weaknesses or biases in the model and generating visualizations of what the network has learned. For example, these methods have revealed the hierarchical organization learned by the kernels going from filters capturing primitive geometric properties (e.g., local orientation) to more complicated shapes (e.g., object boundaries). Interestingly, these learned organizations are reminiscent of the hierarchical structure in primate visual cortex as well as earlier hand-crafted computer vision approaches~\cite{rodriguez2015hierarchical}. Still kernel visualization methods are fundamental qualitative in nature and place the burden of interpretability on the user.

Activation maximization is a commonly used method for kernel visualization. This method was introduced in application to stacked auto-encoders and deep belief networks~\cite{erhan2009visualizing} and subsequently applied to supervised ConvNet models~\cite{simonyan2013deep}. The overall approach of this method is to generate inputs that maximally excite some specific units (neurons) in the network. These images help to understand the nature of features produced or patterns detected by the network. These methods work by iteratively updating a synthetic image to maximize the activation of the target units to highlight the features that the ConvNet is detecting in that layer. The optimization process can be gradient-based~\cite{simonyan2013deep} or via other other approaches~\cite{xiao2019gradient}. This approach is able to illustrate class appearance models learnt by ConvNets by visualizing different aspects of class appearance in a single numerically generated image for a class. In addition this method can also be used to generate image-specific class saliency maps. However, the insights revealed still are qualitative in nature.

There also have been various other visualization techniques to gain insight into the function of intermediate layer filters of ConvNets~\cite{zeiler2014visualizing,mahendran2015understanding}.  Deconvolutional Network (DeconvNet) has been used to visualize the feature activation projected back to pixel space and revealed the hierarchical nature of the feature representation in the network~\cite{zeiler2014visualizing}. Another approach, network inversion, used an approximation of the inverse function of the ConvNet model and sampled possible approximate reconstructions to analyze various invariance properties captured by ConvNet representations~\cite{mahendran2015understanding}. The findings reveal that ConvNet models gain certain invariances (e.g., to shift and rotation) as network depth increases. 

Finally, a recent study directly visualized the numerical values of learned convolutional kernels across all layers of several networks, i.e., their point spread functions~\cite{hadji2020convolutional}. Notably, not only were the kernels visualized, but an analytic explanation of the results was provided. The visualizations tended to appear like oriented bands filters, which was explained in terms of complex exponentials being the eigenfunctions of convolution.

\subsubsection{Analytical approach to representation understanding}
Statistical methods of multivariate similarity are available for quantitative comparisons of representations within and across neural networks~\cite{andrew2013deep,raghu2017svcca,morcos2018insights,kornblith2019similarity}. Canonical Correlation Analysis (CCA) and Centered Kernel Alignment (CKA) based approaches are among the dominant statistical techniques for neural network representation analysis. CCA is a linear method that finds a linear transformation for two sets of variables that maximizes their correlation~\cite{andrew2013deep}. In the context of neural networks, CCA can be used to compare the similarity of the activations of different layers in the same network or different networks~\cite{andrew2013deep}. In contrast, CKA is a non-linear method that measures the similarity between the representations of two neural networks by comparing the similarity of their centered Gram matrices~\cite{cortes2012algorithms, kornblith2019similarity}. The Gram matrix is a matrix that contains the dyadic product of the vectors in the representations, after mean subtraction. By centralizing the Gram matrix, CKA accounts for the differences in the mean activation levels of the two networks. CKA then computes a statistical similarity using the Hilbert-Schmidt independence criterion (HSIC)~\cite{gretton2007kernel} of these Gram matrices. Recent approaches use mini-batches to approximate the unbiased estimator of HSIC~\cite{song2012feature,nguyen2020wide,raghu2021vision}. CKA is invariant to the orthogonal transformation of representations (e.g., permutation of units under analysis) and invariant to isotropic scaling. For these reasons, CKA can provide a good similarity comparison and analysis of internal features of deep networks.

In addition to the above approaches, another notable analytical approach to probe insights into the internal representation of deep networks include frequency domain analysis. Recently, with the help of frequency domain filtering of input images, it has been revealed that conventional vision transformers (\eg ViT) show less effectiveness in capturing high-frequency image components~\cite{bai2022improving}. In particular, applying low pass and high pass filters on input images of ImageNet classification, it is observed that ViT is less effective than ConvNet models in capturing high frequency information. Based on this observation, a method is presented for improving the performance of ViT models by influencing the high-frequency components~\cite{bai2022improving}.  


\subsection{Interpretability using input contribution analysis}
Post-hoc methods also can operate by creating a heat map that shows the regional importance of input contributing to an output (e.g., a classification or estimate of some quantity). The majority of these techniques fall into attribution-based, gradient-based and perturbation-based methods. Here we present an overview of the most commonly used approaches for input contribution analysis. 

\subsubsection{Attribution-based saliency methods}
Attribution-based saliency methods are a class of interpretability methods that explain a model by assigning credit to each input feature based on how much it influenced a final output, e.g., a classification~\cite{binder2016layer,lundberg2017unified,ribeiro2016should,shrikumar2017learning}. Attribution-based saliency methods help explain the reasoning behind a network's outputs and can be used to identify the most influential features. These methods help provide insights into how neural networks make decisions and can help identify biases, debug models and improve model performance. Various applications in image recognition and natural language processing use this method to measure the importance of features.

A notable early attribution-based method, Layer-wise Relevance Propagation (LRP)~\cite{bach2015pixel,samek2016evaluating,montavon2017explaining}, is used for interpreting outputs by assigning relevance scores to each neuron in the network and propagating them back to the input layer to generate attribution scores for each input feature. LRP is a principled method that satisfies desirable properties such as conservation of relevance scores, implementation consistency and sensitivity~\cite{bach2015pixel}. Conservation of relevance score is an important property that ensures relevance scores assigned to the input features are conserved throughout the layers of a network, \eg the sum of relevancy scores from all input paths to a node equals the sum of relevancy scores to all output paths from that node. Consistency requires that the total relevance assigned to the input should be distributed among the features in a manner that aligns with the model's prediction. Finally, sensitivity measures how sensitive the output is  to changes of the relevance score. Systematic perturbation of input components with higher relevance score is expected to result in a corresponding change in the model's output. An extension to this method, partial LRP, uses LRP to quantitatively evaluate the importance of different heads at each layer to the output~\cite{voita2019analyzing}. Another prominent method, Deep Learning Important FeaTures (DeepLIFT)~\cite{shrikumar2017learning}, compares the activation of each neuron in the network for a given input to a reference activation and calculates their difference to determine the importance of each feature. DeepLIFT can handle non-linear interactions between different feature maps and is applicable to various types of deep network. 

Total Relevance Propagation (TRP), also known as transformer attribution, uses a class specific visualization method based on a relevance score similar to Layerwise Relevance Propagation (LRP) for each attention head in each layer of a transformer model~\cite{chefer2021transformer}. This method assigns local relevance based on the Deep Taylor Decomposition (DTD) principle~\cite{montavon2017explaining} and then propagates the relevancy score weighted by gradients with respect to a target class through the attention graph. The DTD rule is based on the Taylor expansion of the network's output with respect to its input, which allocates relevance to the neurons based on their importance in the network's output. This approach follows a conservation rule stating that total input relevance should equal total output relevance. For this result, it considers only the elements with a positive weighed relevance. In addition, it added provision to maintain the conservation rule for the non-parametric binary operations in the attention head, \eg skip connection and matrix multiplication. Unlike some other interpretation methods, such as attention visualization, it takes into account the entire sequence of input tokens, rather than just focusing on the attention weights of individual tokens. It does this by propagating relevance scores from the output layer back through the network, using a set of rules to assign relevance scores to each input token based on its contribution to the output.

Another line of research uses Class Activation Map (CAM) as a visualization technique to highlight the important regions of an input image that contribute to some specific output from a model~\cite{zhou2016learning}. A class activation map is a target-specific visualization method capable of detecting the discriminative image regions used by ConvNets to identify that category. From this point, CAM can simultaneously localize the class-specific image regions in a single forward pass with a ConvNet model trained on image classification. The approach works by mapping the output class score for a specific class using the weights of the last fully connected layer. Hence, it generates the activation map at a lower resolution than the input. This class activation map is upsampled to the input's resolution to compute the input relevance. This method is designed based on the findings of earlier work that ConvNets trained for classification inherently learn object detection without any supervision on the location of the object~\cite{zhou2014object}. CAM generalizes this ability of ConvNets in localizing objects in identifying the regions of an image that contribute to the final decision.

\subsubsection{Gradient-based saliency methods}
Gradient-based interpretability methods explain how a deep network arrives at a particular output by analyzing the gradients of the model with respect to its inputs. These methods allow an understanding of the reasoning behind a network's outputs and they apply to various types of networks, including convolutional networks and transformers. 
A pioneering approach in this category, Gradient×Input, computes saliency by multiplying the gradients with the input~\cite{denil2014extraction}. Another approach computed the gradients of the output to the input, then measured input sensitivity and visualized it as a heatmap~\cite{li2016visualizing}. Yet another gradient-based approach, Integrated gradients (IG), uses an additional reference input in the computation~\cite{sundararajan2017axiomatic}. The reference input can be a black image for computer vision models and a zero embedding vector for natural language models. It computes the actual input from the baseline input and measures the difference in predictions and attributes relevance scores to input features based on their contribution to this difference. While the previous method measures sensitivity to the input data, this method measures the contribution of each input element. 

Gradient-weighted Class Activation Mapping (Grad-CAM) combined the gradient-based approach with Class Activation Mapping (CAM)~\cite{selvaraju2017grad} to produce visual explanations for decisions from a large class of ConvNet-based models. Grad-CAM uses the gradients of any target label flowing into the final convolutional layer to produce a coarse localization map highlighting the important regions in the image for predicting the label. Similar to other localization approaches, \eg CAM~\cite{zhou2016learning}, Grad-CAM visualizations are also highly class-discriminative. Moreover, by fusing other pixel-space gradient visualizations with Grad-CAM, a method results that can generate high resolution class-discriminative visualizations. This method is applied to various ConvNet models for the tasks of image classification, image captioning and visual question answering.

Gradient-based and attribution-based saliency provide information on \textit{what} features or inputs influence an output. However, these methods are limited in their ability to explain \textit{how} the features interact and \textit{why} certain features are important. In response, the feature interaction analysis approach attempts to explain pairwise feature interactions in a network~\cite{janizek2021explaining}. A notable work in this category is an extension to integrated gradients that uses an architecture agnostic method with integrated Hessians~\cite{janizek2021explaining}. The Integrated Hessian matrix is a second-order derivative matrix that captures the interaction between the different features in the input image. In particular, the Integrated Hessian matrix for the output of the network represents a quantitative measure between all input features. Correspondingly, it yields insights on how different features in the input image interact with each other to produce the final output.

\subsubsection{Perturbation-based methods}
Perturbation-based methods identify the most influential features by input perturbation~\cite{shrikumar2017learning,fong2019understanding}. These methods are a class of techniques for interpreting convolutional neural networks by perturbing the input image and analyzing the effect on the model's output. The basic idea is to identify the most important input features, such as pixels or regions of the image, by measuring how sensitive the model's output is to perturbations of those features. External perturbation~\cite{fong2019understanding} can be applied to both input and intermediate layers. These methods then visualize the difference  between perturbed and unperturbed activation using a representation inversion technique.  Occlusion-based perturbation methods use regular or random occlusion patterns and measure how that affects the model~\cite{li2016understanding,petsiuk2018rise}. These methods are generally applicable as model-agnostic to any black box model without requiring changes. A shortcoming of these approaches is that generation of their saliency maps is computationally intensive.

\subsubsection{Miscellaneous approaches to highlight input importance}
SHapley Additive exPlanations (SHAP) is an approach for interpreting the outputs of a model by assigning an importance value to each feature of the input data, indicating how much that feature contributes to the model's output for a particular instance~\cite{lundberg2017unified}. SHAP is based on the concept of Shapley values from cooperative game theory. It assigns a contribution value to each player in a cooperative game based on their individual contribution to the total outcome. In the context of machine learning, Shapley values are used to determine the contribution of each feature to the final output by considering all possible coalitions of features. Another method, Local Interpretable Model-Agnostic Explanations (LIME)~\cite{ribeiro2016should} is a black box method to provide a human-understandable explanation for why a particular instance was classified in a certain way by a machine learning model. This method creates a local linear model that approximates the behavior of the underlying architecture in the vicinity of a particular instance. This local model is then used to generate explanations by highlighting the most important features or regions of the input that contributed to the model's output. Decision tree based approaches also have been used to explain ConvNet outputs quantitatively and semantically \cite{zhang2019interpreting}. These decision trees are able to encode decision modes of the ConvNet as quantitative rationales for each ConvNet output and work without requiring any modification to the training pipeline.

\subsection{Attention interpretability}
Attention specific interpretability methods use various approaches to explain the role of attention in correspondence to the outputs made by transformer models. These methods compute how attention propagates in addition to visualization of the attention maps. Some methods are specifically designed for self-attention, while there are other generalised approaches that encompass multimodal and cross-attention. Given the model agnostic approaches to interpretability, these attention specific methods are also popular for transformer interpretability for natural language and vision tasks. In this section we present different approaches commonly used for attention interpretability.

\subsubsection{Visualizing raw attention}
Raw attention visualization has been used to analyse the role of self-attention in transformer models. One notable work in this area focused on multiscale visualization of attention by presenting techniques for visualizing attention at three levels of granularity: the attention-head level, the model level and the neuron level~\cite{vig2019visualizing}. These techniques were initially proposed for language models. The attention-head level visualizes the self-attention patterns in attention heads of a transformer layer as flow paths from input tokens to output tokens. The model level provides a high-level view of attention across all of the model’s layers and heads in a tabular form for a particular input. The neuron level is used to visualize the contribution of individual neurons in the query and key vectors in computing attention.  

\subsubsection{Attention rollout and attention flow}
Attention rollout and attention flow have been proposed as measures to quantify the attention to input tokens that flows through the self-attention layers in a transformer  model~\cite{abnar2020quantifying}. Initially, these methods were applied to Natural Language Processing (NLP) tasks, and subsequently were extended to vision models~\cite{dosovitskiy2020image}. These methods can indicate a set of input tokens that are important for the final decision from the model. They use attention weights as the relative importance of the input tokens. They provide views on the information flow through attention layers complementary to the attention visualization methods. In addition, the information flow through attention layers shows higher correlation with importance scores of input tokens from gradient methods presented earlier for ConvNets. The presented visualizations suggest that these methods are better approximations than raw attention of how input tokens contribute to an output. Attention rollout uses the assumption that the identities of input tokens are linearly combined through the layers based on the attention weights. So, this method rolls out the weights to capture the propagation of information from input tokens to intermediate hidden embeddings. In contrast, attention flow considers the attention graph as a flow network and computes maximum flow values from hidden embeddings to input tokens using a maximum flow algorithm. Both methods take the residual connection in the network into account to better model the connections between input tokens and hidden embedding. These methods recursively compute the token attentions in each layer of a model from the embedding attentions.

\subsubsection{Gradient self-attention maps}
Gradient Self-Attention Maps (Grad-SAM) is a gradient-based interpretation method for analyzing the self-attention units in transformer models and to identify the influence of input elements on the model’s output~\cite{barkan2021grad}. This is a target specific post-hoc interpretability method that computes the attention maps weighted by their rectified gradients at each attention head of all the layers. Then, it aggregates the weighted attention scores of all the attention heads in all of the layers by computing the arithmetic average. The motivation for using rectified gradients is to prevent the accumulated negative gradient from cancelling out the positive influence during the averaging. Grad-SAM was initially demonstrated on the BERT~\cite{devlin2018bert} model fine-tuned for several NLP tasks including sentiment analysis and article classification. However, the general approach of Grad-SAM also is suitable for application to computer vision models.

\subsubsection{Attentive class activation tokens}
Attentive Class Activation Tokens (AttCAT) is a method to generate token-level explanations that leverages features, their gradients and their self-attention weights~\cite{qiang2022attcat}. This method was demonstrated on several transformer models for NLP including BERT~\cite{devlin2018bert}, DistillBERT~\cite{sanh2019distilbert} and RoBERTa~\cite{liu2019roberta}. However, the approach also is suitable for application to computer vision models. This method attempted to address the shortcomings of earlier approaches that use attention-weights without considering feature maps. The generated token-level explanations are considered as impact scores to quantify the influence of inputs on the model’s outputs. This class activation based method is capable of discriminating positive and negative impacts on the model’s output using the directional information of the gradients. For this purpose, the method defines a new impact score to measure the influence of a token on the output. The impact score measures both the magnitude and direction of the impact to interpret how much influence a token has on the output and whether it is a positive or negative influence.

\subsubsection{Generic attention-model explainability}
The generic attention interpretability method was presented to address the limitation of previously discussed interpretability approaches focused on self-attention only~\cite{chefer2021generic}. This method is a class-dependent interpretability method that can explain output of any transformer-based architecture. Toward this end, the method extends attention interpretability beyond self-attention to include bi-modal attention (\eg text and image) and cross-attention (\eg transformer decoders). This method produces relevancy maps from the attention layers for each of the interactions between the input modalities in the network. For bi-modal attention, the intra-modal relevance is initialized as ones and inter-modal relevance as zeros. Subsequently, the relevancy maps are updated based on the attention maps between the token embeddings. To account for multi-head attention, gradient weighting was used to average across the heads considering only the positive weighted relevancy. In addition to the generalization, this method also shows superiority in explaining self-attention compared to other approaches~\cite{abnar2020quantifying,chefer2021transformer,selvaraju2017grad,voita2019analyzing}.

\subsection{Understanding the role of the temporal dimension in video models}
\subsubsection{Visualization studies}
Activation maximization~\cite{erhan2009visualizing} has been extended to work on video to understand what has been learned by various convolutional approaches to action recognition. In particular, the problem is formulated as a gradient-based optimization to search for the preferred spatiotemporal input in the input spacetime domain~\cite{feichtenhofer2020deep}. In this work, multiple two-stream architectures were visualized to show that local detectors for appearance and motion arise to form distributed representations for recognizing human actions.

In the context of a specific convolutional model for early action recognition \cite{zhao2021interpretable}, visualizations and quantitative results suggest that temporal differences of intermediate layer features provide better predictions than framewise features or features from other layers. It also was found that the learned filters for extrapolating observed features into the future in support of early recognition operate as a convolutional approximation of optical flow-based warping.


\subsubsection{Static vs. dynamic information}
Research has suggested that scene context can dominate motion and more general dynamic information in video understanding \cite{vu2014predicting,he2016human,choi2019can}. Such observations have motivated a number of studies on deep learning models that specifically address the importance of single frame static information vs. dynamic information revealed through consideration of multiple frames. 

\paragraph{Understanding scene bias.}
Of particular interest here is an approach that addressed scene bias in action recognition by mitigating the effect during training \cite{choi2019can}. The methodology was based on masking the actors in videos to document scene bias in state-of-the-art action recognition models. Notably, actor masking to document reliance on scene context during recognition originally was performed considerably earlier \cite{derpanis2012action}.

\paragraph{Temporal importance analysis.}
An approach for temporal importance analysis was proposed to benchmark video models using a meta dataset of temporal classes curated from a subset of an action recognition dataset \cite{sevilla2021only}. In particular, human performance was compared on temporally shuffled data and original data to select a subset of categories where temporal information is most important. This subset of classes are then used to measure the capability of different models in learning temporal correspondence. This methodology is different from standard input perturbation approaches, because it relies on human observers. The presented experiments demonstrate that using the curated temporal classes for training improves generalization of the model to unseen classes. In addition, the activation maps generated from the model trained on temporal classes are more intuitive than the activation maps from the same model trained without these temporal classes. 

\paragraph{Static vs dynamic bias.}
Static and dynamic bias metrics are used in perturbation-based interpretability approaches to analyze the capability of video understanding models to capture dynamic information~\cite{kowal2022deeper}. The metrics are based on the mutual information between sampled video pairs corresponding to static and dynamic factors. These video samples generally are produced by some perturbation on the original videos and/or their optical flow. The method was applied to various models and datasets for two video understanding tasks, action recognition and object segmentation. It was determined that the majority of models and datasets are biased toward static information.

\paragraph{Appearance free dataset.}
A challenge in studying the relative contribution of static vs. dynamic information is the disentangling of these two attributes in input video. This challenge motivated the Appearance Free Dataset (AFD) for video action recognition \cite{ilic2022appearance}. AFD is a a synthetic derivative of the UCF101 dataset~\cite{soomro2012ucf101} where individual frames show no pattern relevant to action recognition but the video reveals the relevant underlying motion. The dataset is generated by animating spatial noise using image motion extracted from the UCF101 dataset. This dataset is evaluated with human observers to reveal that the human recognition from purely temporal information is very similar to that of RGB data. In contrast, all contemporary action recognition algorithms had their performance markedly reduced when their input was AFD compared to RGB. The observation was used to define a novel action recognition algorithm that explicitly considered optical flow input along with RGB, which equaled human level performance.

\paragraph{Adaptive temporal selection.}
ATemporal Probe (ATP) is based on a lightweight transformer to select the most representative frame from a video clip\cite{buch2022revisiting}. The approach uses this ATP model to evaluate the performance of an image-only baseline on the adaptively selected video frame to justify the necessity of temporal information for the specific input in the task under consideration. The approach is demonstrated in the context of standard discriminative video-language tasks, such as video question answering and video-language retrieval. This method is able to create static and dynamic subsets of a dataset based on the importance of dynamics. It also was able to evaluate the capability of video models to capitalize on temporal dynamics by comparing with the lower bound from ATP augmented image-language models. The experiments show that image models with adaptively selected video frames can achieve on par or better performance compared to state-of-the-art video models. This observation reveals the limitation of these models in learning temporal features.

\subsection{Ante-hoc methods for inherently interpretable models}\label{sec:antehoc-methods}
Recently there is a growing interest in developing inherently interpretable deep learning models for real world application, in contrast to adopting post-hoc model agnostic interpretability methods with black-box models~\cite{lipton2018mythos,rudin2019stop}. An inherently interpretable deep learning model is a type of neural network designed to be transparent and explainable, allowing for better understanding and insights into its decision-making process. Such an inherently interpretable model is expected to be capable of generating decisions based on human-understandable categories grounded in domain expertise. Among many, one of the most highlighted arguments in support for ante-hoc interpretable models suggests that interpretability is best learned during model training~\cite{rudin2019stop}. These models use techniques such as explicit feature representation~\cite{zhang2018interpretable}, attention mechanisms~\cite{rigotti2021attention} or decision trees~\cite{ribeiro2016should} to make their internal workings more accessible to human understanding. This interpretability can help identify and correct errors or biases, increase trust and acceptance and improve model performance in various fields such as healthcare, finance and law.

Several inherently interpretable models have been presented. A general method for modifying a traditional ConvNet into an interpretable one has been presented where each filter in deeper layers represents a specific object part~\cite{zhang2018interpretable}. As another approach, Concept Bottleneck Models (CBMs) were introduced with the idea to first predict an intermediate set of human-specified concepts and then use those concepts to predict the final output~\cite{koh2020concept}. As an example, wing colour can be a concept for the objective of bird species identification. Notably, training of these models requires an additional annotation of concepts other than the target label. In addition to facilitating inherent interpretability, these models also allow for intermediate intervention at the concept level. As yet another approach, concept whitening was introduced for interpretable ConvNet models for image recognition~\cite{chen2020concept}. The concept whitening layer normalizes feature maps (e.g., analogous the batch normalization) and decorrelates (whitens) the latent space. Another approach augmented a base model with additional modules that generate concept explanations, which is trained jointly with the original model \cite{sarkar2022framework}. The explanation generation module is easy to integrate with existing networks and also provides the flexibility to incorporate different forms of supervision (\eg both weak and strong supervision). This method has shown better predictive performance compared to recently proposed concept-based explainable models through use of full concept supervision.

A more radical approach is to replace standard learned model components with more readily interpretable ones. One notable approach in this direction, B-cos Networks, introduced the B-cos transform as a drop-in replacement for linear units to increase the interpretability of deep networks by promoting weight-input alignment during training~\cite{bohle2022b}. The B-cos transform is formulated as a variation of the linear transform that can affect the parameter optima of the model when combined with the binary cross entropy loss. This transform does not require a non-linear activation, allowing the full model computation to be summarised by a single linear transform. This summarized linear transform can help explain output logits by directly visualizing the corresponding row. The B-cos transform is designed to be compatible with existing architectures allowing it to be integrated into common models such as VGGs~\cite{simonyan2015very}, ResNets~\cite{he2016deep}, InceptionNets~\cite{szegedy2016rethinking} and DenseNets~\cite{huang2017densely}. Finally, ConvNet models have been presented wherein all convolutional filters are analytically defined based on mathematical and signal processing principles \cite{bruna2013invariant,hadji2017spatiotemporal}, with the analytic specifications being inherently interpretable. The analytic specification also is used to define network pooling and nonlinearities in a principled fashion.


\section{Vision transformer interpretability}
In this section we present a summary of notable experimental results for transformer interpretability. This includes experiments with both model agnostic methods and attention specific methods. 

\subsection{Internal representations in transformers}
\begin{figure}[!ht]
    \centering
        \centering
        \includegraphics[width=0.58\textwidth]{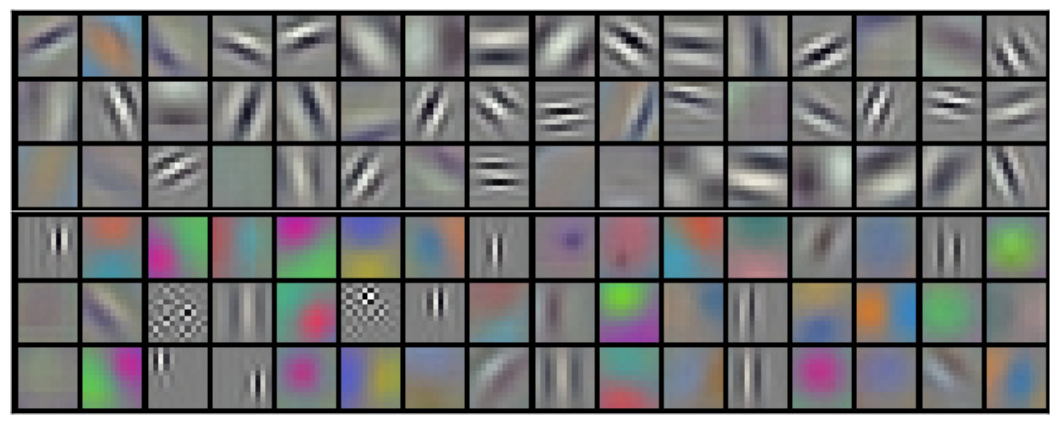}
        \includegraphics[width=0.41\textwidth]{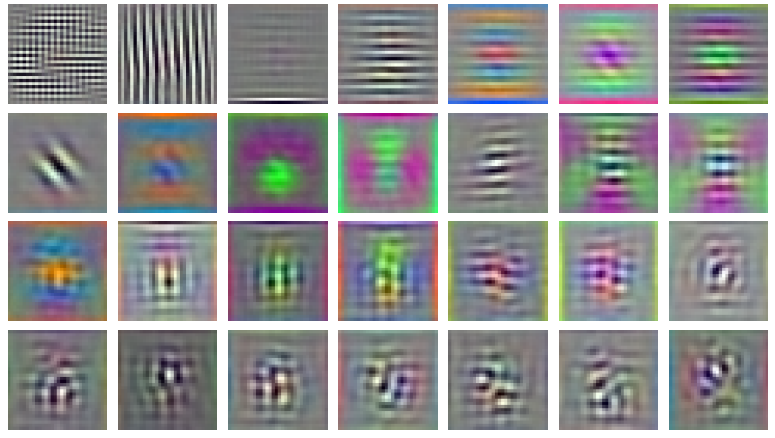}
\caption[Filter Map Visualizations of ConvNet and Vision Transformer]{Filter Map Visualizations of ConvNet and Vision Transformer. Shown are filter map visualizations of early convolution layers of a ConvNet and the linear embedding layer of a vision transformer. Left: The 96 kernels sized 11×11×3 learned in initial convolutional layer of AlexNet~\cite{krizhevsky2017imagenet} where the first 48 kernels were learned using GPU 1, and the remaining 48 kernels were learned using GPU 2. Right: First 28 principal components of linear embedding layer of ViT~\cite{dosovitskiy2020image}. Images reproduced from \cite{krizhevsky2017imagenet} and \cite{dosovitskiy2020image}.}
\label{fig:filter_viz}
\end{figure}

\paragraph{Filter visualization.}
Visualization of learned filter kernels has been used for analyzing deep networks. Visualization of early filters in ConvNets has shown that they largely resemble oriented bandpass filters, e.g., \cite{zeiler2014visualizing,mahendran2015understanding,krizhevsky2017imagenet,hadji2020convolutional}. Theoretical reasons for the emergence of such filters also has been provided in terms of the correspondence between convolution and linear shift invariant systems \cite{hadji2020convolutional}. In contrast, visualization of the linear embedding layer of transformers suggests that while oriented structure is observed, it has more global support compared to ConvNets, which is consistent with the relative ability of transformers to perform data association across their entire input domain. Also, beyond unimodal orientation, transformers appear to implement more complicated data associations. Figure~\ref{fig:filter_viz} presents visualizations of the filter maps of early convolutional layers of AlexNet~\cite{krizhevsky2017imagenet}(left) and the linear embedding layer of ViT~\cite{dosovitskiy2020image}(right).


\begin{figure}[!ht]
    \centering
    \includegraphics[width=0.95\textwidth]{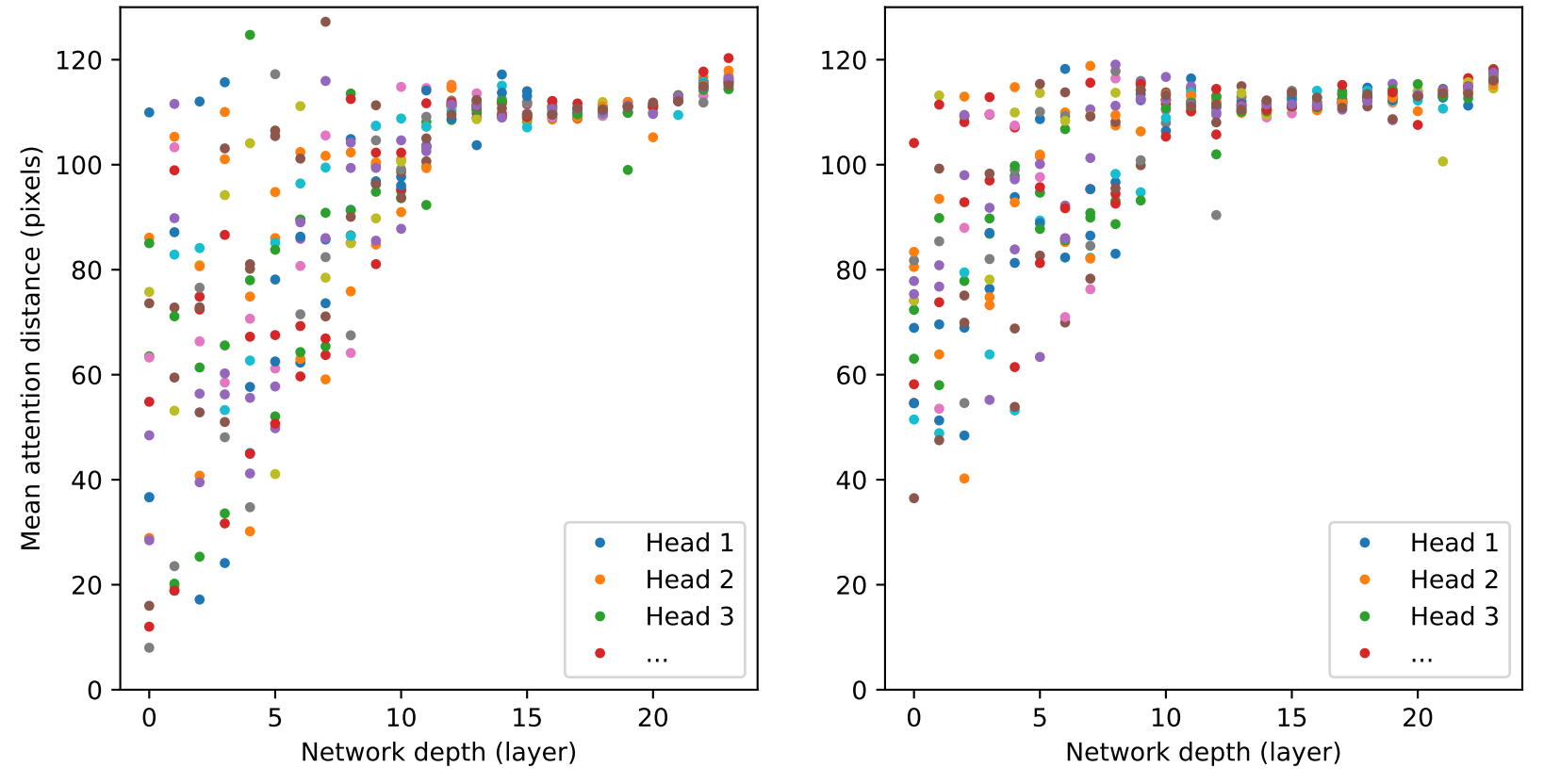}
    \caption[Attention Distance in Vision Transformer.]{Attention Distance in Vision Transformer. Attention distance of ViT (left) vs. ResNet+ViT. Figure reproduced from \cite{dosovitskiy2020image}.}
    \label{fig:vit_attention_distance}
\end{figure}

\paragraph{Attention distance.}
ViT used attention distance between tokens on various heads of the multi-head attention layer as an analogue to the receptive field in ConvNets~\cite{dosovitskiy2020image}. In particular, attention weights were used to compute the average distance in image space across which information is integrated. A comparison was presented showing attention distance of ViT and a hybrid model with ConvNet+ViT. As shown in the Figure~\ref{fig:vit_attention_distance}, it is evident that the attention distance is large for both the pure transformer model and the hybrid model. This evidence shows that in both of the cases deeper layer attention can integrate information across a long distance in image space. In contrast, the early attention layers for both of the models has attention spreading from short to long distance. Again, the early attention heads for the pure transformer model has highest spread of attention distance, encompassing short to long. This result suggests that some attention heads in the early layers learn localized attention similar to the smaller receptive fields of ConvNets. The overall results motivate the use of pure transformer models, without relying on ConvNets to provide localized support.

\subsection{Input and feature contribution analysis in vision transformers}

\begin{figure*}[t]
	\begin{center}
		\setlength\tabcolsep{0.7pt}
		\def\arraystretch{0.95}
		\resizebox{0.99\textwidth}{!}{
			\begin{tabular}{c}
                \includegraphics[width=0.95\textwidth]{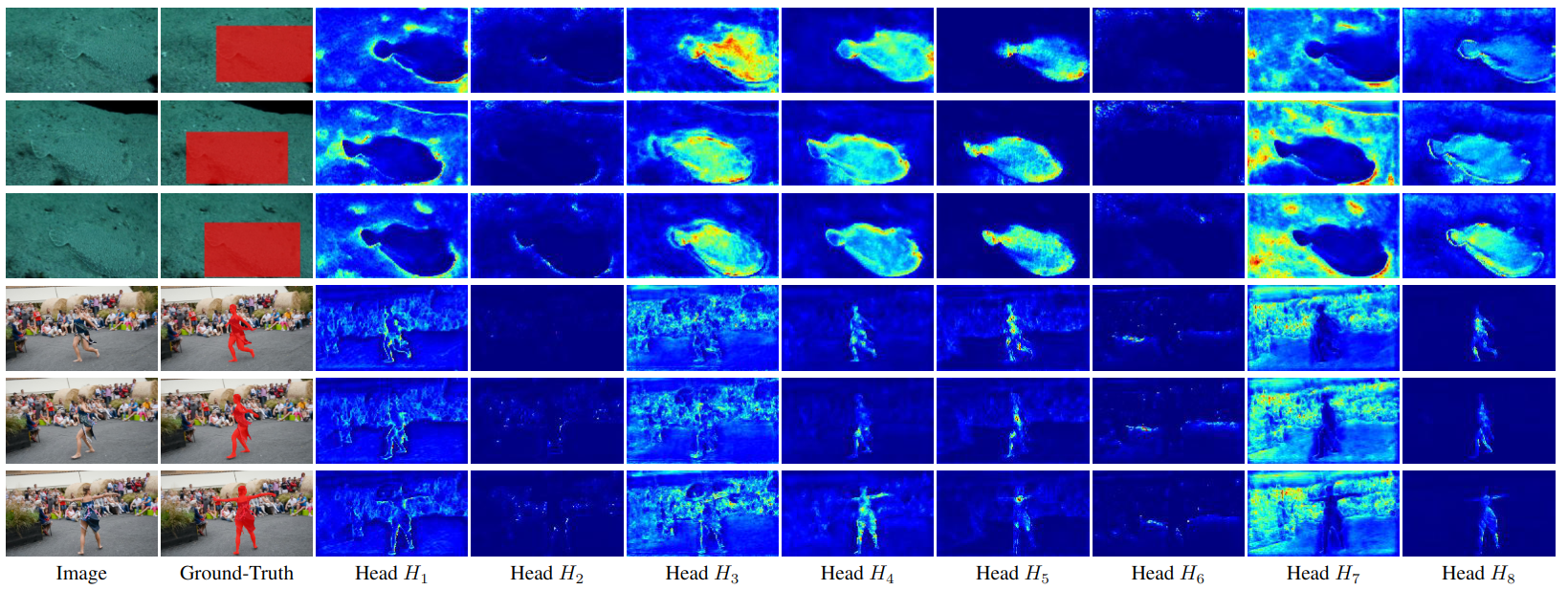}\\				
			\end{tabular}}
			\caption[Visual Explanation of Transformer Decoder Queries]{Visual Explanation of Transformer Decoder Queries. Shown are visual explanations of object related semantic properties learned by the dynamic queries of a transformer decoder used in video object segmentation. The visualization in the form of heat maps shows the semantic attributes generated in different attention heads of the object attention map of the transformer decoder on video sequences from two datasets. The top three rows show three example frames from the MoCA dataset \cite{lamdouar2020betrayed} and the bottom three rows show three example frames from the DAVIS 2016 dataset~\cite{perazzi2016benchmark}. It is seen that different heads highlight different object and background components. Figure reproduced from~\cite{karim2023med}.}
		\label{fig:medvt_attn}
		\end{center}
\end{figure*}

\paragraph{Decoder query understanding with visualization.}
Recently MED-VT used visualization to analyze the object attention maps generated by the learnable queries of the transformer decoder for video object segmentation \cite{karim2023med}. As shown in Figure~\ref{fig:medvt_attn}, different attention heads in the object semantic query attention maps are able to disentangle different levels of semantic abstraction related to the primary video object. In particular, each of the attention heads learns to attend to different aspects of object representation, including overall object shape as well as its extremities, \eg boundaries. Notably, most attention heads generate well localised saliency maps for the primary objects in the video. Conversely, some of the attention heads localise complimentary background. This evidence suggests that decoder queries tend to disentangle high level attributes such as object boundaries, foreground and background in different heads. In particular, the heads are representing the two classes of the task, foreground and background. These results not only document the localization ability of the adaptive object queries but also provide evidence on their ability to disentangle class semantic attributes.

\paragraph{Optimization-based feature visualization.}
Similar to its use with ConvNet models, gradient descent based optimization has been used to explore the inner workings of transformer models. One particular example is an optimization-based feature visualization that analyses several transformer models~\cite{ghiasi2022vision}. The approach initially generated synthetic images that maximize the activation of a particular model component and subsequently identifies natural images that have the same behaviour. It was found that the feed forward layer features with high-dimensionality produce more interpretable visualizations than the low-dimensional features from the key, value and query of self-attention. This work conducted visualizations for a variety of vision transformer models. The results show that early layers of vision transformer models also learn to recognize local features, \eg colour, edges and texture, while deeper layers learn semantic abstraction similar to ConvNets. However, different from ConvNets, vision transformers are found to use background information in addition to object features and show resilience to dependence on texture.

\paragraph{Canonical correlation analysis.}
Canonical correlation analysis based methods have been employed for quantitative comparisons of representations within and across three networks~\cite{raghu2017svcca,morcos2018insights,kornblith2019similarity}. The experiments reveal several key differences between the representation learning mechanism of the two currently dominant classes of models, ConvNets and transformers. One of the key findings is that vision transformers are able to learn global features at early layers in contrast to ConvNets, which is to be expected based on the relative early layer support of transformers vs. ConvNets. Another finding indicates that vision transformers produce comparatively stronger intermediate features compared to ConvNets, making the vision transformer a better backbone for dense estimation tasks.


\subsection{Attention interpretability}
\begin{figure}
    \centering
    \includegraphics[width=0.31\textwidth]{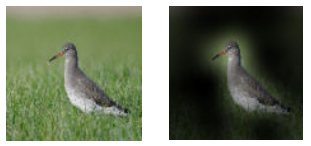}
    \includegraphics[width=0.31\textwidth]{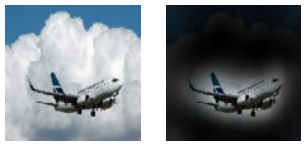}
    \includegraphics[width=0.31\textwidth]{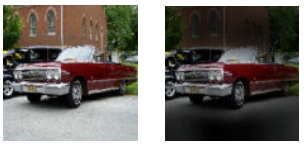}
    \includegraphics[width=0.31\textwidth]{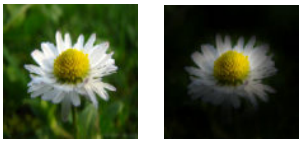}
    \includegraphics[width=0.31\textwidth]{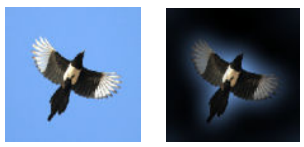}
    \includegraphics[width=0.31\textwidth]{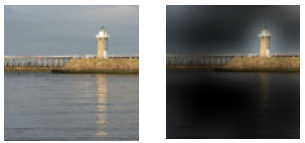}
    \caption[Attention Maps using Attention Rollout.]{Attention Maps using Attention Rollout. Shown are attention maps for several input images in a vision transformer. Each horizontal image pair shows the input (left) and the attention map (right). The attention maps tend to highlight the key object in the image. Figure reproduced from \cite{dosovitskiy2020image}.}
    \label{fig:vit_attention_maps}
\end{figure}

\paragraph{Attention rollout.}
ViT used attention maps from the output token to the input space to demonstrate that the model attends to image regions that are semantically relevant for classification~\cite{dosovitskiy2020image}. In particular, attention rollout was used to project the attention map to the input space~\cite{abnar2020quantifying}. The approach includes averaging attention weights across all heads and then recursively multiplying the weight matrices of all layers for the mixing of attention across tokens through all layers. A subset of the attention map visualizations from ViT are presented in Figure~\ref{fig:vit_attention_maps}. The attention maps largely correspond to the semantically important input image regions for making the desired classifications. However, this approach does not provide insights on the information captured in different heads at different attention layers.

\begin{figure*}[!ht]
	\vspace{-0.1cm}
	\begin{center}
		\setlength\tabcolsep{0.7pt}
		\resizebox{0.91\textwidth}{!}{
			\begin{tabular}{m{6em}m{6em}m{6em}m{6em}m{6em}m{6em}m{6em}}
				\includegraphics[width=0.14\textwidth]{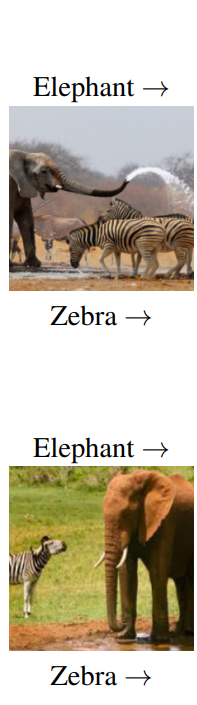}
				&
				\includegraphics[width=0.14\textwidth]{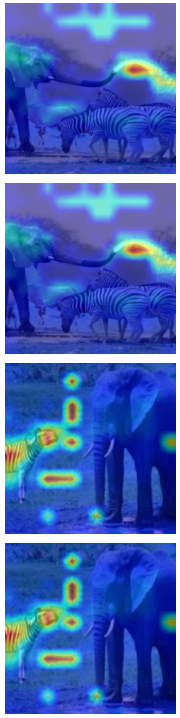}
                    &
				\includegraphics[width=0.14\textwidth]{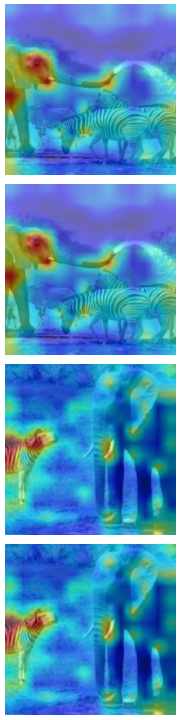}
				&
				\includegraphics[width=0.14\textwidth]{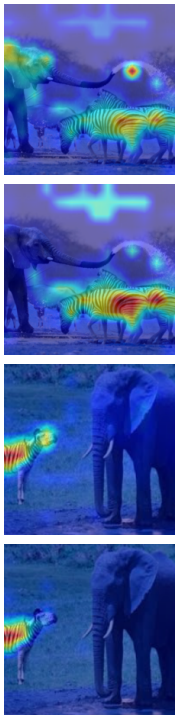}
                    &
				\includegraphics[width=0.14\textwidth]{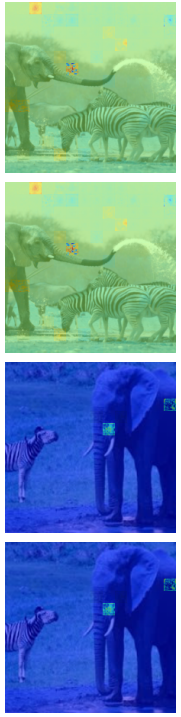}
				&
				\includegraphics[width=0.14\textwidth]{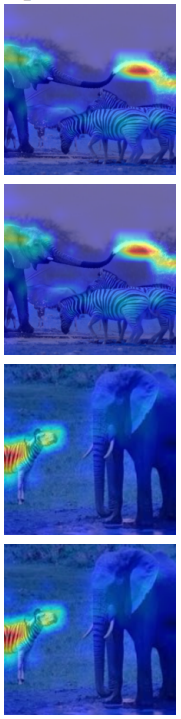}
				&
				\includegraphics[width=0.14\textwidth]{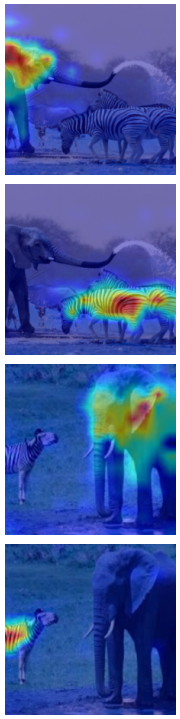}
                    \\
                    \multicolumn{1}{c}{Input} &\multicolumn{1}{c}{raw-attention} & \multicolumn{1}{c}{rollout~\cite{abnar2020quantifying}} & \multicolumn{1}{c}{GradCAM~\cite{selvaraju2017grad}} & \multicolumn{1}{c}{LRP~\cite{voita2019analyzing}} & \multicolumn{1}{c}{partial LRP~\cite{voita2019analyzing}} & \multicolumn{1}{c}{TRP~\cite{chefer2021transformer}} \\ 
			\end{tabular}}
               \caption[Visual Comparison of Attention Map Generation Methods.]{Visual Comparison of Attention Map Generation Methods. Shown are attention maps from various methods for input images from ImageNet~\cite{russakovsky2015imagenet} containing two objects from two different classes. Approaches vary in their ability to concentrate on the target objects. Figure reproduced from \cite{chefer2021transformer}.}

            \label{fig:ch5_trp_two_class}
		\end{center}
\end{figure*}

\paragraph{Relevance propagation.}
Total Relevance Propagation (TRP) has been demonstrated on the vision transformer ViT~\cite{dosovitskiy2020image}. The experiments include comparing classification outputs and interpretability saliency using segmentation metrics. The results suggest that the networks focus on the most salient components of the objects for classification. For instance, the model tends to focus on the abdomen to capture the stripes for the class zebra. In comparison, in classifying elephants, the focus tends to be on the head region. The evaluation with segmentation metrics considers the input relevance visualization as soft-segmentation of the image and compares it to the segmentation ground truth using standard evaluation metrics for segmentation. According to these metrics, TRP yielded saliency maps that are in better accord with the ground truth compared to rollout~\cite{abnar2020quantifying}, GradCAM~\cite{selvaraju2017grad}, LRP~\cite{binder2016layer} and partial LRP~\cite{voita2019analyzing}. A visualization of generated attention maps for input containing objects from two classes is shown in Figure ~\ref{fig:ch5_trp_two_class}. 

\begin{figure*}[!ht]
	\begin{center}
		\setlength\tabcolsep{0.7pt}
		\resizebox{0.91\textwidth}{!}{
			\begin{tabular}{c}
				\includegraphics[width=0.15\textwidth]              {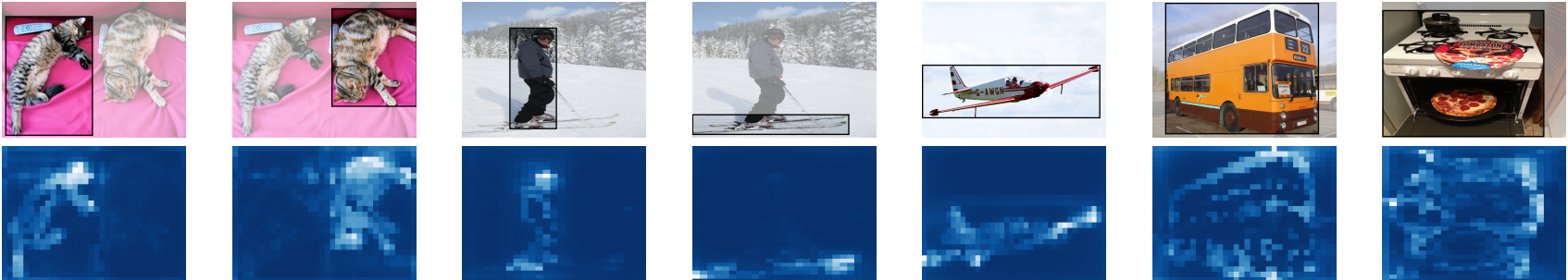}\\
                \end{tabular}}
			\caption[Segmentation Mask using Input Relevance.]{Segmentation Mask using Input Relevance. Shown are visualizations comparing bounding box masks from DETR~\cite{carion2020end} (top row) with the segmentation mask from the decoder attention visualization method (bottom row). The results show the generation of segmentation masks consistent with the bounding boxes. Figure reproduced from~\cite{chefer2021generic}.}
			\label{fig:chefer2021generic_fig6}
		\end{center}
\end{figure*}

\paragraph{Generic attention-model explainability.}
An approach for generic relevancy propagation presented experiments using positive and negative perturbation tests \cite{chefer2021generic}. The approach was demonstrated on transformer models containing various categories of attention, \eg self-attention, multimodal co-attention and cross-attention decoder. This approach used self-attention vision-language model VisualBERT~\cite{li2019visualbert} and vision model ViT~\cite{dosovitskiy2020image} to demonstrate the relevance score for self-attention. For VisualBERT, perturbing the tokens with high relevance shows dramatic drop in model accuracy indicating the practical significance of the generated relevance. The presented results using ViT showed superior performance in producing class-specific visualizations. For the experiment with multimodal attention, positive and negative perturbation tests were performed on each modality separately using the multimodal transformer LXMERT~\cite{tan2019lxmert} for visual question answering~\cite{antol2015vqa}. Again, perturbing the tokens with high relevance shows dramatic drop in model accuracy indicating the practical significance of the generated relevance. The experiment with an encoder-decoder model uses DETR~\cite{carion2020end} trained for object detection on the MS COCO dataset~\cite{lin2014microsoft}. In particular, Otsu's threthreshold method~\cite{otsu1979threshold} was used to produce segmentation masks from the saliency; subsequently, standard segmentation evaluation protocols were followed. The segmentation masks produced from input relevancy shows that the generated masks are consistent with the bounding boxes, as shown in Figure~\ref{fig:chefer2021generic_fig6}.

\subsection{Understanding temporal modeling in video transformers}
The static and dynamic bias metric was applied to video transformers to understand how well they capture these information types~\cite{kowal2022deeper}. In particular, the findings suggests that the popular transformer models MViT~\cite{fan2021multiscale} and Timesformer~\cite{bertasius2021space} both encode more static information than dynamic. It also was noted that most standard datasets for training video understanding algorithms are biased toward static information. The Appearance Free Data (AFD) paradigm has been applied to a wide variety of deep learning action recognition models, including the transformer MViT \cite{fan2021multiscale}. Like all of the models evaluated, MViT's performance was severely compromised when forced to deal with only dynamic information. These results show the limited capability of inherently modeling temporal dynamics in state-of-the-art video action recognition models. The ATemporal Probe (ATP) model was used in interpreting the static bias present in video datasets and limitations of temporal modeling in video models \cite{buch2022revisiting}. The presented experiments reveal that many state of the art video-language benchmarks are heavily static biased and lack representation of temporal correspondence. It is evident from the results that this static bias can allow models to achieve good results using static information of individual frames; however, they also can result in poor generalization to data points requiring strong temporal modeling. The results also indicate that integrating ATP into state-of-the-art video models can improve accuracy of these models, especially when temporal correspondence is important.

\subsection{Inherently interpretable vision transformers}
The ongoing debate between post-hoc explainability vs. inherently interpretable models has resulted in recent proposals of inherently interpretable transformers. A notable work, ConceptTransformer (CT) presented a transformer-based module designed to be used as a classifier head in deep learning models~\cite{rigotti2021attention}. The module generates classification outputs using cross-attention  between input features and a set of embeddings representing high level interpretable concepts. This CT classifier is capable of enhancing the model with domain knowledge in the form of plausible cross-attention weights. The CT classifier module only adds an overhead of a modification to the loss function in the host model to enforce plausibility of the explanation during the training phase. An empirical evaluation of the approach on three image benchmark datasets was presented: MNIST Even/Odd~\cite{barbiero2022entropy}, CUB-200-2011~\cite{welinder2010caltech} and aPY~\cite{farhadi2009describing}. 


As discussed in Sec~\ref{sec:antehoc-methods}, B-cos transformers were introduced as a holistically explainable vision transformer model capable of inherently providing explanations for its decisions~\cite{bohle2023holistically}. 
This formulation allows for summarizing the entire model with a single linear transformation. The generated summaries from the transformer model with dynamic linear layers are highly interpretable as the parameters are implicitly optimised to align with relevant input patterns. Experimental result of applying this approach to the ViT model shows that Bcos-ViTs can perform competitively to the baseline with additional advantage of inherent explainability. The inherent linear summaries also are found to outperform state-of-the-art post-hoc explanation methods.

\subsection{Interpretability for efficient transformer design}
IA-RED${}^2$ is an interpretability-aware redundancy reduction method for vision transformers to improve computational efficiency~\cite{pan2021ia}. The paper argues that some of the input patches are redundant and such redundancy is input-dependant. On that basis, an interpretable module is introduced to dynamically drop several input patches with minimal correlation to reduce the redundant computation. A policy network is adopted by leveraging the idea of dynamic inference to decide and discard uninformative patches. This way of dynamically dropping some less informative patches in the original input reduces the length of the input sequence to yield computational efficiency without suffering accuracy. Further, by using the policy network to discriminate regional importance, this approach is inherently interpretable. Empirically, on both images and videos a notable speed up on state-of-the-art models was achieved with minimal drop in accuracy. In addition, qualitative examples were provided to verify the ability of the multi-head interpreter compared to raw-attention. Also, quantitative results were presented using weakly-supervised image segmentation on ImageNet-segmentation~\cite{guillaumin2014imagenet} to quantify the superiority of the multi-head interpreter in identifying semantically important image regions. 

\section{Summary}
In this chapter,  we presented discussion of notable interpretability approaches used for deep learning models to date. These include visualization techniques, quantifying input contribution, attention interpretability and temporal modeling in video understanding. Following the general discussion, we presented discussion on application of interpretability analysis to various transformer models. Finally, we presented a brief discussion on application of interpretability for efficient and self-explainable transformers.

\pagebreak

\chapter{Conclusion and Future Work}\label{CH5}
Throughout this survey, we have discussed various transformer models for video segmentation tasks. These include video segmentation ranging from unimodal to multimodal scenarios. This survey also contains insights learned from different interpretability methods for transformers and video understanding. Even though recent efforts using video transformers have advanced the performance on video segmentation tasks using heuristics and intuition, analytic interpretation-guided improvement and optimization remain less explored. In this section, we review the current trends in the application of transformers to video segmentation. In light of these trends, we highlight various research gaps found in the literature and potential urgent challenges waiting to be solved.

\section{Current trends}
While various transformer-based models have proven to push the state of the art further over ConvNet models on video segmentation, most models are hybrid architectures with transformers in conjunction with ConvNets. Currently, the most most common such approach is to use a transformer only for global context aggregation, rather than as a more complete model. Again, this shift to partial component-wise transition might be because of less explored research in transformer interpretability. To conclude this review, we now discuss the current trends in transformer applications to video segmentation and interpretability methods.

\subsection{Application of transformers to video segmentation}

\paragraph{Spatiotemporal feature encoder.}
Trending approaches to video segmentation greatly rely on transformer-based models for encoding RGB pixels from a video clip into a high dimensional feature space suitable for disentangling the semantics~\cite{li2022mvitv2,liu2021swin}. Following initial ConvNet-based feature extraction, a transformer is used to provide context encoding \cite{wang2021end,duke2021sstvos,karim2023med,botach2022end}. Interpretability analysis of intermediate feature maps suggests that vision transformers produce comparatively stronger intermediate features than ConvNets \cite{raghu2021vision}. Further analysis reveals that transformers generate features that are more robust than ConvNet features with less sensitivity to high frequency textures \cite{ghiasi2022vision}. This evidence suggests that transformers can serve as better backbones for dense estimation tasks. Overall, due to the above mentioned benefits and strong performance, the use of spatiotemporal transformers, which can learn to attend to both spatial and temporal information in video frames, is growing for encoding video.


\paragraph{Multimodal feature encoder.}
In multimodal video segmentation, \eg text guided or speech guided video segmentation, a transformer-based cross-attention mechanism is used to communicate contextual information between modalities~\cite{khoreva2018video,li2022you,botach2022end,pan2022wnet}. The approach of using cross-attention-based multimodal feature aggregation has pushed the state of the art over simpler concatenation of multimodal features. Further, using a transformer provides uniform modeling of different modes of input for a generalised framework.

\paragraph{Query decoder.}
As discussed in Sec.~\ref{sec:mask_decoder}, design of the transformer-based decoders to learn queries for object localization in video frames is dominant in various video segmentation tasks. In unimodal video segmentation, queries are learned in the decoder to localize objects in videos in the form of a bounding box or saliency map~\cite{karim2023med,cheng2021mask2formervid}. In text guided video segmentation, state-of-the-art approaches use object queries based on contextual guidance from text features~\cite{khoreva2018video,seo2020urvos,botach2022end}. Another line of trending research used a transformer decoder to learn queries to generate dynamic kernels for convolution to segment objects~\cite{wu2022language}.

\paragraph{Efficient transformers.}
Sequences produced from flattened video patches incur considerable computation and memory demands during attentional processing \cite{bertasius2021space}. Therefore, a recent research trend has focused on computationally efficient attention mechanisms. Various heuristic based optimized attention mechanisms have been proposed that use separable attention~\cite{bertasius2021space} and sparse attention~\cite{duke2021sstvos}. In contrast, deformable attention uses dynamic sampling of tokens~\cite{zhu2020deformable}. Another line of research is to use interpretability-aware dynamic sampling of tokens~\cite{pan2021ia}.

\subsection{Interpretability of vision transformers}

\paragraph{Model agnostic interpretability methods.}
In recent years, model agnostic interpretability methods have been used on transformers and to compare internal representations of transformers and ConvNets. Both visualization based and statistical analysis based model agnostic methods have been dominant in recent research. One possible reason for the popularity of model agnostic methods is that they already have shown impressive results in understanding ConvNet models. A related reason is that they allow comparisons between a transformer's internal representation and those of other models, \eg ConvNets. Among the dominant visual interpretability methods, kernel visualization~\cite{erhan2009visualizing,simonyan2013deep,krizhevsky2017imagenet,dosovitskiy2020image} and feature map visualization~\cite{ghiasi2022vision} were used to compare transformer and ConvNet representations. Several saliency based input contribution visualization approaches also have been widely explored in computer vision research~\cite{abnar2020quantifying,selvaraju2017grad,binder2016layer,voita2019analyzing,chefer2021transformer,chefer2021generic}. Among the prominent quantitative methods, statistical correlation based methods are used in understanding transformer representations and to compare them with ConvNets~\cite{andrew2013deep,raghu2017svcca,morcos2018insights,kornblith2019similarity}. Other approaches employed include Canonical Correlation Analysis (CCA) and Centered Kernel Alignment (CKA)~\cite{raghu2021vision}. These model agnostic interpretability methods revealed insights into feature representations of transformer models along with various similarities and dissimilarities with ConvNets. These insights enabled the computer vision community to understand the internal workings of transformer models, such as feature abstraction hierarchy, the relevance of input features in various models and the temporal modelling capabilities of state-of-the-art vision transformers.

\paragraph{Attention specific interpretability.}
A notable number of attention specific interpretability methods have been presented in recent years. Attention distance was used to quantify and understand the receptive fields of attention heads at different transformer depths~\cite{dosovitskiy2020image}. Attention heatmap visualization was used to visually understand the inter-token correspondence propagation~\cite{vig2019visualizing}. Another line of research uses attention weighted class specific input saliency visualization~\cite{abnar2020quantifying,barkan2021grad,qiang2022attcat,chefer2021generic}. These methods were mainly used to interpret input token contributions to different class outputs. Recent work on video segmentation also presented visual analysis of object queries learned by transformer decoders~\cite{karim2023med}. These attention specific interpretability methods have uncovered crucial information regarding the working of the attention mechanism from an analytic point of view. The revealed information allowed for understanding the representation learned by attention mechanism at different components with respect to network depth, such as feature aggregation ranges corresponding to network depth at encoding and learning object extremities in decoding.

\paragraph{Video temporal modeling interpretability.}
Recently, several curated or synthetic temporal datasets were used for the analysis of temporal modeling of various transformer and ConvNet models. These approaches mainly use a curated subset of large scale datasets~\cite{sevilla2021only} or synthetic datasets~\cite{ilic2022appearance,kowal2022deeper} with representative videos containing strong temporal dynamics. These datasets variously have been used to study models at the inference level as well as at finer levels of analysis (e.g., individual layers within a model as well as units within a layer). 
Another line of research uses feature map visualization to show temporal correspondence in latent representations of intermediate layers of video models~\cite{zhao2021interpretable}. Perhaps the most striking result that has emerged from these studies is that current transformer (as well as ConvNet) models are far weaker in representing and exploiting the temporal dimension compared to the spatial dimension.

\paragraph{Application of interpretability.}
Over recent years, interpretability has been applied beyond scientific understanding and debugging of model prediction~\cite{sundararajan2017axiomatic} to the development of self-explainable~\cite{rigotti2021attention,bohle2023holistically} and efficient models~\cite{tan2018learning, wu2018beyond,pan2021ia}. The recent increasing concern about transparent decision making has motivated research on inherently interpretable models \cite{european2021laying,ontario2021regulating}. ConceptTransformer (CT) was proposed as a general decoder module that can generate additional explanation in addition to the classification output~\cite{rigotti2021attention}. Another notable work in this area are B-cos transformers, which are capable of inherently providing explanations for their decisions~\cite{bohle2023holistically}. Although this area is less explored and mainly contains models for image classification, current trends indicate a growing interest in self-explainable models for other vision tasks, \eg video segmentation. A notable example of using interpretability analysis for model improvement is IA-RED${}^2$, which used interpretability to reduce model redundancy for improved computational efficiency~\cite{pan2021ia}.

\section{Future Work}
In this final section, several directions for future work are suggested.

\paragraph{Temporal modeling in videos.}
State-of-the-art transformer models have demonstrated promising success in various video understanding tasks, \eg action recognition~\cite{wu2022memvit,li2022mvitv2,liu2021swin} and video segmentation~\cite{kim2022tubeformer,wu2022seqformer,pan2022wnet,karim2023med}. Despite the performance gains on current benchmarks, interpretability experiments reveal their limited ability in capturing temporal dynamics and attributes the performance gain to the static bias of the datasets~\cite{kowal2022deeper,ilic2022appearance}. These findings explain their significant performance drop in scenarios requiring strong temporal dynamics modeling~\cite{sevilla2021only}. Improvement in encoding of temporal dynamics for video classification tasks can provide powerful feature encoding for video segmentation as well. In particular, improvement in temporal dynamics based feature disentangling has potential to achieve superior performance in video segmentation in challenging dynamic scenarios. From this point of view, future research on harnessing temporal information in video transformers is of paramount importance.

\paragraph{Better multiscale modeling for segmentation.} 
The requirement of fine localization in video segmentation has motivated approaches for aggregating features from multiple resolutions during encoding~\cite{karim2023med} and learning object attributes from multiscale features during decoding~\cite{cheng2022masked,cheng2021mask2formervid}. However, these methods show limitation in their multiscale modeling in both the encoder and decoder. First, the majority of the methods use attention guided multiscale feature encoding separately at individual scales \cite{fan2021multiscale,li2022mvitv2,cheng2022masked,cheng2021mask2formervid}, with only a few efforts using attention across scales~\cite{chen2021crossvit,karim2023med}. Second, multiscale feature generation with top down semantic propagation is limited to different variants of FPNs~\cite{lin2017feature,kirillov2019panoptic,liu2018path,ghiasi2019fpn,tan2020efficientdet}. A few recent efforts explored using attention-based approaches for top down semantic propagation~\cite{zhu2020deformable} and multiscale decoding~\cite{karim2023med,cheng2022masked,cheng2021mask2formervid}. These efforts on some specific categories of video segmentation indicate potential of attention-based approaches for multiscale feature aggregation. Further research is required to consider their generalization across a broader range of segmentation tasks.

\paragraph{Multimodal video segmentation.}
Prominent multimodal video segmentation includes text referred video segmentation~\cite{seo2020urvos,li2022you,botach2022end,wu2022language}, speech referred video segmentation~\cite{pan2022wnet} and audio-visual sound source segmentation~\cite{zhou2022audio}. It is evident from current research that cross-attention mechanisms have shown success in multimodal context propagation from text or speech to visual features. Recent work on multimodal video segmentation emphasized text referring video segmentation, leaving both speech-referred video segmentation and audio-visual sound source segmentation as critically under explored. The potential for both speech referring video segmentation and audio-visual sound source segmentation remains an area ripe for further exploration. 


\paragraph{Interpretability of spatiotemporal modeling.}
Current analytic methods to evaluate the temporal modeling capability of video transformers are mainly limited to evaluating models on curated or synthetic datasets~\cite{sevilla2021only,ilic2022appearance,kowal2022deeper}. Again, the majority of interpretability methods for transformer models are focused on image classification plus modest work on video action recognition, which leaves video segmentation significantly under explored. Visualization based interpretability of video segmentation transformers is very subjective and does not provide quantitative understanding. Lack of research for transformer interpretability with a focus on video segmentation is a notable constraint on the development of video segmentation models capable of generalisation over a wide range of scenarios. Improved interpretability of algorithms and representations is necessary to truly understand the role of attention on dense video understanding tasks and their enabling mechanisms. From this perspective, there is a demand for further research on video transformer interpretability methods equipped with statistical tools focused on segmentation tasks.

\paragraph{Scalable and efficient video transformers.}
Previous research toward scalable and efficient video transformers primarily uses various heuristics, \eg attention locality and separability. Such methods proposed to improve transformer efficiency using sparse attention~\cite{duke2021sstvos}, axially separable attention~\cite{bertasius2021space} or dynamic sampled attention~\cite{zhu2020deformable}. Currently, interpretability analysis in natural language processing has improved efficiency by pruning some attention heads. Interpretability-aware redundancy reduction of vision transformers also has been proposed that dynamically drops selected input tokens for computational efficiency~\cite{pan2021ia}. Such interpretability driven scalable and efficiency improvement is an under explored area and can contribute to the computer vision community.

\paragraph{Self-explainable video transformers.}
The growing interest in explainable AI due to social and regulatory reasons has lead to research interest in inherently interpretable vision transformer models~\cite{european2021laying,ontario2021regulating}. These concerns only add to the fundamental scientific motivation of understanding the details of model operations. Although there are some recently proposed approaches, \eg concept transformers~\cite{rigotti2021attention} and
B-cos transformers~\cite{bohle2023holistically}, research in this direction is under explored. Again, self-explainable models are mainly limited to image classification. Hence, self-explainable video transformers are worth further investigation.

\ssp
\renewcommand{\bibname}{References}
\bibliographystyle{plain}
\bibliography{APPENDIX/References}

\end{document}